\documentclass{article}

\usepackage[final]{nips_2018-1}

\usepackage[utf8]{inputenc} %
\usepackage[T1]{fontenc}    %
\usepackage{hyperref}       %
\usepackage{url}            %
\usepackage{booktabs}       %
\usepackage{amsfonts}       %
\usepackage{nicefrac}       %
\usepackage{microtype}      %

\usepackage{amsmath}
\usepackage{amssymb}
\usepackage{amsthm}
\usepackage{cases}

\usepackage{algorithm}
\usepackage{algorithmic}

\usepackage{multirow}
\usepackage{tabularx}

\usepackage{graphicx}
\usepackage{epstopdf}
\usepackage{subfigure}
\usepackage{wrapfig}

\usepackage{natbib}
\usepackage[capitalize]{cleveref}
\newtheorem{theorem}{theorem}[section]

\newtheorem{remark}[theorem]{Remark}

\usepackage{xspace}
\usepackage[disable]{todonotes}
\usepackage{url}

\newcommand{\trace}[1]{\textbf{trace}(#1)}

\newcommand{\IND}[1]{\mathbb{I}_{\{ #1 \}}}

\newcommand{\R}{\mathbb{R}}

\newcommand{\one}{\mathbf{1}}

\newcommand{\E}{\mathbb{E}}
\newcommand{\A}{\mathcal{A}}
\newcommand{\X}{\mathcal{X}}
\newcommand{\eps}{\varepsilon}

\newcommand{\ksd}{\mathcal{S}}

\DeclareMathOperator{\argmax}{argmax}

\newlength\myindent
\newcommand\bindent{%
  \begingroup
  \setlength{\itemindent}{\myindent}
  \addtolength{\algorithmicindent}{\myindent}
}
\newcommand\eindent{\endgroup}

\title{Adaptive MCMC via Combining Local Samplers}

\author{
  Kiarash Shaloudegi$^1$ \\[1ex]
  $^1$Depeartment of Electrical and\\  Electronic Engineering \\
  Imperial College London, London, UK \\
  \texttt{k.shaloudegi16@imperial.ac.uk} \\
  \And
  Andr\'as Gy\"orgy$^{2,1}$ \\[1ex]
  $^2$DeepMind \\
  London, UK \\
  \texttt{agyorgy@google.com}
  \And \mbox{} \hspace{-1cm} \\
}

\begin{document}

\maketitle

\begin{abstract}
Markov chain Monte Carlo (MCMC) methods are widely used in machine learning. One of the major problems with MCMC is the question of how to design chains that mix fast over the whole state space; in particular, how to select the parameters of an MCMC algorithm. Here we take a different approach and, similarly to parallel MCMC methods, instead of trying to find a single chain that samples from the whole distribution, we combine samples from several chains run in parallel, each exploring only parts of the state space (e.g., a few modes only). The chains are prioritized based on kernel Stein discrepancy, which provides a good measure of performance locally. The samples from the independent chains are combined using a novel technique for estimating the probability of different regions of the sample space. Experimental results demonstrate that the proposed algorithm may provide significant speedups in different sampling problems. Most importantly, when combined with the state-of-the-art NUTS algorithm as the base MCMC sampler, our method remained competitive with NUTS on sampling from unimodal distributions, while significantly outperforming state-of-the-art competitors on synthetic multimodal problems as well as on a challenging sensor localization task.\\[0.3em]
\end{abstract}

\section{Introduction}

We consider the problem of computing expectations $\mathbb{E}_P[f(X)]=\int_{\X}f(x)p(x)dx$ for some complicated target distribution $P$ with density $p$ over a set $\X \subset \R^d$ and a target function $f:\mathcal{X}\to\mathbb{R}$. Such expectations often arise in Bayesian inference and maximum likelihood estimation \citep{andrieu_introduction_2003,brooks_handbook_2011}. Oftentimes, $p$ has a closed form except for an unknown normalization constant, making the computation of the integral especially challenging  \citep{andrieu_introduction_2003}.\footnote{In most of  real world applications finding this normalization constant is as hard as computing the original expectation under $P$. For example, when $P$ is obtained as a posterior distribution, $p$ can usually be written as $p(x)=\hat{p}(x)/\int_\X \hat{p}(x') dx'$ from Bayes rule, where $\hat{p}(x)$ is available in closed form for any $x \in \X$, but the value of the integral is unknown and is hard to compute.}
Markov chain Monte Carlo (MCMC) methods are a family of numerical estimation methods, which are successfully applied to estimate the aforementioned expectations, especially in high-dimensional problems.    MCMC algorithms take random samples from an ergodic Markov chain with stationary distribution $P$, and approximate the expectation via averaging over the produced sample. 

The challenging problem in designing MCMC methods is to ensure that the distribution of the samples converge to $P$ fast, and,
in practice,  usually some  domain-specific knowledge is used in the design of their proposal distributions to achieve fast convergence  \citep{andrieu_introduction_2003}.  This need for specialized  design led to the development of dozens of methods for each problem,  each of which has their own tunable parameters \citep{neufeld_adaptive_2014}. Consequently, choosing the right method with corresponding parameters to achieve fast convergence is quite difficult and requires considerable time and effort. 

A large body of work has been devoted in the literature to address this difficulty and to find ways to set the algorithms' parameters optimally; for instance, optimal tuning  of the Metropolis-Hasting algorithm (\citealp{roberts_optimal_2001,bedard_optimal_2008,roberts_weak_1997,atchade_towards_2011}; \citealp[Chapter  4]{brooks_handbook_2011}). The problem with this line of research is that the solutions rely on  some Markov chain parameters %
that are typically unknown \citep{latuszynski_adaptive_2013}.

A more promising  line of research to address the parameter setting issue is based on adaptive MCMC methods. In this framework,  the MCMC samples are used to learn about the target distribution, and the algorithms adjust their parameters as they progress \citep{latuszynski_adaptive_2013}. To do so, they rely on optimizing some objective function such as  expected squared jumping distance \citep{pasarica_adaptively_2010,wang_adaptive_2013}, the area  under the autocorrelation function up to some specific lag  \citep{mahendran_adaptive_2012}, the difference between the proposal covariance matrix and its empirical estimation         \citep{haario_adaptive_2001,haario_dram:_2006,sejdinovic_kernel_2014,mbalawata_adaptive_2015},  or the difference between the optimal acceptance rate (in practice, a recommendation thereof) and its empirical estimate  \citep{yang_automatically_2014}. Perhaps the most successful adaptive method for finding the optimal parameters (i.e., the number of leapfrog steps and the step size) in Hamiltonian Monte Carlo (HMC), called no-U-turn sampler (NUTS), is based on monitoring when the sample trajectories ``turn back'', and the resulting algorithm provides state-of-the-art performance in a number of problems \citep{hoffman_no-u-turn_2014}. For more details about adaptive MCMC methods, the reader is referred to the tutorial paper by \citet{andrieu_tutorial_2008}.

In practice, making sure that a sampler can move between distant modes of the target distribution cannot be guaranteed by the aforementioned adaptive methods.\footnote{To be more precise, instead of modes the ``high-probability'' regions of the target distribution are of interest, but, for simplicity and following the standard language of MCMC literature, we will often refer to these as modes throughout the paper.} There are  two main approaches to deal with distant modes: (i) running parallel chains and combining their final samples; and (ii) sampling from powers of the target function (the inverse power is referred to as temperature), known as annealing. Indeed, in the literature, there are several successful methods based on combining these two ideas, such as parallel tempering \citep{earl_parallelTT_2005} and (annealed) sequential Monte Carlo samplers, also known as particle filters \citep{doucet_into_sequential_2001,demoral_sequential_2006}. 
The core idea of these methods and their variants is to run several chains with different temperatures and  periodically exchange information between them. By increasing the temperature of a target distribution, it flattens (becomes more uniform), and the chains at high temperatures act as global search algorithms. They pass the information regarding the location of the  modes to the chains with lower temperatures, which can effectively  search each mode.  An alternative to parallelization is the idea of \textit{regeneration} \citep{mykland_regeneration_1995,ahan_regeneration_2013}. Regeneration partitions a long Markov chain into smaller independent segments such that the samples are unbiased in each segment, hence can be combined together without any further consideration. This process makes it possible to combine samples from different samplers and tune the parameters of the Markov chain after each regeneration. Although theoretically elegant, the application of regeneration methods is limited in practice since they require a properly tuned distribution for detecting regenerations.  

In this paper we combine the strengths of adaptive and parallel MCMC methods. Instead of trying to find a single sampler that approximates the target distribution well on its whole domain, we run several samplers and select which sampler to use at any given time in a \emph{sequential} manner, based on all the samples obtained before.
Our main contributions are (i) adapting the bandit-based  adaptive Monte Carlo (MC)\footnote{Throughout the paper, we refer to samplers taking unbiased, independent samples as MC methods} method of \citet{neufeld_adaptive_2014} to MCMC; and (ii) a novel method for combining samples from multiple chains. The resulting algorithm is suitable for sampling from challenging multimodal distributions and is fairly insensitive to the choice of its parameters. The next subsection gives a more detailed overview of our approach and the corresponding challenges.

\subsection{Approach and challenges}
\label{sec:approach}

In the simple case when all MCMC samplers mix well over the whole domain, our goal is to use the best sampler (which mixes the fastest) most of the time. This is very similar to the case of choosing from several unbiased MC samplers. For the latter, \citet{neufeld_adaptive_2014} showed that scheduling which samplers to use at any point in time is equivalent to a stochastic multi-armed bandit problem \citep{bubeck_regret_2012}. This makes it possible to apply bandit algorithms to select which sampler to use to get the next sample, and the decision depends on the overall performance of the samplers so far, measured by the variance for each sampler. Extending the same idea to the MCMC case is not trivial, since measuring the quality of MCMC samplers is a much harder task. In fact, until recently, there has not been any  empirical measure that can monitor the sample convergence. Common MCMC diagnostics such as effective sample size, trace and mean plots, or asymptotic variance  assume the chain asymptotically converges to the target distribution, so they cannot detect asymptotic bias \citep{gorham_measuring_2015}.  To address this issue, \cite{gorham_measuring_2015}  developed an empirical sample-quality measure that can detect non-convergence (or bias) based on Stein's method. A kernelized version of this measure, called \emph{kernel Stein discrepancy} (KSD) was subsequently developed by \citet{liu_kernelized_2016,chwialkowski_kernel_2016,gorham_measuring_2017}, which can be used to compare the quality of different samplers. 
\emph{As our first contribution, we extend the bandit-based racing method of \citet{neufeld_adaptive_2014} to MCMC samplers by using the KSD measure as the loss function in the bandit algorithms.} This is described in detail in \Cref{sec:unimodal} while the background on KSD is given in \Cref{sec:KSD}.

On the negative side, KSD is not able to detect underfitting if the target distribution has well-separated modes, so it cannot distinguish between two samplers such that one samples only one mode while the other samples both modes, and the samples are equally good locally  \citep{liu_kernelized_2016,chwialkowski_kernel_2016,gorham_measuring_2017}. This brings us to the next problem: namely, MCMC methods using a single chain usually fail to explore the whole domain if the support has reasonably high-probability regions separated by low-probability regions (of course, such notion of separation depends on the actual sampler used). Setting the parameters of the samplers to deal with this issue, or simply detecting its presence, is hard, and to our knowledge no practical solutions thereof are available. To alleviate this problem, following the parallel MCMC framework, we run several chains in parallel only expecting that they provide good samples locally. The hope is that the multiple instances will explore the space sufficiently, finding all the important regions of the support. Then, in the end, we combine all the samples (from all the samplers) to approximate the target distribution.
This is challenging for two reasons: (i) it is not obvious how the samples from different samplers should be weighted, and (ii) we do not want to waste resources to run several samplers exploring the same region of the domain. For (ii), we apply our bandit-based racing method locally (\Cref{sec:localmix}), while
to address (i), \emph{we develop a method to estimate the probability of the region a set of samples cover based on R\'enyi-entropy estimates \citep{pal_renyi_2010}, and use these to weight the samples, which is our second main contribution.} This is described in \Cref{sec:weight_estimation}.

Our final sampling algorithm is put together in \Cref{sec:finalalg}.
Lastly, in \Cref{sec:exp} we demonstrate through a number of experiments that our method is competitive with state-of-the-art adaptive MCMC methods, such as 
the no-U-turn sampler NUTS \citep{hoffman_no-u-turn_2014} or the recent sample reweighting method of \cite{liu_black-box_2016}  on simpler cases when the distribution is concentrated on a single ``connected'' region, while significantly outperforming the competitors, including parallel tampering and sequential MC, on harder problems where high-probability regions are separated by areas of low-probability, as well as on a challenging sensor-localization problem.

\section{Measuring sample quality}
\label{sec:KSD}

As mentioned in \Cref{sec:approach}, measuring the quality of samples produced by an MCMC algorithm is crucial in our approach. To this end we are going to use the recently introduced  \emph{kernel Stein discrepancy} (KSD) \citep{liu_kernelized_2016,chwialkowski_kernel_2016,gorham_measuring_2017}. 

To measure the quality of a set of samples, we want to quantify how well a probability distribution $Q_n$ over $\R^d$ can approximate the target distribution $P$; in this section we assume that the density $p$ of $P$ is positive and differentiable on the whole $\R^d$. Oftentimes, $Q_n$ is given by a weighted sample $\{(x_i,q_i)\}_{i \in [n]}$, where $[n]$ denotes the set $\{1,\ldots,n\}$, $x_i \in \R^d$ and $Q_n(x_i)=q_i>0$ with $\sum_{i \in [n]} q_i =1$.  One way to do this is to measure the maximum expectation error over a class of real-valued test functions $\mathcal{H} \subset \{f:\R^d \to \R\}$:
\begin{equation}
\mathcal{D}_{\mathcal{H}}(Q_n, P) \triangleq \sup_{h\in \mathcal{H}} |\E_{Q_n}[h(Z)]-\E_{P}[h(X)]|,
\label{eq:ipm}
\end{equation}
where, in case of a weighted sample, $\E_{Q_n}[h(Z)] = \sum_{i=1}^n q_i h(x_i)$ (we use $Z$ to distinguish the sample from $X$).
If the class of test functions $\mathcal{H}$ is large enough, for any sequence of probability measures $(Q_n)_{n\geq1}$,  the convergence of  $\mathcal{D}_{\mathcal{H}}(Q_n, P)$  to zero implies that $Q_n$ converges weakly to  $P$ ($Q_n \Rightarrow  P$). 
One advantage of using this formulation is that we can recover different  metrics in the form \eqref{eq:ipm} by changing the class of test functions $\mathcal{H}$ \citep{gorham_measuring_2015}; for example, using $\mathcal{H}\triangleq\{h:\X\to\mathbb{R}|\sup_{x\in\X}|h(x)|\leq1\}$, the measure in \eqref{eq:ipm} becomes the \emph{total variation distance}, while using the class of Lipschitz-continuous functions
$\mathcal{H}\triangleq\{h:\X\to\mathbb{R}|\sup_{x\neq y\in\X}\frac{\|h(x)-h(y)\|_2}{\|x-y\|_2} \le 1 \}$
recovers the Wasserstein distance. Note that typically finding the exact solution in \eqref{eq:ipm}  is as hard as the original problem since one needs to calculate  $\E_{P}[h(X)]$; however, if we can find a class of functions $\mathcal{H}$ such that $\E_{P}[h(X)] = 0$ for any function $h \in \mathcal{H}$ and $\mathcal{H}$ is sufficiently rich so that \eqref{eq:ipm} is still a meaningful measure of similarity, we can avoid this problem. 

To this end, let $k: \mathbb{R}^d \times \mathbb{R}^d \to \mathbb{R}$ be a positive definite kernel and  $\mathcal{K}$ its associated reproducing kernel Hilbert space (RKHS),  $\|\cdot\|_{\mathcal{K}}$ be the induced norm from the inner product in $\mathcal{K}$. %
Then, for $x\in\mathbb{R}^d$, and $f\in \mathcal{K}$, we have $f(x)=\langle f,k(x,\cdot)\rangle$. 
Given this kernel and a norm $\|\cdot\|$ on $\R^d$ with dual norm $\|\cdot\|^*$,%
\footnote{The dual norm $\|\cdot\|^*$ is defined as $\|x\|^*=\sup_{y\in\R^d, \|y\|=1}\langle x, y \rangle$  for any $x \in \R^d$.}
\citet{gorham_measuring_2017} defined the \emph{kernel Stein set} $\mathcal{F}_{k,\|\cdot\|}$ as 
\[
\mathcal{F}_{k,\|\cdot\|}\triangleq\{f=(f_1,\ldots,f_d)\in\mathcal{K}^d| \;\|v\|^*\le 1 \text{ for } v_j=\|f_j\|_{\mathcal{K}}, j=1,\ldots,d\},
\]
and the Langevin Stein operator $T_p:\{f:\R^d\to\R^d\} \to \{h:\R^d \to \R\}$ as 
\[
	(T_p f)(x)\triangleq \frac{1}{p(x)}\langle \one, \nabla(p(x)f(x)) \rangle=\langle \nabla\log (p(x)),f(x)\rangle+\langle \one, \nabla f(x)\rangle,
\]
where $\one$ denotes the $d$-dimensional all-one vector (with each component being $1$).
\citet{gorham_measuring_2017} showed that that if $k$ is a continuous and bounded function with a continuous and bounded second derivative $\nabla_x\nabla_y k(x,y)$ and $\E_{P}[\|\nabla \log p(X)\|_2]<\infty$, then $\E_{P}[((T_pf)(X))]=0 $ for all $f\in\mathcal{F}_{k,\|\cdot\|}$. Thus, with $\mathcal{H}=T_p  \mathcal{F}_{k,\|\cdot\|}=
\{h | \; f=T_p f, f \in \mathcal{F}_{k,\|\cdot\|}\}$, $P$ cancels from the definition \eqref{eq:ipm} reducing it to 
\begin{equation}
\label{eq:KSD1}
\ksd(Q_n)=\mathcal{D}_{T_p\mathcal{F}_{k,\|\cdot\|}}(Q_n,P)=\sup_{h\in T_p \mathcal{F}_{k,\|\cdot\|}} |\E_{Q_n}[h(Z)]|
\end{equation}
where we suppressed the dependence on $P$, $T_p$, and $\mathcal{F}_{k,\|\cdot\|}$ in the notation $\ksd(Q_n)$.  This makes it possible to compute the optimum in the definition under the above conditions on $k$. Let $s_p(x)\triangleq \nabla \log p(x)$ with $s_{p,i}(x)$ denoting its $i$th coordinate for any $i=1,\ldots,d$, and define
	\begin{equation*}
		\begin{split}
		k_p^i(x,x')& \triangleq  s_{p,i}(x)^\top k(x,x') s_{p,i}(x')+s_{p,i}(x)^\top\nabla_{x'_i}k(x,x')
			+s_{p,i}(x')^\top\nabla_{x_i}k(x,x')+\nabla_{x_i}\nabla_{x'_i}k(x,x'), \\
			k_p(x,x')& \triangleq  \sum_{i=1}^d k_p^i(x,x') \\
			& = s_p(x)^\top k(x,x') s_p(x')+s_p(x)^\top\nabla_{x'}k(x,x')
			+s_p(x')^\top\nabla_{x}k(x,x')+\trace{\nabla_{x}\nabla_{x'}k(x,x')},
		\end{split}
	\end{equation*}
Then, if $\sum_{i=1}^d \E_{Z,Z' \sim Q_n}[k_p^i(Z,Z')^{\frac{1}{2}}]<\infty$ (where $Z$ and $Z'$ are independently drawn from $Q_n$) the resulting maximum becomes
\begin{equation*}
\ksd(Q_n)=\sqrt{\E_{Q_n\times Q_n}[k_p(Z,Z')]}.
\end{equation*}
When $Q_n$ is a weighted sample $\{(x_i,q_i)\}_{i=1,\ldots,n}$, this simplifies to
\begin{equation}
\label{eq:KSD}
\ksd(Q_n)=\sqrt{\mathbf{q}^\top K_{p,n} \mathbf{q}}
\end{equation}
where $\mathbf{q}=(q_1,\ldots,q_n)$ and $K_{p,n}=[k_p(x_i,x_j)]_{i,j\in[n]}=[\sum_{l=1}^d k_p^l(x_i,x_j)]_{i,j\in[n]}$. We call $\mathcal{S}(Q_n)$ the \emph{kernel Stein discrepancy}. Note that $\mathcal{S}(Q_n)$ can be computed with our information about $p$, since it only depends on $p$ through $s_p(x)= \nabla \log p(x)$, which cancels the effect of the unknown normalization constant. Also, $\ksd(Q_n) \to 0$ for any norm $\|\cdot\|$, if and only if it converges to zero for $\|\cdot\|=\|\cdot\|_2$; hence, in the rest of the paper we assume that $\|\cdot\|$ is the Euclidean norm.

Under some technical conditions the KSD measure goes to $0$ if and only if $Q_n$ converges weakly to  $P$ \citep{gorham_measuring_2017}:
In particular,  if $k$ is the inverse multiquadratic (IMQ) kernel $k(x,y)=(c^2+\|x-y\|_2^2)^\beta$ for some $c>0$ and $\beta \in (-1,0)$, 
$Q_n \Rightarrow P$ if $\ksd(Q_n) \to 0$.  Furthermore, if $\nabla \log p$ is Lipschitz with $\E_P[\|\nabla \log p(X)\|_2^2]$,
$\ksd(Q_n) \to 0$ whenever the Wasserstein distance of $Q_n$ and $P$ converges to zero (which, in turn, implies $Q_n \Rightarrow P$).
Also note that popular choices of kernels, such as the Gaussian kernel or the Mat\'ern kernel do not detect the non-convergence of the distribution $Q_n$ to $P$ for $d \ge 3$ (there are cases when $\ksd(Q_n) \to 0$ but $Q_n \not\Rightarrow P$), and the situation is the same for the IMQ kernel defined above with $\beta<-1$ for $d \ge 2\beta/(1+\beta)$. Note that these kernel do work for $d=1$.

However, despite of the above nice theoretical guarantees, one might not be able to detect convergence based on KSD for practical sample sizes, especially when the target distribution is multimodal and the modes are well-separated: \citet{gorham_measuring_2016} demonstrated that for a one-dimensional Gaussian mixture target distribution (with two components), for practical sample sizes, the KSD measure fails to distinguish between two sets of samples, one drawn independently from one mode and the other drawn independently from the whole target distribution. Furthermore, KSD  requires even more samples to distinguish between the two cases as the modes' distance increases (see Section 6.1 of \citet{gorham_measuring_2016} for more details).   
Another issue is that the complexity of computing the KSD for an empirical distribution is quadratic in the sample size, which quickly becomes infeasible as the sample size grows.

An interesting property of the KSD measure is that it is convex: if $Q^1,\ldots,Q^M$ are distributions and $Q=\sum_{i=1}^M w_i Q^i$ for nonnegative weights $w_i$ satisfying $\sum_{i=1}^M w_i=1$, we have
\begin{align}
\ksd(Q) &= \sup_{f \in \mathcal{F}_{k,\|\cdot\|}} \left| \E_{Q}\left[  \left( T_p f\right) (Z)  \right] \right| = \sup_{f \in \mathcal{F}_{k,\|\cdot\|}} \left|  \sum_{i = 1}^{M} w_i \E_{Q^i} \left[ \left( T_p f\right) (Z) \right] \right|  \nonumber \\
& \le  \sup_{f \in \mathcal{F}_{k,\|\cdot\|}}   \sum_{i = 1}^{M} w_i \left| \E_{Q^i} \left[ \left( T_p f\right) (Z) \right] \right| 
 \le   \sum_{i = 1}^{M} w_i   \sup_{f \in \mathcal{F}_{k,\|\cdot\|}} \left| \E_{Q^i} \left[ \left( T_p f\right) (Z) \right] \right| 
 = \sum_{i=1}^{M}w_i  \ksd( Q^i)~.
\label{eq:KSD-convex}
\end{align}

\section{Sequential selection of samplers}
\label{sec:problem_formulation}

In this section we present several strategies to select from a pool of MCMC samplers in a sequential manner. In all of our algorithms the selection of the sampler to be used next depends on the quality of the samples generated by the different samplers, where the quality will be measured by the KSD measure (or its approximations). Formally, assume we have access to $M$ MCMC samplers (e.g., multiple sampling methods and/or multiple instances of the same sampling algorithm with different parameters, such as starting point or step size), and denote the set of samplers by $[M]$. At every step of the algorithm, we select one of the samplers and use it to produce the next  batch of samples.

\subsection{Mixing samplers}
\label{sec:unimodal}
First we consider the case when each sampler is \emph{asymptotically unbiased} (or \emph{consistent}), that is,  generates samples with an empirical distribution converging weakly to the target distribution $P$ almost surely (this is usually satisfied for any standard MCMC sampler when $P$ is unimodal). Our task is to sequentially allocate  calls among the $M$ samplers  to minimize the KSD measure of the set of samples we collect.  In order to do so, we design an algorithm which gives preference to samplers where the convergence is faster.  This setup is similar to the one considered by \citet{neufeld_adaptive_2014}, who designed sequential sampling strategies for MC samplers generating independent and identically distributed (i.i.d.) samples, by selecting samplers with smaller variances based on multi-armed bandit algorithms \citep{bubeck_regret_2012}. 
In this section we generalize their method to MCMC samplers, aiming to minimize the total KSD measure (cf. Eq.~\ref{eq:KSD}) of the sample instead of the variance (recall that for practical sample sizes--before convergence--the variance of the sample is not a good measure of quality for MCMC samplers). However, computing the KSD measure is quadratic in the sample size, and so it becomes computationally infeasible  even for relatively small sample sizes---note that any computation we spend on selecting samplers could also be used for sampling.  Therefore, we are going to approximate the KSD measure as the average KSD over smaller blocks of samples.

For a sampler with total sampling budget $n$, we break the sampling process into $T$ rounds: At each round the sampler takes a batch of samples of size $n_b=n/T$.  Let $S_t$ be the KSD measure of samples from the $t$th round; we approximate $\tilde{S}_n$, the KSD of the full sample of size $n$ with the average $(1/T)\sum_{t=1}^T S_t$.
We  call this the \emph{block-diagonal approximation}, as it corresponds to a block-diagonal approximation of the kernel matrix $K_{p, n}$ in computing  \eqref{eq:KSD}.
To quantify the accuracy of the approximation, we assume that there exists a function $g(t,n_b)$ such that  $\lim_{\frac{n_b}{t}\to \infty} g(t,n_b)=0$ and
$\frac{1}{t}\sum_{b=1}^{t} \E[S_b]-\E[\tilde{S}_{t n_b}] \le g(t,n_b)$ for all $t=1,\ldots,T$.
Using the block-diagonal approximation, we relax our goal to competing with a sampler with the smallest average approximate block-KSD measure $(1/T) \sum_{t=1}^T S_{i,t}$, where $S_{i,t}$ is the KSD measure of the $n_b$ samples generated by sampler $i$ when it is called the $t$th time. 
We also assume that this is close to the KSD measure $\tilde{S}_{i,t}$ of a sample of size $t n_b$ obtained from using sampler $i$ for $t$ blocks, that is,
$\frac{1}{t}\sum_{b=1}^{t} \E[S_{i,b}]-\E[\tilde{S}_{i,t n_b}] \le g(t,n_b)$ for all $t=1,\ldots,T$.
Experimental results presented in \Cref{sec:exp_block_error} indicate that the block-diagonal approximation mostly preserves the ranking of the samplers (as defined by the true KSD measure), hence we pay very little price for the computational advantage we get.

Furthermore, solving this problem is well-suited for any bandit algorithm; here we adapt the UCB1 method of \citet{auer_finite-time_2002}. The resulting algorithm, which we call  \emph{KSD-UCB1} and is given in \Cref{algorithm:ucb}\footnote{The algorithm assumes that $S_{i,t} \in [0,1]$, which can be achieved by rescaling the KSD measures. In practice, a reasonable estimate for the range can be obtained from $S_{i,1}$ for each sampler $i$.}, keeps track of an optimistic estimate of the average approximate KSD value for each sampler, and every time selects a sampler whose performance is estimated to be the best possible (based on high-probability confidence intervals).

\begin{algorithm}[tb]
	\caption{KSD-UCB1}\label{algorithm:ucb}
	\begin{algorithmic}
		\FOR{$i \in \{1,\ldots,M\}$}
		\STATE - Use sampler $i$ to generate $n_b$ samples; observe $S_{i,1}$; set $\bar{\mu}_{i,1} = S_{i,1}$ and $T_{i}(1)= 1$.
		\ENDFOR
		\FOR {$t\in \{M+1,M+2,\ldots,n\}$}
		\STATE - Play arm $i$ that minimizes $ \bar{\mu}_{i,t-1}-\sqrt{\frac{2\log t}{T_i(t-1)}}$.
		\STATE - Set $T_i(t)=T_i(t-1)+1$ and $T_j(t)=T_j(t-1)$ for $j\neq i$.
		\STATE - Observe $S_{i,T_i(t)}$, and compute $\bar{\mu}_{i,t}=(1-\frac{1}{T_i(t)})\bar{\mu}_{i,t-1}+\frac{S_{i,T_i(t)}}{T_i(t)}$.
		\ENDFOR
	\end{algorithmic}
\end{algorithm}

If the KSD values $S_{i,1},\ldots,S_{i,T}$ were i.i.d., the standard bandit regret bound \citep{bubeck_regret_2012} would yield
$\sum_{t=1}^T \E[S_t] - \min_{i \in [M]}\sum_{t=1}^T \E[S_{i,t}] = O(\log T)$. 
In this case the convexity of the KSD measure (recall Eq.~\ref{eq:KSD-convex}) would imply that after $T$ rounds of sampling,
\[
\E[\tilde{S}_{T n_b}] - \min_i \E[\tilde{S}_{i,T n_b}] \le \frac{1}{T} \left(\sum_{t=1}^T \E[S_t]  - \min_{i \in [M]} \sum_{t=1}^T \E[S_{i,t}] \right)+ 2g(T,n_b)
\approx \frac{O(\log T)}{T}+ 2g(T, n_b)~.
\]
This shows that increasing $T$ and $n_b$, the performance of KSD-UCB1 would be close to that of the best sampler. However, in our case, the $S_{i,t}$ are not i.i.d. Assuming that the samplers mix (which is reasonable for a single mode distribution), the $S_{i,t}$ are getting closer and closer to be sampled i.i.d. as $n_b$ increases. Also, as mentioned above, the effect of the block-diagonal approximation (and hence that of $g(T,n_b)$) is small in practice (see also \Cref{sec:exp_block_error}).
The exact parametrization of KSD-UCB1 corresponds to the assumption that the block-KSD values are in $[0,1]$. While this is not true in general, in practice we estimate the maximum KSD value from samples and normalize the values accordingly, before feeding them to the KSD-UCB1 algorithm.

\subsection{Locally mixing samplers}
\label{sec:localmix}
In practice, if the target distribution is multimodal and the modes are far from each other, MCMC methods often get stuck in some of the modes and fail to explore all the regions where $P$ is supported; while eventually all asymptotically consistent methods reach each mode, this may not happen after any practically reasonable time. To model this situation, we assume that the support of $P$ is partitioned into sets $A_1,\ldots,A_K$ with $P(A_j)>0$ for all $j \in [K]$ (the $A_j$ are pairwise disjoint and their union is the support of $P$) such that the empirical distribution of the samples generated by sampler $i \in [M]$ converges weakly to $P(\cdot|A_j)$ for some $j \in [K]$, where $P(\cdot|A)$ is the conditional distribution of $P$ over $A$. We  refer to the sets $A_j$ as \emph{regions}, and a sampler satisfying the above condition almost surely a \emph{locally mixing} sampler.

For simplicity we first consider the case where there is one sampler in each region $A_i$ (consequently $M = K$). This  setup  is similar to stratified sampling: The idea is to partition the domain into non-overlapping regions (a.k.a., strata), draw samples from each region, and combine the final samples to estimate $\E_p[f(X)]$ \citep{owen_montecarlobook}. The problem in stratified sampling is to find the optimal number of samples that need to be taken from each stratum in order to minimize the  Monte Carlo integration error. Given the total number of sample $n$, the optimal strategy for minimizing the mean squared error (MSE) is to sample each stratum $n_i = \frac{\sigma_{i} P(A_i)}{\sum_{i=1}^{K}\sigma_{i} P(A_i)}n$ times (relaxing the integrality constraints), where $\sigma_i$ is the conditional standard deviation of $f(X)$ given that $X$ falls into the $i$th region \citep{carpentier_adaptive_2015}. 

One can immediately see that the problem we consider in this subsection is very similar to stratified sampling, with the important differences that our samplers are not i.i.d., and we do not minimize the squared error but the KSD measure. Denoting the distribution of samples from region $A_i$ by $Q_{n,i}$ after taking $n$ samples in total, let $w_i$ denote the weight of sampler $i$ generating these samples (recall that here, by assumption, we have one sampler in each region). Then our total weighted sample distribution becomes
\begin{equation}
\label{eq:Qn}
\textstyle Q_n = \sum_{i=1}^M w_i  Q_{n,i}.
\end{equation}
Since according to our assumptions, $Q_{n,i}$ converges weakly to $P(A_i)$ almost surely, for every $i$, we need to have $w_i \to P(A_i)$ for all $i \in [M]$ to ensure that $Q_n$ converges to $P$ weakly (we will refer to this as the $w_i$ being asymptotically consistent). 
A procedure for estimating $w_i$ this way will be given in the next section.
Assuming for a moment that $w_i=P(A_i)$, using the convexity of the KSD measure (see Eq.~\ref{eq:KSD-convex}), %
we have
\begin{align}
\tilde{S}_n =\ksd(Q_n) \le  
 \sum_{i=1}^{M}w_i  \ksd( Q_{n,i})
= \sum_{i=1}^{M} w_i \tilde{S}_{i,n_i}, \label{eq:tSbound}
\end{align}
where, as before, $\tilde{S}_n$ and $\tilde{S}_{i,n_i}$  denote the KSD measures of the whole sample and, resp., that of sampler $i$ (with number of samples $n_i$  obtained by sampler $i$).
Based on this inequality, we could aim for minimizing $ \sum_{i=1}^{M}w_i   \tilde{S}_{i,n_i}$ and use the ideas from adaptive stratified sampling \citep{carpentier_adaptive_2015}; however, multiple challenges preclude us from doing this: (i)  $w_i$ is not known in advance; (ii) the stratified sampling algorithm is based on the known convergence rate of the sample average, but we do not know how fast $ \tilde{S}_{i,n_i}$ approaches zero (as a function of $n_i=w_i n$); and (iii) the computational complexity of calculating $ \tilde{S}_{i,n_i}$ is $O(n_i^2)$. To handle (i), we will address the problem of estimating $w_i$ in \Cref{sec:weight_estimation}.  For (ii), a conservative approach is to uniformly minimize  $w_i \tilde{S}_{i,n_i}$, hence selecting sampler $i$ for which this quantity is the largest. For (iii), we  again use a block-diagonal approximation to $\tilde{S}_{i,n_i}$, which causes problems with (ii), since the estimate does not converge to $0$ for a fixed block size. In \Cref{sec:exp_multimode_single} we present experiments with different strategies under different setups, but the strategies considered seem to perform rather similarly. 
The simplest strategy considered is to select regions uniformly at random. Another approach is to minimize the maximum $\tilde{S}_{i,n_i}$, that is, choosing $i=\argmax \tilde{S}_{i,n_i}$, which does not require the knowledge of the weights $w_i$, while we also test strategies selecting regions based on their estimated weights or, similarly to stratified sampling, based on their estimated variance. Although quite different, the strategies considered seem to perform rather similarly in various settings.

\begin{algorithm}[tb]
	\small
	\caption{KSD-UCB1-M} \label{algorithm:ksducb1m}
	\begin{algorithmic}
		\STATE {\bfseries Given:} (Unnormalized) density $\hat{p}$; partition of the domain $A_1,\ldots,A_K$; $M$ samplers in $K$ classes: samplers in $\A_i$ sample from $p(\cdot|A_i)$ for all $i\in[K]$, $\cup_{i\in [K]} \A_i = [M]$; total number of rounds: $T$; batch size: $n_b$.
		\STATE {\bfseries Initialize:} Draw a batch of samples from each sampler $i$, observe $S_{i,1}$ and set $\hat{\mu}_{i,1} = S_{i,1}$ and $T_i(1) = 1$ for all $i \in [M]$.
		\FOR{$t \in \{M+1, \ldots, T\}$}
			\STATE - Select a region $I_t$. %
		 	\STATE - Draw a batch of samples from arm $i_t = \arg\min_{i \in A_{I_t}} \left( \mu_{i, t-1} - \sqrt{\frac{2 \log t}{T_i(t-1)}}\right)$. 
			\STATE - Set $T_{i_t}(t)=T_{i_t}(t-1)+1$ and $T_j(t)=T_j(t-1)$ for $j\neq i_t$.
			\STATE - Observe $S_{i_t,T_{i_t}(t)}$, and compute $\bar{\mu}_{i_t,t}=(1-\frac{1}{T_{i_t}(t)})\bar{\mu}_{i_t,t-1}+\frac{S_{i_t,T_{i_t}(t)}}{T_{i_t}(t)}$.
		\ENDFOR
		\STATE - Reweight the samples of each mode proportional to its weight, $\frac{w_k}{Tn_b}$.
	\end{algorithmic}
\end{algorithm}

Unfortunately, we cannot guarantee that we start samplers in such a way that we have a single sampler for each region. If we know which sampler belongs to which region (recall that we assume that the samplers are locally mixing and hence belong to a region $A_i$), we can combine any of the region selection strategies described above with the bandit method described in \Cref{sec:unimodal} in a straightforward way: in each region $A_i$, run an instance of KSD-UCB1 over the samplers exploring this region, and use this KSD-UCB-1 instance as a single  locally mixing sampler  in any of the above region selection methods. Finally, when the sampling budget is exhausted ($n$ samples are generated), estimate the weights $w_i$, and reweight the samples corresponding to region $A_i$ according to \eqref{eq:Qn}. We refer to this procedure as \emph{KSD-UCB1-M} (M stands for Multiple regions); the pseudocode of the algorithm is given in \Cref{algorithm:ksducb1m}.  Clearly, since the bandit algorithms sample each sampler infinitely often, we have the following consistency result:

\emph{ Under the local mixing assumption made at the beginning of this section, the final weighted sample \eqref{eq:Qn} obtained by KSD-UCB1-M is asymptotically unbiased as long as the weight estimates $w_i$ are asymptotically unbiased.}

Thus, we need to find some asymptotically unbiased estimates of the probabilities of the regions $A_i$.

\subsection{Weight estimation}
\label{sec:weight_estimation}

In this section we consider the problem of finding the weights $w_i$ in \eqref{eq:Qn}. As discussed after the equation, this amounts to finding the probability of the region the samples cover, which is again challenging since we have access only to an unnormalized density. This problem is faced by every algorithm which tries  to speed up  MCMC methods  by  running parallel chains. As an example, in big-data scenarios it is common to split the data into subsets,  run an MCMC sampler on each subset, and combine the resulting samples in the end. To our knowledge, most work in the literature solves this problem by estimating the density of each batch of samples separately \citep{Angelino_2016,nemeth_merging_mcmc_2018},  using typically either a Gaussian approximation \citep{scott_consensus_2016} or some kernel-density estimation method \citep{neiswanger_2014}. According to \citet{nemeth_merging_mcmc_2018}, the first approach works well in practice,  in spite of not being supported by any theory, while the second, kernel-based estimation  scales poorly with the dimension $d$. 

Here we take a  different approach and rather than estimating the density of the sample batches, we directly estimate the probabilities $P(A_i)$ via R\'enyi entropy.

Formally, suppose the domain of $P$ is partitioned into non-overlapping regions $A_1,\ldots,A_K$, and from each region $A_i$ we have a set of samples $X^{(A_i)}$. 
The R\'enyi entropy  of order $\alpha\neq 1$  for a density $p$ is defined as $R_{\alpha}(p) = \frac{1}{1 - \alpha}\log\int_{\mathbb{R}^d} p^{\alpha}(x) \text{d}x$. The conditional density of $P$ restricted to a set $A$ is denoted as $p(x|A) = \tfrac{p(x)}{P(A)}\IND{x \in A}$, and its R\'enyi entropy is 
$R_{\alpha}(p|A) = \frac{1}{1 - \alpha}\log\int_{A} \left( \frac{p(x)}{P(A)}\right)^{\alpha} \text{d}x$. 
From this definition it trivially follows that
$$ \log P(A) = R_\alpha(p|A) - \frac{1}{1 - \alpha}\log\E\left[p(X)^{\alpha-1}\big|X \in A\right].$$
In our case, instead of $p$ we only have access to $\hat{p}=c p$ for some $c>0$. Replacing $p$ with $\hat{p}$ in the integral, we obtain
\begin{equation}
\label{eq:logP-renyi}
\log (P(A) c) = R_\alpha(p|A) - \frac{1}{1 - \alpha}\log\E\left[\hat{p}(X)^{\alpha-1}\big|X \in A\right].
\end{equation}
Thus, we can estimate $P(A)$ (or, more precisely, $P(A) c$) by estimating the two terms on the right-hand side of the above equation. 

Given a sample $X^{(A)}$ taken i.i.d. from $p(\cdot|A)$, the second term can be estimated by the empirical average
$\hat{B}_\alpha(X^{(A)})=\frac{1}{|X^{(A)}|} \sum_{x \in X^{(A)}} \hat{p}(x)^{\alpha-1}$, while for the first term we can use a graph-based estimate \citep{hero_1999}. In particular, we are going to use the estimator $\hat{R}_\alpha(X^{(A)})$ of \citet{pal_renyi_2010}, which is based on generalized $k$-nearest neighbor graphs of the sample $X^{(A)}$, and it converges to $R_\alpha(p|A)$ almost surely for any $\alpha \in (0,1)$ as the sample size grows to infinity.
Thus, we obtain that $\beta(X^{(A)}) = \hat{R}_\alpha(X^{(A)}) - \frac{1}{1-\alpha} \log \hat{B}_\alpha(X^{(A)})$ is an asymptotically unbiased estimate of $\log (P(A) c)$. Therefore, given a fixed partition $A_1,\ldots,A_K$,
\begin{equation}
\textstyle P(A_i) \approx  w_i = e^{\beta(A_i)}/\sum_{j \in [K]} e^{\beta(A_j)}~.
\label{eq:wi}
\end{equation}
More precisely, if the minimum number of samples in the regions is $m$, then $P(A_i)= \lim_{m\to\infty} w_i$.

We use the following estimator $\hat{R}_\alpha(X^{(A)})$ based on $k$-nearest neighbors \citep{pal_renyi_2010}: Let $N_k(X^{(A)})$ denote that set of all pairs in $X^{(A)}$ defining the edges of its $k$-nearest neighbor graph, and define $L_{p,k}(X^{(A)})= \sum_{(x,x') \in N_k(X^{(A)})} \|x-x'\|_2^p$ for some $p \ge 0$. Then for $\alpha \in (0,1)$, the estimator is defined as
\[
\hat{R}_\alpha(X^{(A)}) =\frac{1}{1-\alpha} \log \frac{L_{p,k}(X^{(A)})}{\gamma n^{1-p/d}}
\]
where $p=d(1-\alpha)$ and $\gamma>0$ is a constant such that $\lim_{n\to\infty} L_{p,k}(Y^n)/n^{1-p/d}=\gamma$ almost surely where $Y^n$ is a set of $n$ i.i.d. samples drawn uniformly from the unit cube $[0,1]^d$. Theorem~1 of \citet{pal_renyi_2010} ensures that $\gamma$ is well-defined, while their Theorem~2 shows that if $A$ is bounded, then for any $\delta \in (0,1)$ and $p<d-1$,
\[
|\hat{R}_\alpha(X^{(A)}) - R_\alpha(p|A)| = O\left(m^{-\frac{\alpha}{d(1+\alpha)}} \left(\log\frac{1}{\delta}\right)^{\frac{1-\alpha}{2}}\right)
\]
with probability at least $1-\delta$ where $m=|X^{(A)}|$ is the number of samples in $A$. Omitting the $\delta$ terms from the notation for simplicity, we see that the 
additive error of the estimator $\hat{R}_\alpha(X^{(A)})$ is of order $m^{-\alpha/(d(1+\alpha))}$ which dominates the $O(m^{-1/2})$ error of $ \hat{B}_\alpha(X^{(A)})$ for $d \ge 2$ (which can be obtained by standard concentration inequalities if $p$ is bounded from below by some number, see, e.g., \citealp{BoLuMa13}). This implies that $|\beta(X^{(A)}) - \log (P(A) c) | = O(m^{-\alpha/(d(1+\alpha)})$. Therefore, $P(A_i)$ can be estimated with a multiplicative error of $O(\exp(const\cdot m^{-\alpha/(d(1+\alpha)})$ where $m$ now denotes the minimum number of samples over the partition cells $A_1,\ldots,A_K$. In other words, $|\log P(A_i) - \log w_i| = O(m^{-\alpha/(d(1+\alpha)})$, and using $e^x=1+O(x)$ as $x \to 0$, we also have $|P(A_i)-w_i| = O(m^{-\alpha/(d(1+\alpha)})$ as $m \to \infty$. Note that although the error of our estimator scales quite unfavorably in the dimension $d$, in practice it seems to work well even for moderately large dimensions (around 20). Furthermore, to maximize the convergence rate, in the experiments we choose $\alpha$ to be close to one (this also controls the variance of $\hat{B}_\alpha(X^{(A)})$, which was hidden in the above crude analysis); see \Cref{sec:weight} for more details.

\begin{remark}\em
Another intuitive way to estimate the constant $c=\hat{p}/p$ is to use importance weighting. Having a distribution $Q$ with density $q$, we have
\[
c P(A) = \int_A \hat{p}(x)\; dx = \int_A \frac{\hat{p}(y)}{q(y)/Q(A)}\cdot  \frac{q(y)}{Q(A)}\; dx~.
\]
Using this formulation, it is possible to estimate $c P(A)$ by taking samples $y_1,\ldots,y_n$ from $A$ independently according to $Q$ restricted to $A$, then compute
$\frac{Q(A)}{n} \sum_{i=1}^n \frac{\hat{p}(y_i)}{q(y_i)} \approx c P(A)$.
Finally, $Q(A)$ can be estimated, for example, by taking $m$ samples $y'_1,\ldots,y'_m$ independently from $Q$, and calculating the empirical frequency $\frac{1}{m}\sum_{i=1}^m  \IND{y'_i \in A} \approx Q(A)$. Note, however, that the latter might be problematic in general (especially in high dimensional spaces), and the method only works if $Q$ is well-aligned with $P$. We have run multiple experiments where the target distribution $P$ was a Gaussian mixture distribution, and for each set $A$ (which essentially contained a number of the mixture components), we chose $Q$ to be a Gaussian distribution with mean and variance set as the empirical mean and empirical variance of a sample taken independently from $P$ restricted to $A$. The experiments with different randomly selected target distributions consistently showed that the performance of the two methods were essentially the same in the nice cases when the number of partition cells was equal to the number of connected regions (each region may contained more than one mixture components which were close to each other), while in every other case the  R\'enyi-entropy-based estimation method outperformed the one based on importance sampling. Hence, in the rest of the paper we only consider the former.
\end{remark}

\section{The final algorithm}
\label{sec:finalalg}

So far we have discussed how to solve our problem if we have either local mixing or all the samplers mix globally. While the latter is the case asymptotically for all the MCMC samplers used in practice, the mixing may be too slow to be observed for any practical number of samples. On the other hand, the problem does not simplify to the local mixing scenario, since--even if well-separated regions are actually present--the chains often jump from one region to another even if they do not cover the whole domain.

To be able to adapt our KSD-UCB1-M algorithm, we need to group together the samplers covering the same region (even though what a region is is not clearly defined in this scenario). The problem is especially hard since the grouping of the samplers is non-stationary, and we should also be able to track when a sampler leaves or joins a region (equivalently, group). Furthermore, if the groups are too large, we do not explore the whole domain, while if they are too small, we waste resources by running multiple samplers for the same region.

To solve this issue, we propose a simple heuristic to  identify samplers that are close together:  In each round of the sampling, we take all the samples from the last batch of each sampler, and for each sample point we look at its $N$ nearest neighbors. Then we find the grouping where any two samplers are grouped together that have points which are nearest neighbors of each other (this can easily be done recursively). By this simple heuristic, in each round we can group together the samplers that are close to each other. Note that here we do not make any assumption regarding the number of the regions (e.g., well-separated modes) of the distribution. By having all the samples, the algorithm can easily identify multiple regions  by running a clustering algorithm. 
The final step of the algorithm is to determine the correct weight for each sample point. Recall that in the locally mixing case we weighted the empirical distribution of each region by their estimated probability (cf. Eq.~\ref{eq:Qn}). Here, since we do not have these regions, in the end we assign the samples into $M$ clusters using $k$-means clustering, and weight the empirical measure within each cluster by the estimated probability of the cluster (using Eq.~\ref{eq:wi}).
Algorithm~\ref{algorithm:mcmc_bandit_global} shows the whole procedure, called KSD-MCMC with reweighting (KSD-MCMC-WR). 
\begin{algorithm}[t]
	\small
	\caption{KSD-MCMC-WR} \label{algorithm:mcmc_bandit_global}
	\begin{algorithmic}
		\STATE {\bfseries Given:} Distribution $p(x)$; $M$ samplers; total number of rounds $T$; batch size $n_b$; number $N$ of nearest neighbors for  clustering; the order $\alpha$ of the R\'enyi entropy.
		\STATE {\bfseries Initialize:} For each $i \in [M]$, draw $n_b$ samples from the $i$th sampler with random initialization; compute $S_{i,1}$ and set $\bar{\mu}_{i,1} = S_{i,1}$ and $T_i(1) = 1$.
		\FOR {$t \in \{M+1, \ldots, T\}$}
		\STATE - Cluster the samplers by clustering the samples from their last batches:
		\bindent
		\STATE - Initially, the last batch of samples from each sampler forms a cluster.
		\STATE - Merge two clusters if any point of one cluster has a point from the other cluster among its $N$ nearest neighbors.\!\!\!\!
		\STATE - Find the number of clusters $n_c$.
		\STATE - Define $\A_i \subset [M]$ for $i \in [n_c]$ as the set of samplers belonging to cluster $i$ ($\A_i\cap \A_j = \emptyset$ for $i\neq j$).\!\!\!\!
		\eindent
		\STATE - Choose a cluster $I_t$ (e.g., uniformly at random).
		\STATE - Draw a batch of samples of size $n_b$ from sampler $i_t = \arg\min_{i \in A_{I_t}} \left( \bar{\mu}_{i, T_i(t-1)} - \sqrt{\frac{2 \log t}{T_i(t-1)}}\ \right)$. 
		\STATE - Set $T_{i_t}(t)=T_{i_t}(t-1)+1$ and $T_j(t)=T_j(t-1)$ for $j\neq i_t$.
		\STATE - Observe $S_{i_t,T_{i_t}(t)}$, and compute $\bar{\mu}_{i_t,T_{i_t}(t)}=(1-\frac{1}{T_{i_t}(t)})\bar{\mu}_{i,T_{i_t}(t-1)}+\frac{S_{i_t,T_{i_t}(t)}}{T_{i_t}(t)}$.
		\ENDFOR
		\STATE - Cluster all the samples into $M$ clusters by k-means clustering; for each cluster calculate its estimated probability $w_i$ using \eqref{eq:wi} and output the reweighted samples with weight $w_i/n_i$ where $n_i$ is the total number of samples in cluster $i \in [M]$.
	\end{algorithmic}
\end{algorithm}

\section{Experiments}
\label{sec:exp}

In this section we empirically evaluate  our choices in the algorithm design process as well as compare the performance of our final algorithm with state-of-the-art methods from the literature on several synthetic problems.

We use three different base sampling methods: Metropolis-Hastings (MH), 
the Metropolis-adjusted Langevin algorithm (MALA) of \citet{mala_2002} and the no-U-turn sampler (NUTS) of \citet{hoffman_no-u-turn_2014}. For the latter, which is a state-of-the-art sampling method \citep{wang_adaptive_2013,liu_black-box_2016}, we used the implementation provided in the pymc3 package \citep{pymc3} which, on top of the original NUTS, also includes several modifications recommended in the literature for boosting its performance (see the  pymc3 package website for more details). Beside the initial point, MH and MALA have a single step size parameter (for MALA the preconditioning matrix is always the identity), while for NUTS we used the default setting (in our experiments the performance of NUTS was insensitive to the choice of its parameters). In our experiments, when a MH or MALA sampler is chosen randomly, the step size is selected uniformly at random from $[0.1,5]$ (the initial points are selected uniformly at random ``not too far'' from the modes of the underlying distribution).

We present one set of experiments for each component of our algorithm. Throughout this section, to compute the KSD measure, we use the inverse multiquadratic kernel $k(x,y)=(1+\|x-y\|_2^2/h)^\gamma$ suggested by \citet{gorham_measuring_2017} with $\gamma=-0.5$ and $h=1$ unless otherwise stated (experiments in  \Cref{sec:exp_block_error} and \ref{sec:parallel} indicate that the performance of our method is quite insensitive to the choice of $h$ in the range $[0.01, 100]$).

\Cref{sec:exp_block_error} analyzes the effect of using the block-diagonal approximation of the KSD measure. The performance of our bandit-based samplers for unimodal target distributions is considered in \Cref{sec:exp_singlemode_multisampler}.
Multimodal densities with separated modes and one sampler in each mode are considered in \Cref{sec:exp_multimode_single}, including an empirical analysis of our weight estimation method in \Cref{sec:weight}.
Our final algorithm KSD-MCMC-WR is tested in the general setting of a multimodal target distribution with an unknown number of modes
in \Cref{sec:exp_general}, and specifically against parallel MCMC methods in \Cref{sec:parallel}. Finally, we consider a realistic task of node localization in a sensor network in \Cref{sec:sensor}.

\subsection{Block-diagonal approximation of the KSD measure}
\label{sec:exp_block_error}
The intuition behind our bandit MCMC  method is that we can identify the best sampler among a group of samplers  from the average KSD for small batch sizes instead of making the decision based on the KSD computed over a large set of  samples. To verify this hypothesis, we checked if the average approximate KSD computed on small blocks preserves the order of two samplers as defined by the true KSD measure. In particular, we compared the average block-KSD measure of two samplers for different block sizes. We drew $n=100000$ samples from each sampler for a non-isotropic two-dimensional independent Gaussian distribution and computed the average block-KSD measure for these samples with block sizes $10, 25, 50, 100, 250, 500, 2000$ (the parameters of the samplers and the distribution were selected randomly\footnote{Throughout this section the variance of the components of a Gaussian distribution are selected uniformly from $[0.1,3]$}). 
Formally, for each sampler $i$ with block size $n_b$ and total sample size $n$, we computed the average block-KSD as $\bar{S}_{i,n_b} = \frac{n_b}{n} \sum_{b=1}^{n/n_b} S_{i,b}$, where $S_{i,b}$ is the average block-KSD computed on the $b$th block of data, and for a given block size $n_b$, $\argmax_i \bar{S}_{i,n_b}$ is declared to be the better sampler according to the measurements. 

\begin{figure}[th]
\vspace{-0.1cm}
	\begin{center}
		\centerline{
			\includegraphics[width=0.8\columnwidth]{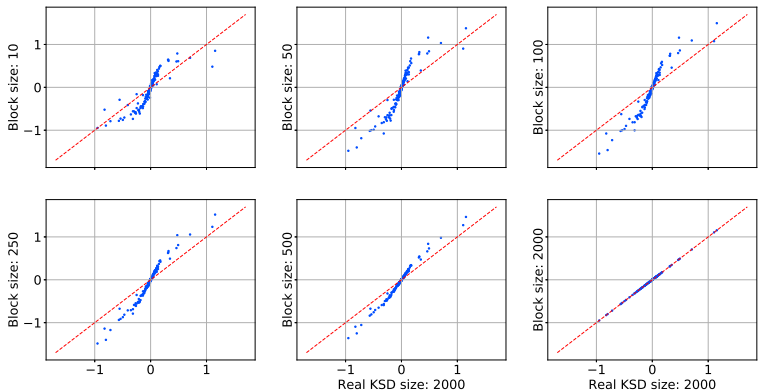} 
		}
		\caption{Block-diagonal approximation of the KSD measure for kernel width $h = 1$.}
		\label{fig:ksd_vs_blk}
	\end{center}
	\vspace{-0.5cm}
\end{figure}

\Cref{table:ksd_blk} shows the fraction of times the average block-KSD for different block sizes gave the same ordering of the samplers as the ordering obtained for block size $2000$ (which is treated as the ground truth in this experiment), both for MH and MALA as base samplers. We can observe that the ordering obtained from the average block-KSD measures is most of the time close to the ``ground truth'', justifying its use for measuring sample quality. The agreement in most cases is around at least $90\%$, and rearly gets below $80\%$. Interestingly, no trend regarding the kernel width $h$ is observable. Scatter plots of average block-KSD differences ($\bar{S}_{1,n_b}-\bar{S}_{2,n_b}$) for $h=1$ and different block sizes are given in \Cref{fig:ksd_vs_blk} for the MH algorithm. These figures show that in cases of incorrect ordering, the average block-KSD values of the samplers are really close for both block sizes, which means the two samplers considered are of approximately the same quality according to both measures (thus, practically it does not matter which of them is used for sampling). 

\begin{table*}[!ht]
	\small
	\centering
	\begin{tabular}{|l|l|l|l|l|l|l|l|l|}
		\cline{1-9}
		Kernel  width &Batch Size &  10 & 25 & 50 & 100 & 250 & 500 & 2000  \\ \cline{1-9}
		$h = 0.01$ &MH & $80\%$ &$84\%$ &$86 \%$ &$94 \%$ &$96 \%$ &$96 \%$ &$100\%$ \\ 
		$h = 0.01$ &MALA & $100\% $ &$100 \%$ &$100 \%$ &$100\%$ &$ 100 \%$ &$100 \%$ &$100\%$\\ \cline{1-9}
		$h =0.1$ &MH & $58\%$ &$68\%$ &$76 \%$ &$84 \%$ &$90 \%$ &$92 \%$ &$100\%$ \\ 
		$h =0.1$ &MALA & $86\% $ &$94 \%$ &$94 \%$ &$96\%$ &$ 98 \%$ &$98 \%$ &$100\%$\\ \cline{1-9}
		$h =1$ &MH & $93\%$ &$95\%$ &$94 \%$ &$94 \%$ &$94 \%$ &$96 \%$ &$100\%$ \\ 
		$h =1$ &MALA & $93\% $ &$92 \%$ &$93 \%$ &$91\%$ &$ 93 \%$ &$94 \%$ &$100\%$\\ \cline{1-9}
		$h =10$ &MH & $92\%$ &$92\%$ &$92 \%$ &$92 \%$ &$94 \%$ &$96 \%$ &$100\%$ \\ 
		$h =10$ &MALA & $68\% $ &$68 \%$ &$76 \%$ &$84\%$ &$ 84 \%$ &$84 \%$ &$100\%$\\ \cline{1-9}
		$h =100$ &MH & $83\%$ &$85\%$ &$85 \%$ &$85 \%$ &$85 \%$ &$89 \%$ &$100\%$ \\ 
		$h =100$ &MALA & $77\% $ &$79 \%$ &$83 \%$ &$85\%$ &$ 81 \%$ &$81 \%$ &$100\%$\\ \cline{1-9}
	\end{tabular}
	\caption{The fraction of times the samplers are ranked in the same order for different block sizes as for block size $2000$ (regarded as the ``ground truth'') for different kernel widths $h$.	}
	\label{table:ksd_blk}
\end{table*}

\subsection{Unimodal target distributions with multiple samplers }
\label{sec:exp_singlemode_multisampler}
In this section we consider sampling from a standard normal Gaussian distribution, where the samplers have access to the unnormalized density $p(x) = e^{-x^\top x/2}$.
The goal of the samplers is to estimate the mean of the distribution with respect to the mean squared error. We consider $5$ different MH and, respectively, MALA samplers (with step size parameters $0.1, 0.2, 0.5, 1, 2$, resp.) and run our  proposed algorithm KSD-UCB1  (\Cref{algorithm:ucb}) on top of them. 
The method is compared with the following baselines and variations:

\begin{itemize}
\item Uniform: This algorithm distributes the computational budget equally among the samplers, and in the end takes all the samples generated from all samplers with equal weights. %
\item KSD-opt: This method again takes the same number of samples from each sampler (i.e., uses the aforementioned uniform algorithm), but reweights them using the method of  \citet{liu_black-box_2016}. Indeed, KSD-opt is a reweighting method that can be used with any sets of samples. However, it has been originally proposed to be used on top of the uniform algorithm. This reweighting can be very effective for  unimodal distributions or multimodal distributions with really close modes, but it is computationally quite demanding.
\item $\eps$-greedy: This method is a variant of KSD-UCB1, but is based on the $\eps$-greedy bandit algorithm \citep{bubeck_regret_2012} instead of UCB1. That is, in every round, with probability $1-\eps$, the sampler with the smallest approximate block-KSD measure is selected, while with probability $\eps$, a sampler is selected uniformly at random. In the experiments $\eps=\frac{0.05}{\sqrt{t}}$ in round $t$.
\end{itemize}

\begin{wrapfigure}[15]{r}{0.45\textwidth}
	\vspace{-0.5cm}
	\begin{center}
		\centerline{
			\includegraphics[width=0.4\columnwidth]{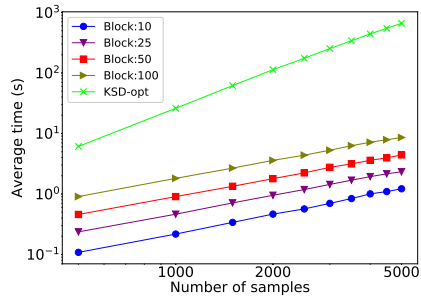} 
		}
		\caption{The average running time for different KSD based method.}
		\label{fig:time}
	\end{center}
\end{wrapfigure}
\begin{figure}[t]
	\begin{center}
		\centerline{
			\includegraphics[width=0.5\columnwidth]{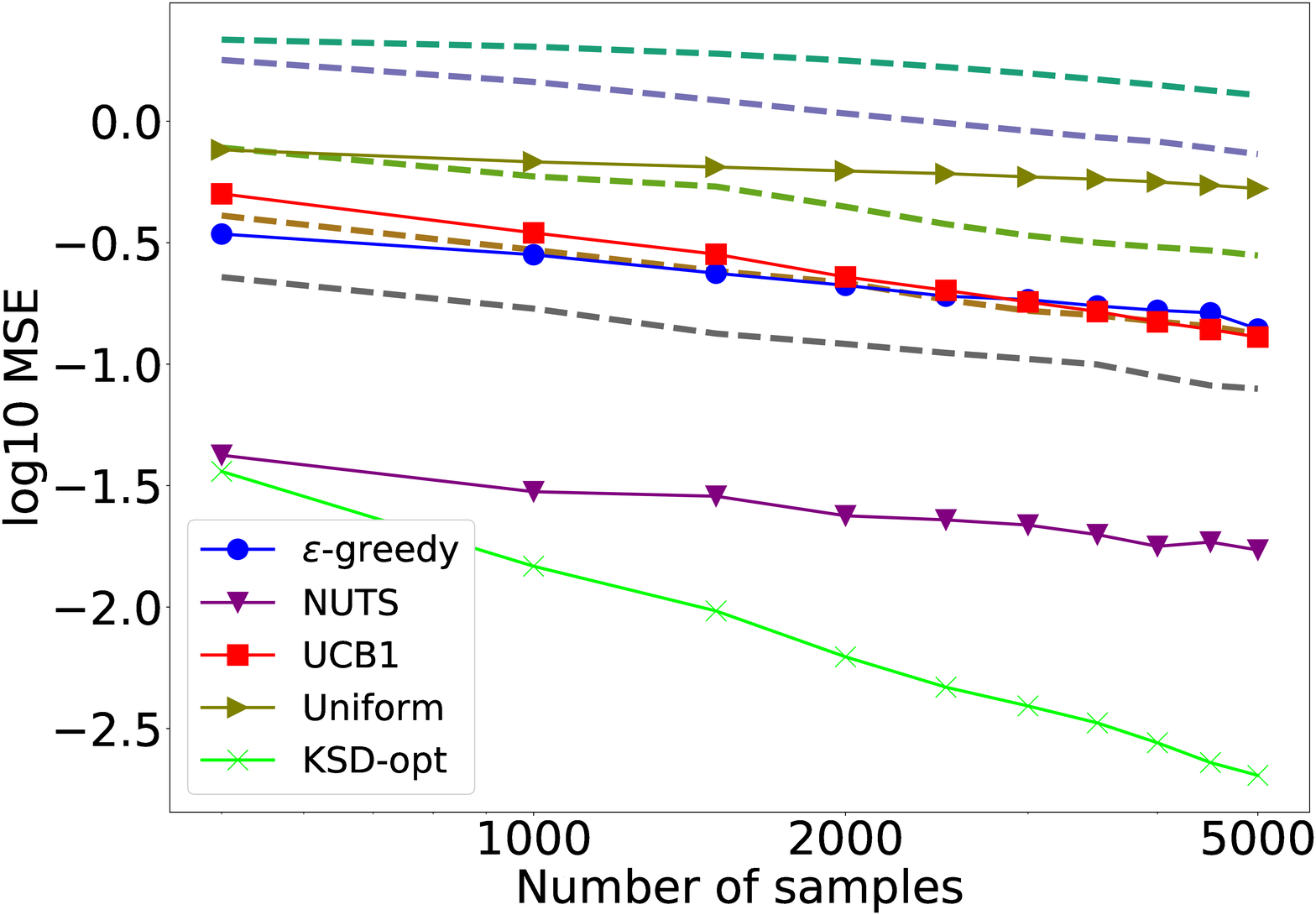} 
			\includegraphics[width=0.5\columnwidth]{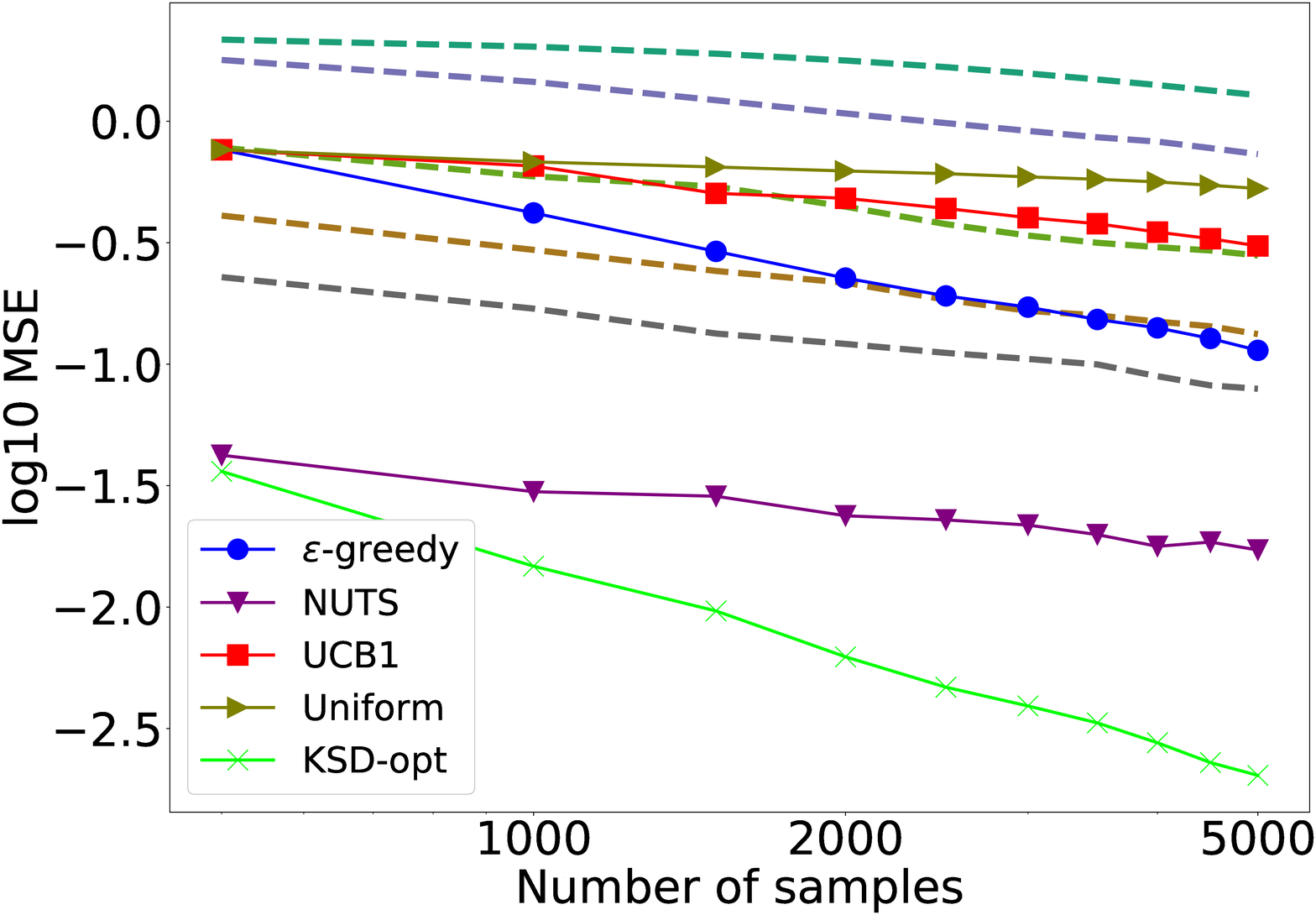}}
		\caption{Unimodal case, Gaussian target distribution: MSE for different sample sizes with MH samplers in $2$-dimensions ($d=2$) with batch size $10$ (left) and batch size $100$ (right). The dashed lines without labels show the performance of the different MH samplers. The total number of samples is $5000$ in each case. }
		\label{fig:singlemode_multiple_chain}
		\centerline{
			\includegraphics[width=0.5\columnwidth]{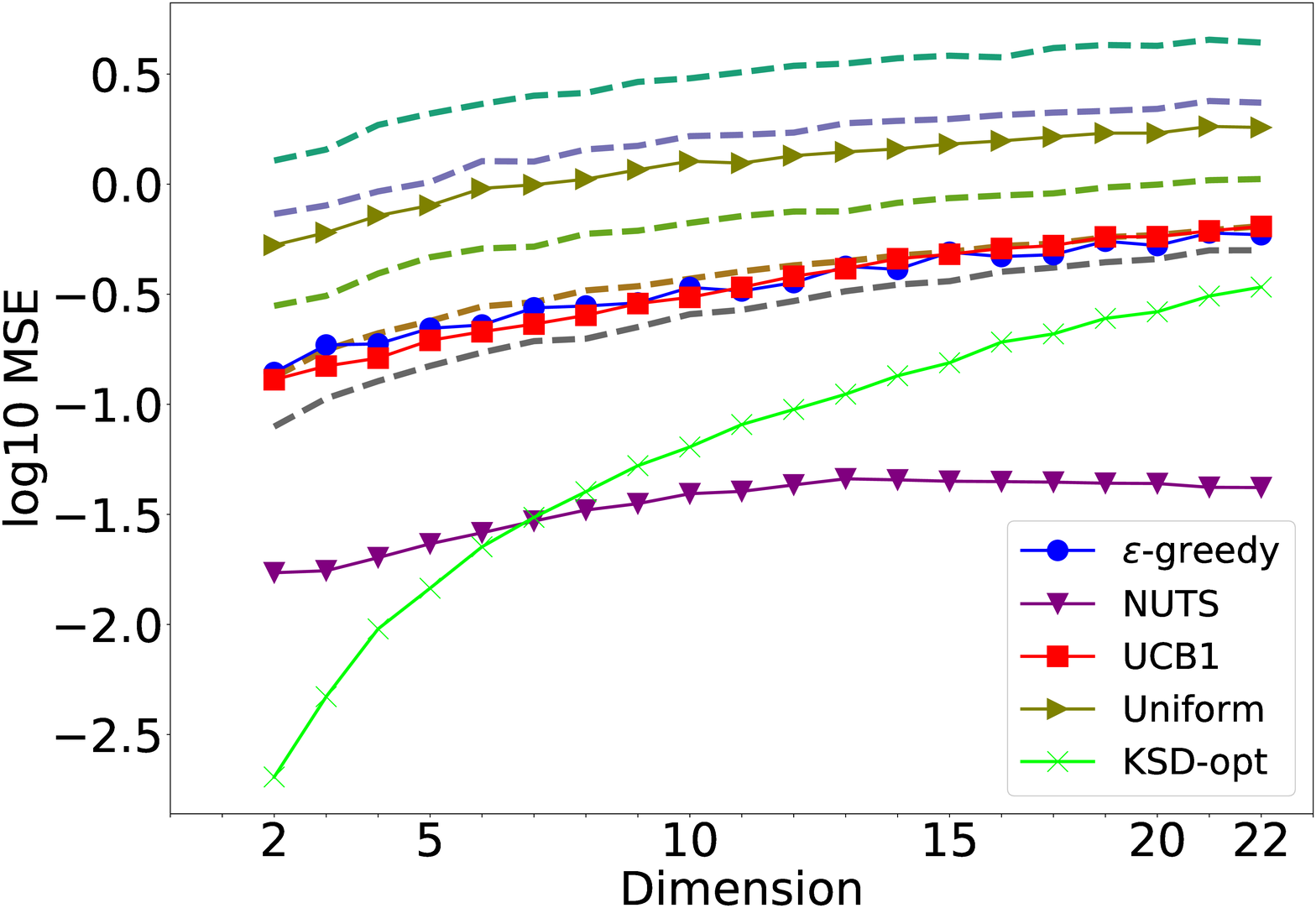} 
			\includegraphics[width=0.5\columnwidth]{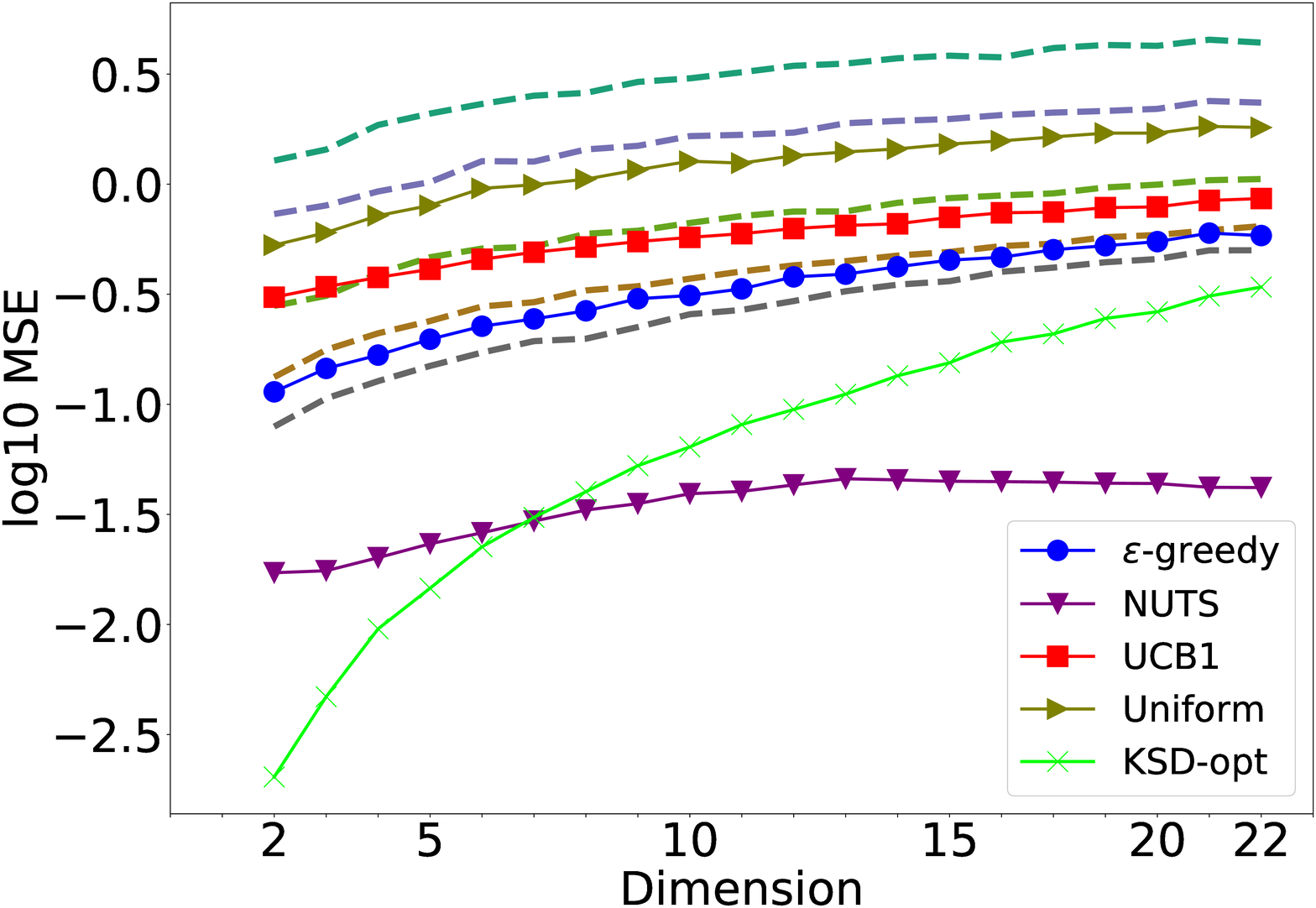}}
		\caption{Unimodal case, Gaussian target distribution: MSE versus dimension $d$ with MH samplers and batch size $10$(left) and batch size $100$ (right). The dashed lines without labels show the performance of the different MH samplers. The total number of samples is $5000$ in each case.}
		\label{fig:singlemode_e_vs_d}
	\end{center}
\end{figure}

We also compared the performance with that of NUTS. Figures~\ref{fig:singlemode_multiple_chain} and~\ref{fig:singlemode_e_vs_d} show the results for MH base samplers with $d=2$ and as a function of $d$, respectively.  One can observe that both bandit-based algorithms (i.e., KSD-UCB1 and KSD-$\eps$-greedy) perform similarly, almost achieving the performance of the best base sampler. Interestingly, there is no significant difference between the two methods, and $\eps$-greedy, which is an inferior bandit algorithm, seems to perform slightly better. This might indicate that in our case less exploration could be preferable. Also, it is interesting to note that the bandit method with the smaller batch size of $10$ outperformed the one with bigger batch size of $100$. Since we have already observed that the ordering of the samplers is not really affected by the batch size, this improved performance is most likely due to the increased number of decision points, which allows better adaptation.
Note that given that the computational complexity of calculating the KSD measure is quadratic in the number of samples, smaller batch sizes have a huge advantage compared to bigger ones. In particular, in our experiment, the computational cost for batch size $10$ is about two orders of magnitude smaller than for batch size $100$. The running time of the different KSD computations are shown in \Cref{fig:time}.

\begin{figure}[t]
	\begin{center}
		\centerline{
			\includegraphics[width=0.5\columnwidth]{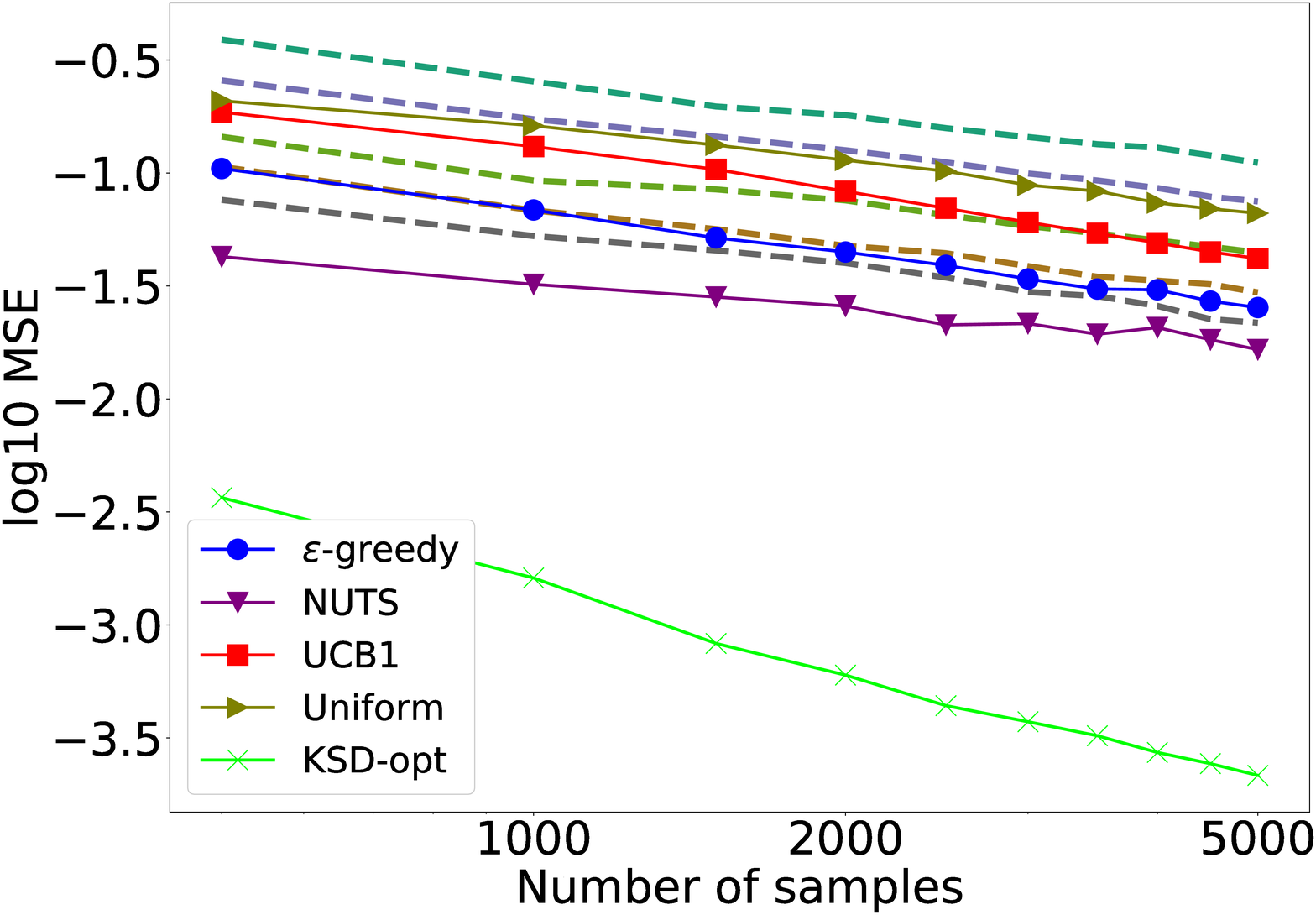} 
			\includegraphics[width=0.5\columnwidth]{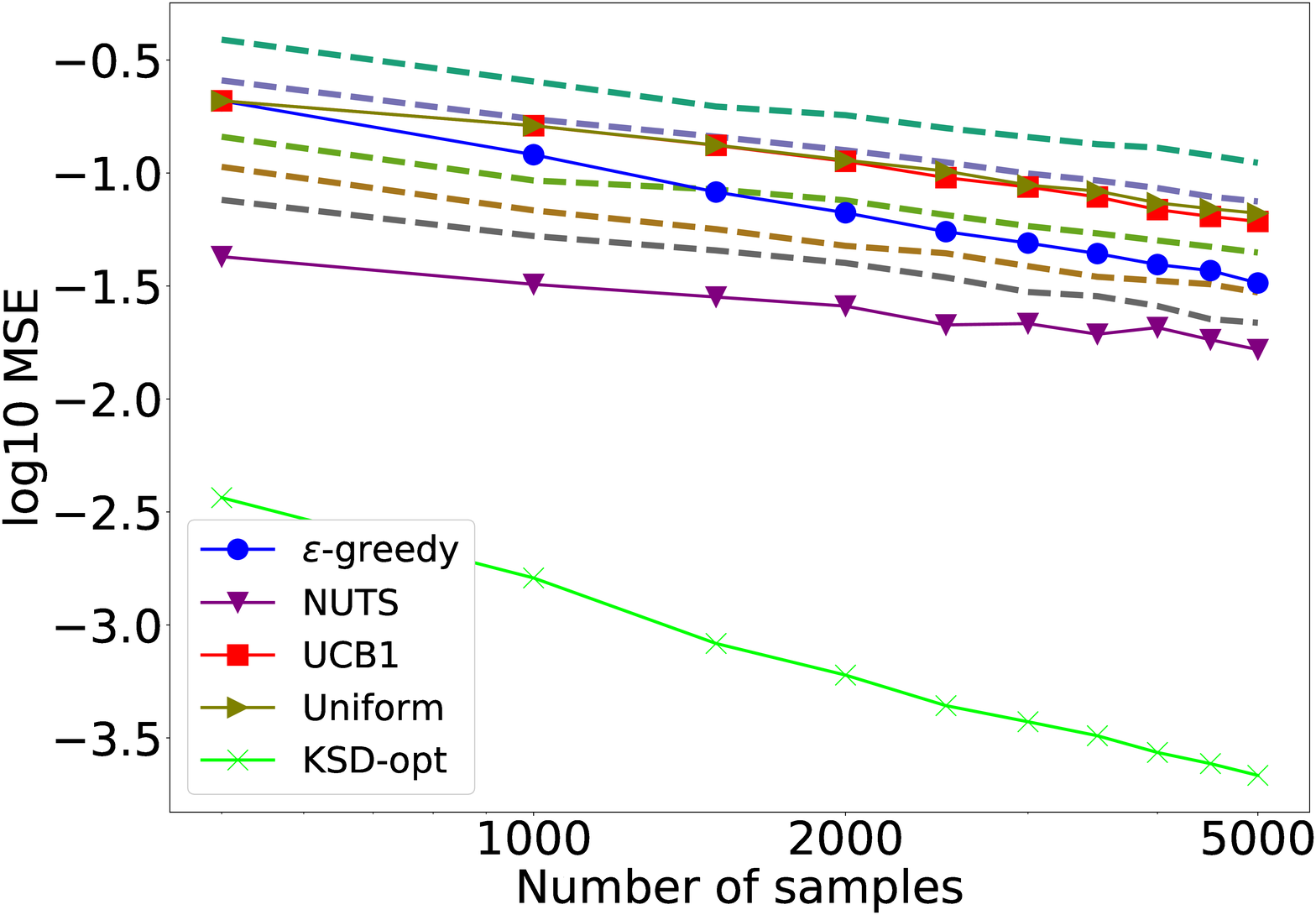}}
		\caption{Unimodal case, Gaussian target distribution: MSE for different sample sizes with MALA samplers in $2$-dimensions ($d=2$) with batch size $10$ (left) and batch size $100$ (right). The dashed lines without labels show the performance of the different MALA samplers. The total number of samples is $5000$ in each case.}
		\label{fig:singlemode_multiple_chain_mala}
		\centerline{
			\includegraphics[width=0.5\columnwidth]{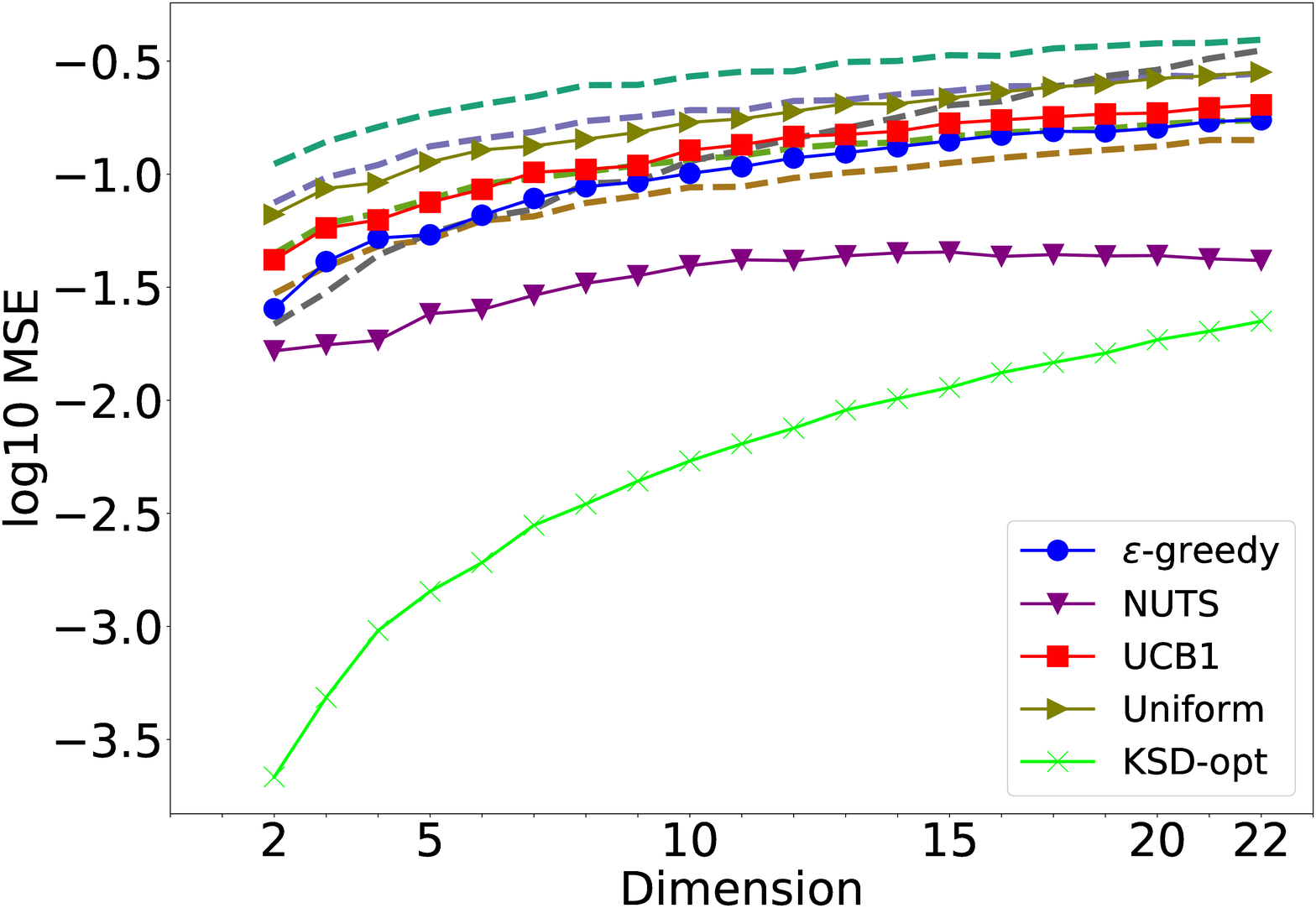} 
			\includegraphics[width=0.5\columnwidth]{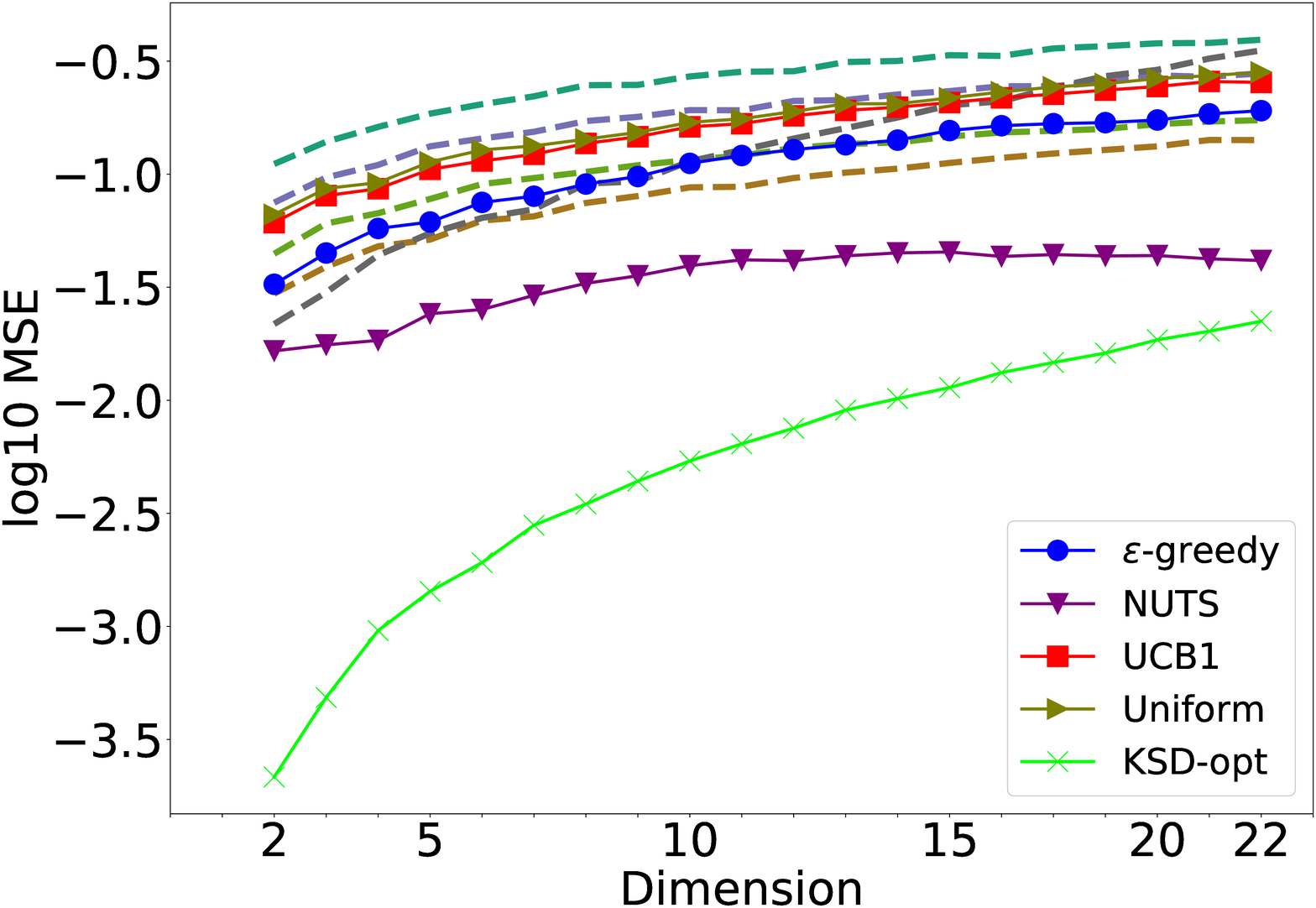}}
		\caption{Unimodal case, Gaussian target distribution: MSE versus dimension $d$ with MALA samplers and batch size $10$ (left) and batch size $100$ (right). The dashed lines without labels show the performance of the different MALA samplers. The total number of samples is $5000$ in each case.}
		\label{fig:singlemode_e_vs_d_mala}
	\end{center}
\end{figure}

On the other hand, although our methods are competitive with the MH samplers, one can observe that NUTS significantly outperforms all of them, hence also their combinations. Furthermore, the reweighting mechanism of KSD-opt can provide a significant boost in performance for lower dimensions.

The performance of the aggregating sampling algorithm can be significantly improved by changing the base samplers. This is shown in Figures~\ref{fig:singlemode_multiple_chain_mala} and~\ref{fig:singlemode_e_vs_d_mala}, where MH is replaced with the much better MALA samplers, making the best sampler and our bandit-based sampling method competitive with NUTS. Interestingly, due to the better quality samples, in this case KSD-opt outperforms MALA for the whole range of dimensions (up to $d=22$) we consider, although the performance difference vanishes as $d$ increases, and indicates (as expected) that NUTS will be better for large values of $d$.

\subsection{Separated modes with one sampler for each mode}
\label{sec:exp_multimode_single}
In this section we consider the case of a multimodal target density (Gauss mixture) with separated modes, where we have one sampler for each mode. 
Here we run several versions of KSD-UCB1-M, which differ in how the region (mode) $I_t$ is selected in the algorithm. In particular, we consider the following methods for selecting $I_t$:
\begin{itemize}
\item Uniformly at random (this is called "Equal probability" in the figures).
\item Random, proportionally to the estimated weights $w_i$ (or, equivalently, the estimated probability of the corresponding region), as described in \Cref{sec:weight_estimation} ($w$).
In all the experiments, we used the  Information Theoretical Estimators package of \citet{szabo14information} to calculate estimates of the R\'enyi entropy.
\item $I_t$ is selected as the sampler (region) with the largest average block-KSD measure (KSD).
\item $I_t$ is selected randomly with probability proportional to $w_i \tilde{S}_{i,n_i}$, as suggested by \eqref{eq:tSbound} (KSD.$w$).
\item $I_t$ based on the stratified sampling idea of \citep{carpentier_adaptive_2015}, with probability proportional to $\hat{\sigma}_i w_i$, where $w_i$ is the estimated weight and $\hat{\sigma}_i$ is the estimated standard deviation of the samples ($\sigma.w$)
\end{itemize} 
For completeness, for the last reweighting step we conducted experiments with both the true weights and also with estimated weights.

\begin{figure}[t]
	\begin{center}
		\centerline{
			\includegraphics[width=0.5\columnwidth]{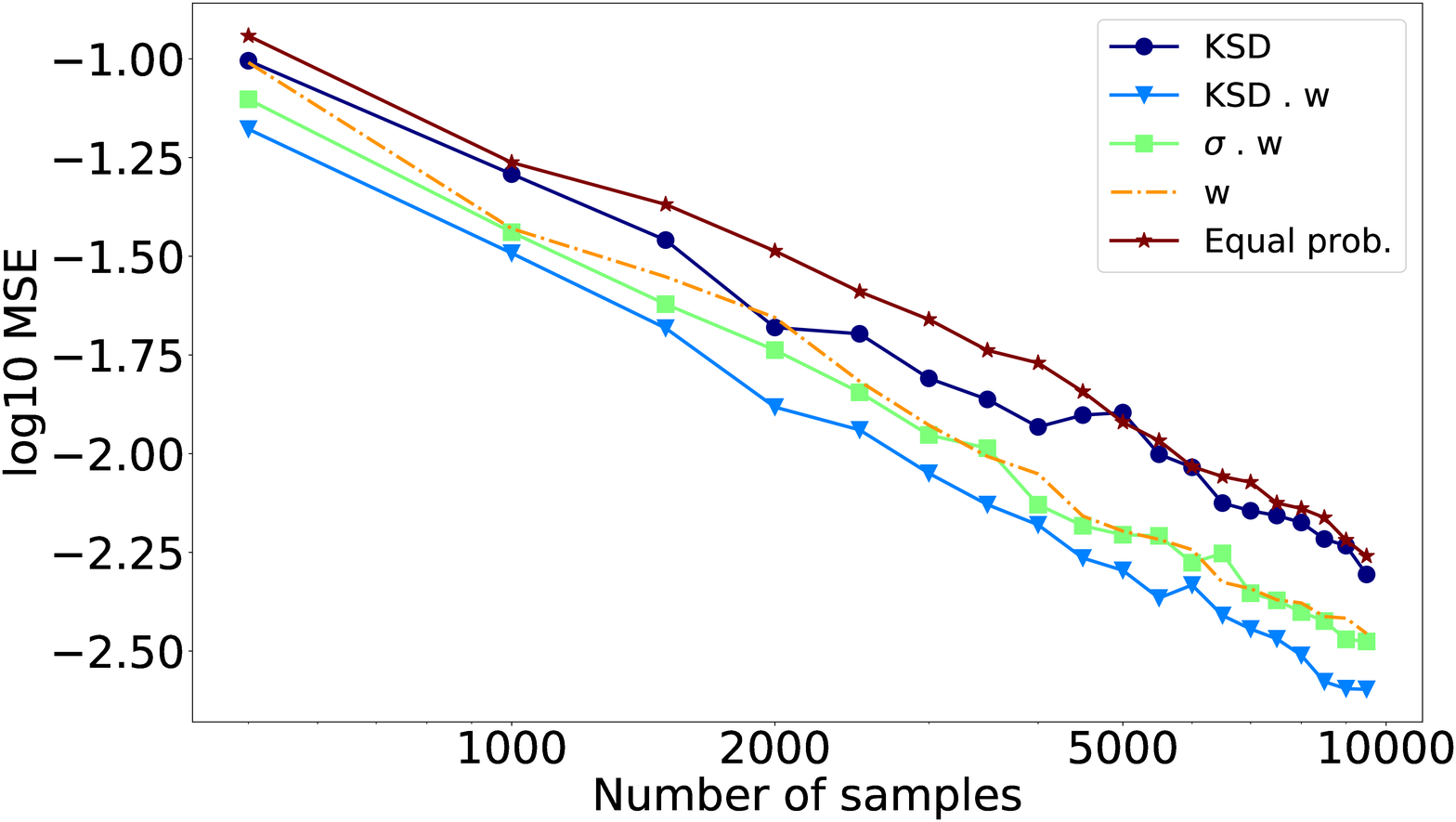} 
			\includegraphics[width=0.5\columnwidth]{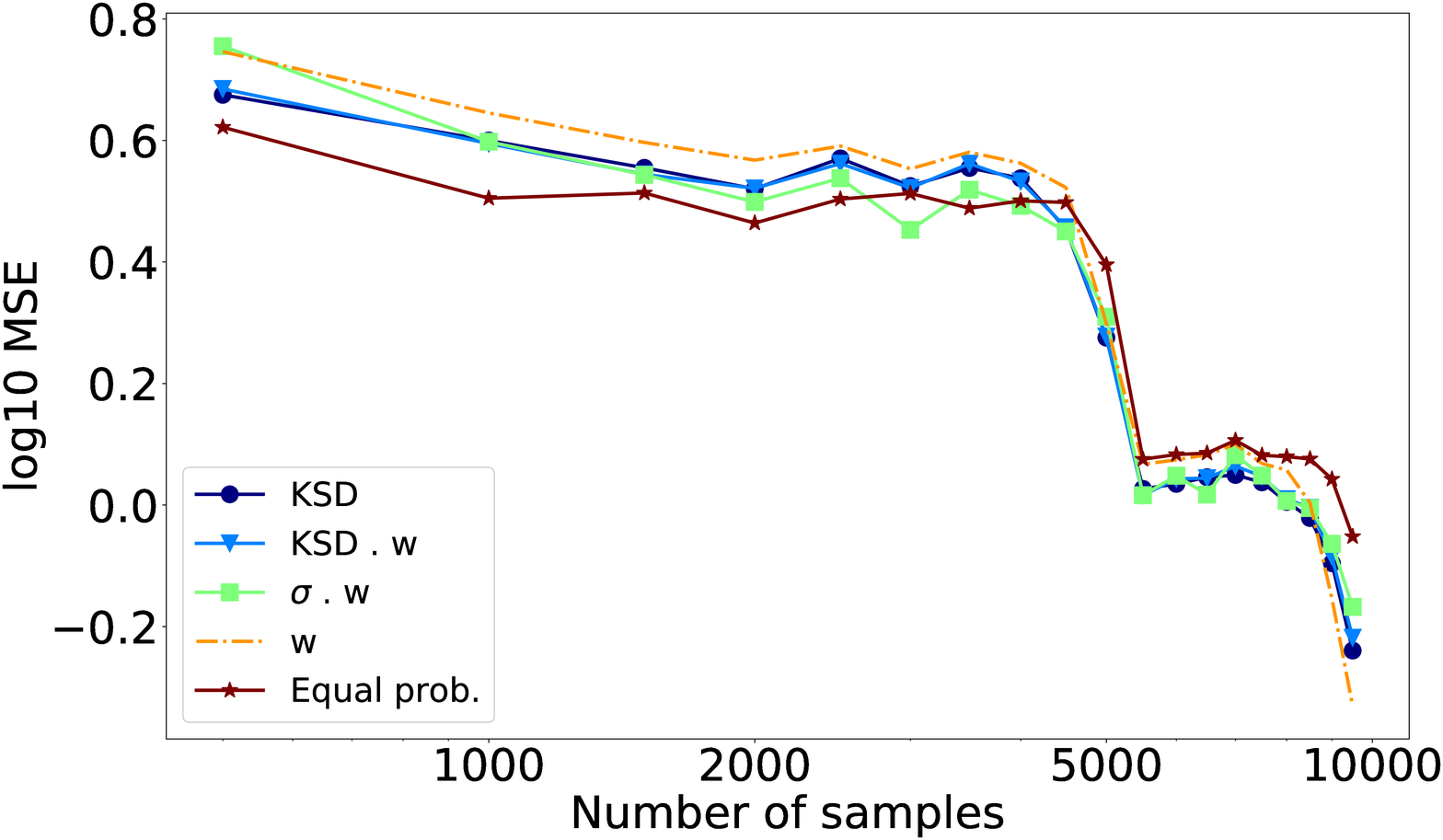}}
		\caption{Multimodal distributions (Gauss mixtures): MSE for different sample sizes when the samplers are MH when the weight of each mode is know in advance (left); and when the weights are estimated with our proposed weight estimation method  based on R\'enyi entropy with $\alpha=0.95$ (right). The samples are taken from the following   unnormalized distribution $p(x) = \sum_{i=1}^{3}\beta_i \exp\left\lbrace (x-\mu_i)^\top \Sigma_i^{-1}(x-\mu_i)\right\rbrace$ where  $\beta= [0.5, 0.3, 0.2]$, $\Sigma_i = \sigma_iI$ with $\sigma=[0.9, 0.4, 0.5]$ and $\mu_1 = [6, 6]^\top$, $\mu_2 = [-6, 6]$ and $\mu_3 = [0, -6]$.  On the right figure one can observe that the error slightly goes up after a sharp decline for all the methods (after about $5000$ samples). This is due to a single chain that visited some areas of low probability that affected the weight estimation. The error starts to decline again when the chain returns to a high probability area. }
		\label{fig:mse_vs_sample_multimode_mh}
	\end{center}
\end{figure}

\begin{figure}[t]
	\begin{center}
		\centerline{
			\includegraphics[width=0.5\columnwidth]{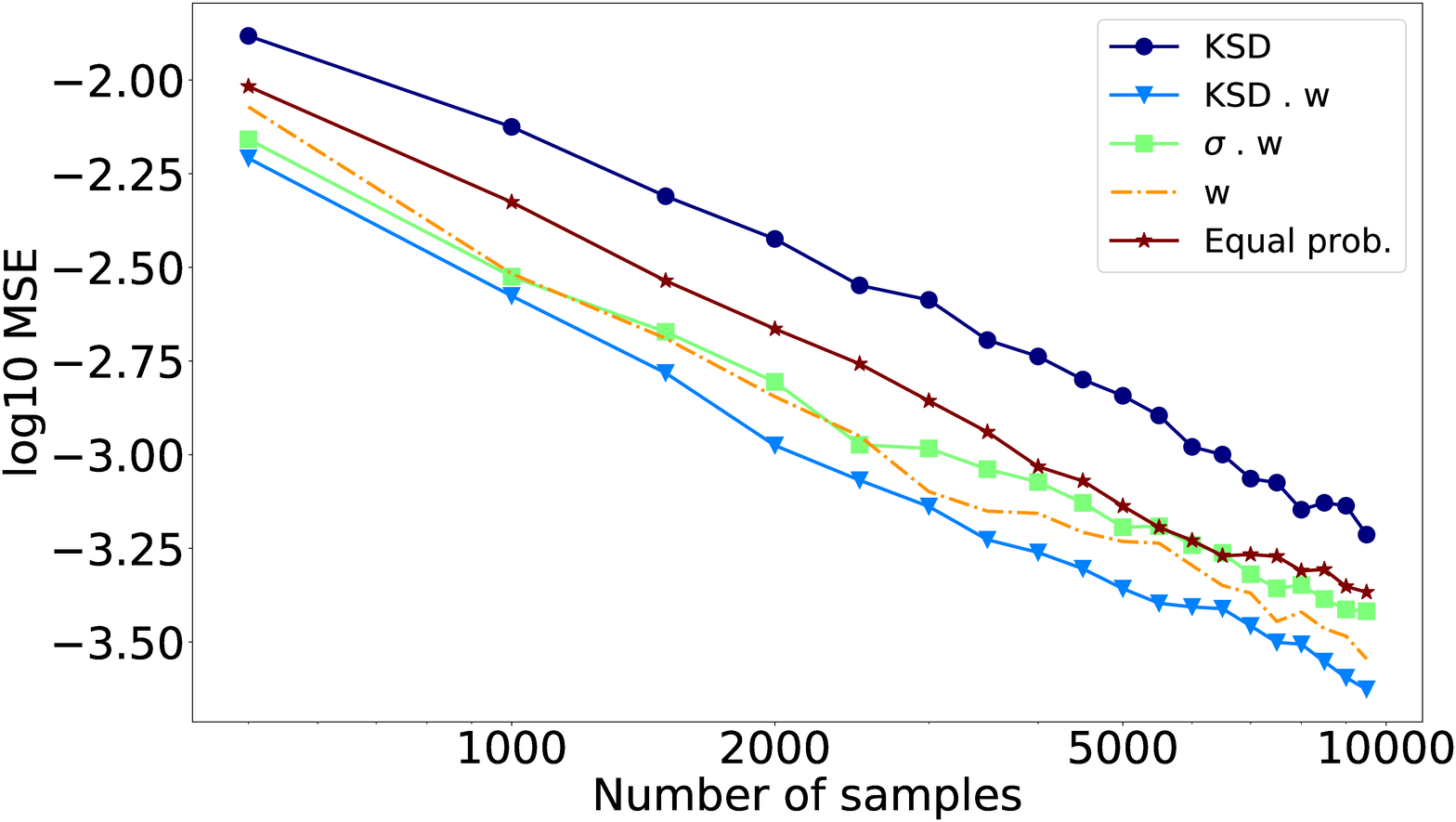} 
			\includegraphics[width=0.5\columnwidth]{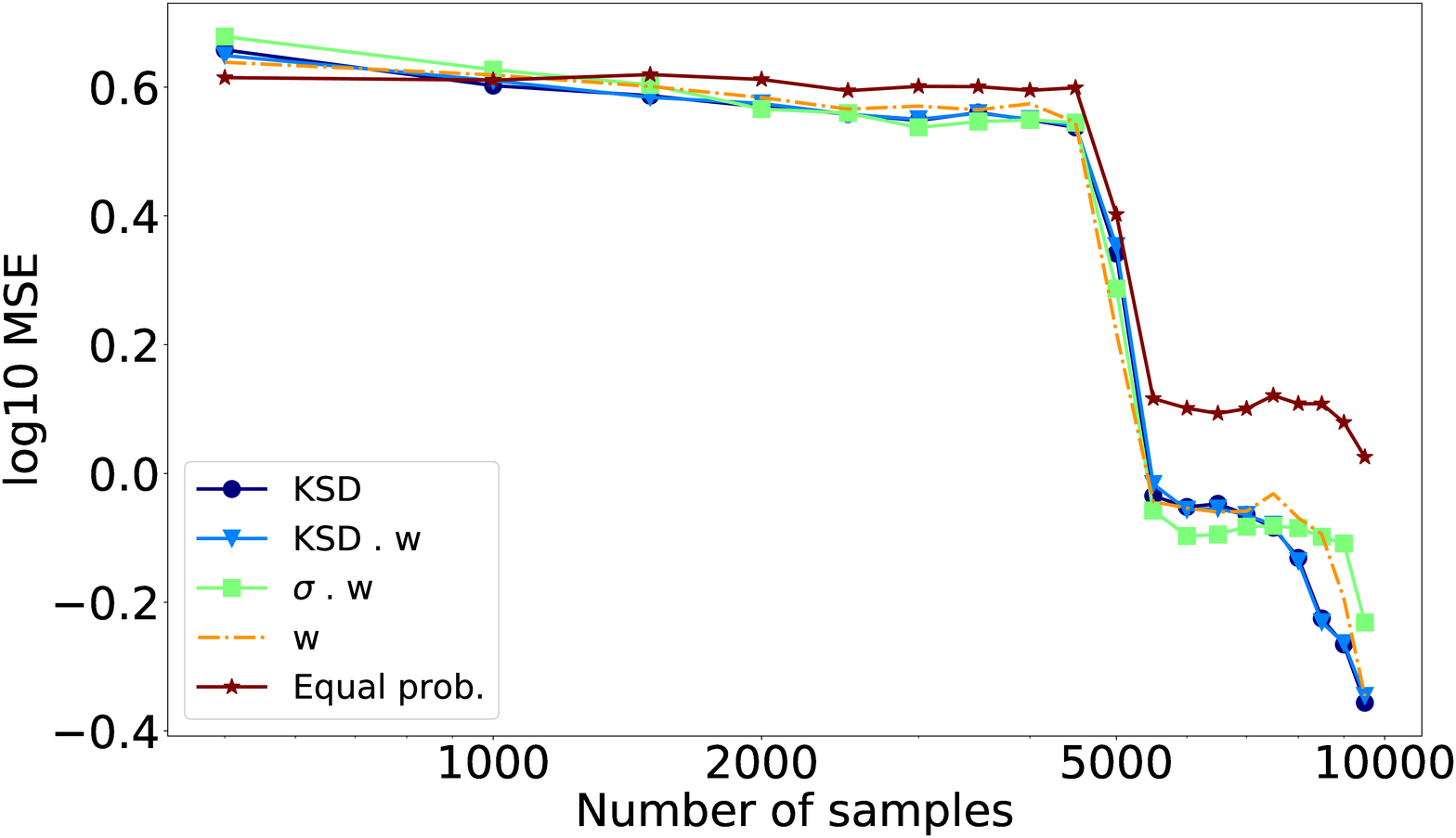}}
		\caption{Multimodal distributions (Gauss mixtures): MSE for different sample sizes for the same target distribution as in \Cref{fig:mse_vs_sample_multimode_mh} with MALA base samplers when the weight of each mode is know in advance (left); and when the weights are estimated with our proposed weight estimation method  based on R\'enyi entropy with $\alpha=0.95$ (right). %
	}
		\label{fig:mse_vs_sample_multimode_mala}
	\end{center}
\end{figure}

We ran experiments for a target distribution with three separated modes, where the goal again was to estimate the mean of the distribution with respect to the mean squared error. We selected one sampler for each mode (with random parameters). The experiment were repeated $100$ times, both for MH and MALA base samplers, with different random initializations for the MCMC chains. Finally, the whole process was repeated $20$ times, drawing different parameters for the samplers.
\Cref{fig:mse_vs_sample_multimode_mh} shows the results for one of the $20$ settings with MH samplers (the results for all the $20$ settings followed the same pattern); both with and without the a priori knowledge of the weights. The results with MALA instead of MH are presented in \Cref{fig:mse_vs_sample_multimode_mala}. One can observe that, for known weights, the best method for selecting $I_t$ in both experiments is the random choice with probability proportional to $w_i \tilde{S}_{i,n_i}$. On the other hand, when the weights are needed to be estimated, the performance of our more informed methods (i.e., all except for the uniform random selection) is approximately the same, which is due to the unavoidable errors in estimating the weights (this effect is more visible for small sample sizes).

\subsection{Weight estimation}
\label{sec:weight}

\begin{figure}[t]
	\begin{center}
		\centerline{
			\includegraphics[width=0.5\columnwidth]{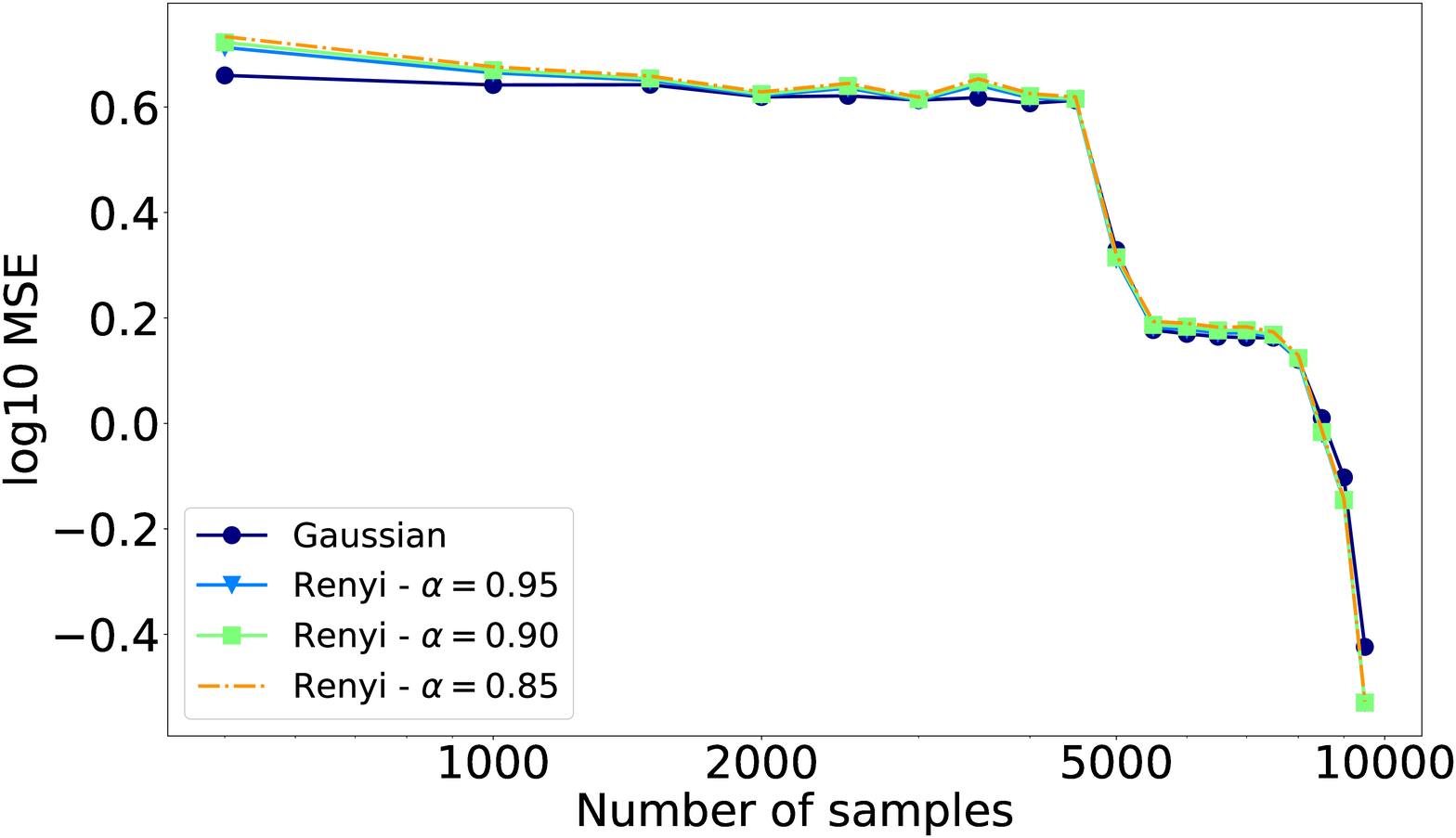}
			\includegraphics[width=0.5\columnwidth]{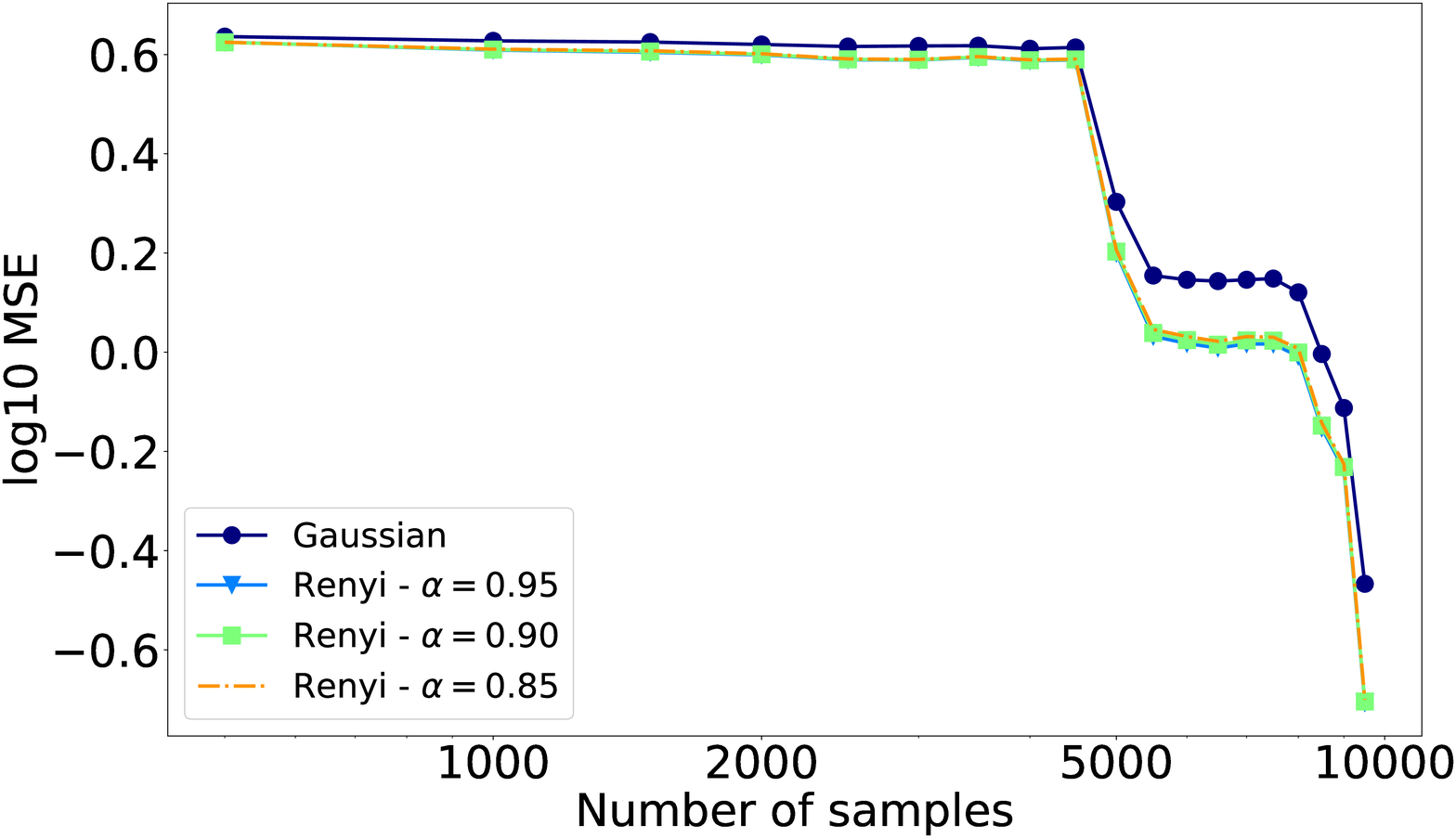} }
				\caption{Multimodal distributions (Gauss mixtures):  Mean squared error of different weight estimation methods for the setup in \Cref{fig:mse_vs_sample_multimode_mh} when the samplers receive equal budget. ``Gaussian'' refers to considering a Gaussian distribution for each mode and calculating its weight by estimating its covariance matrix, while ``R\'enyi'' refers to our proposed weight-estimation method based on R\'enyi entropy. Results for MH base samplers are shown on the left, while for MALA samplers on the right.}
		\label{fig:std_vs_renyi_multimode}
	\end{center}
\end{figure}

\begin{figure}[th]
	\begin{center}
		\centerline{
			\includegraphics[width=0.5\columnwidth]{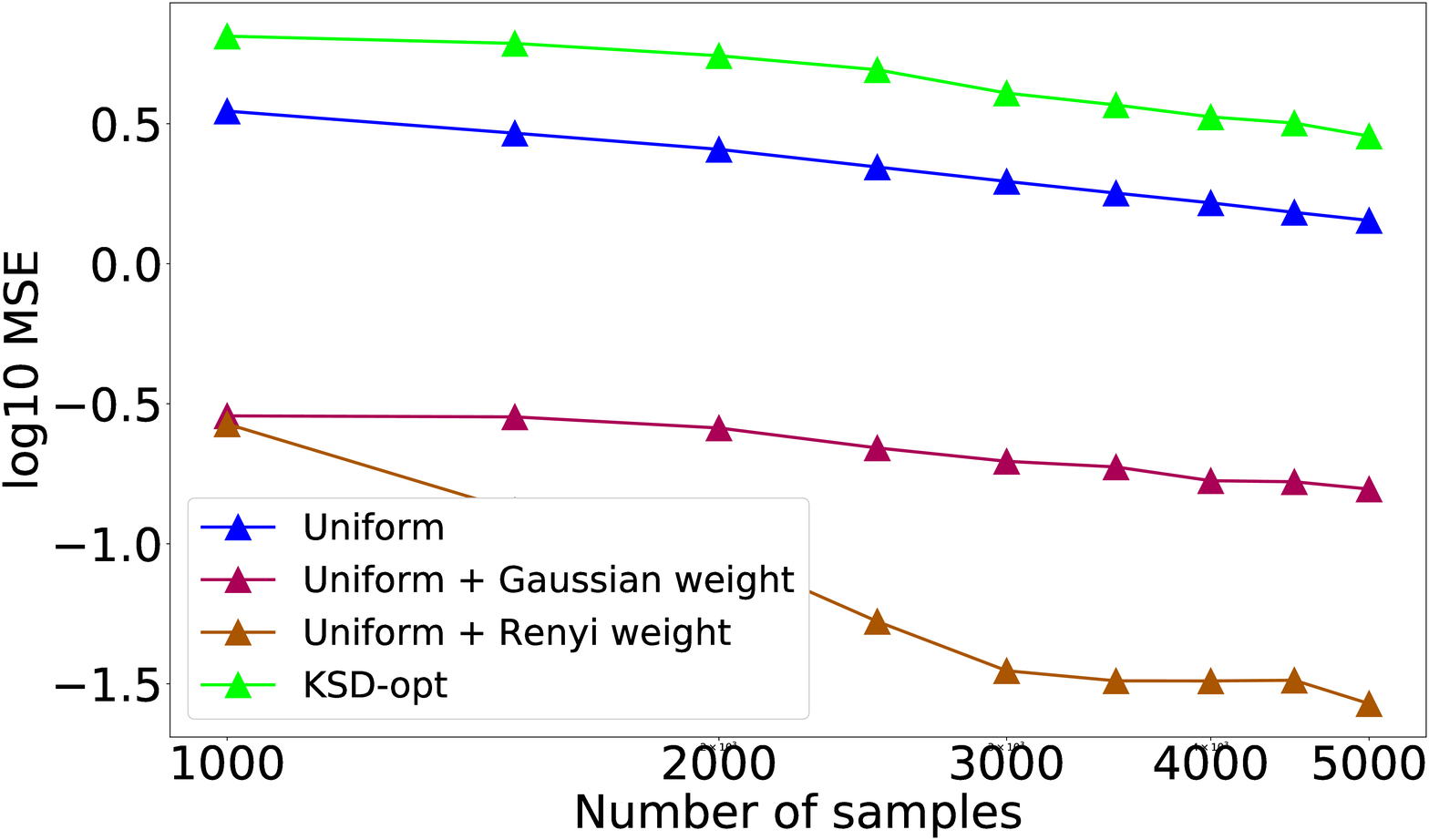}
			\includegraphics[width=0.5\columnwidth]{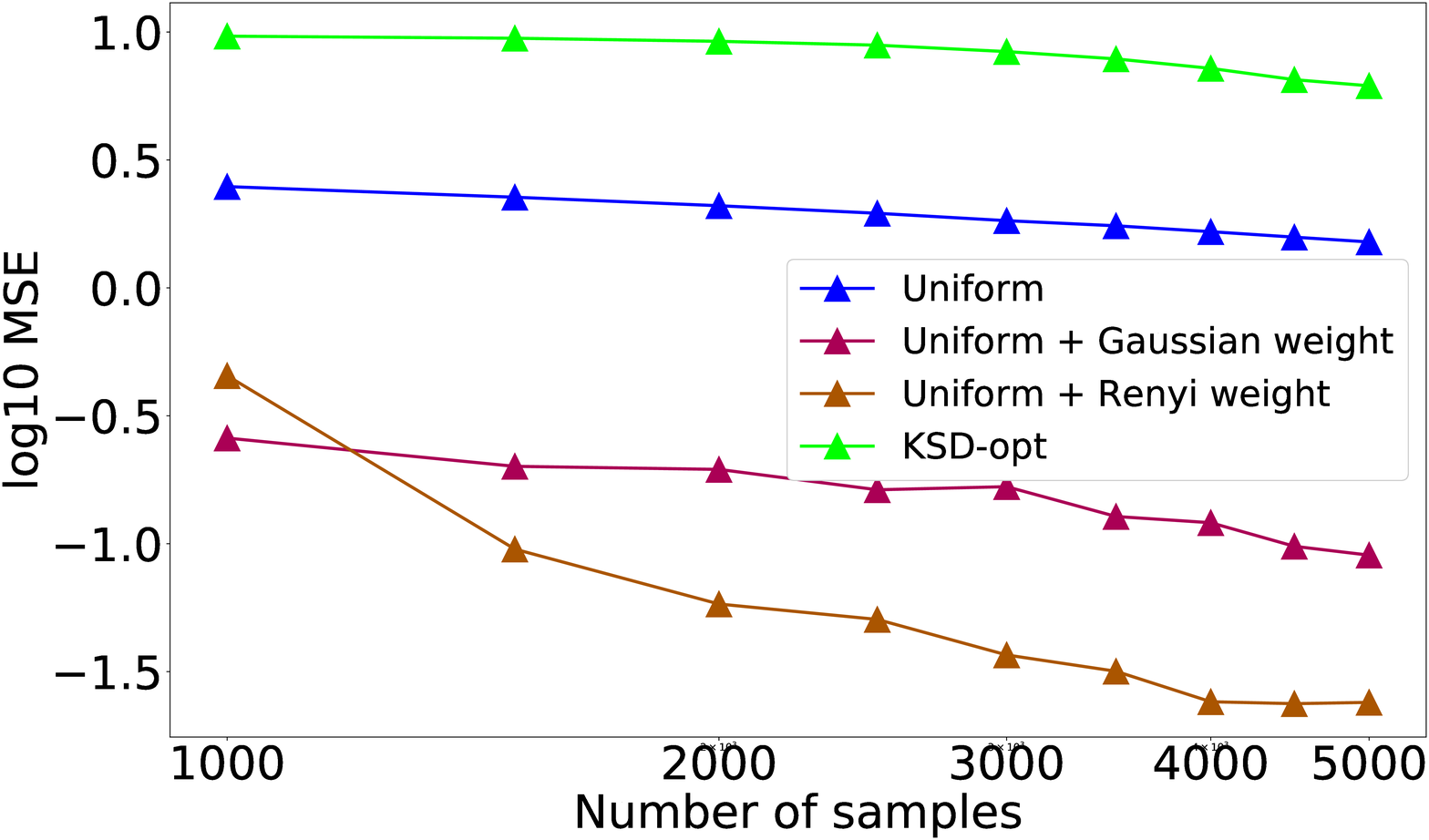} }
		\caption{Unknown number of modes:  Mean squared error of different weight estimation methods for a multimodal distribution in 2-dimensions (d=2)  when combining $10$ randomly initialized samplers. The samples are taken from a 2-dimensional Gaussian mixture model with 5 isotropic modes. The mean of each mode is selected uniformly at random from $[-5, 5]^2$ and the variance of each component is selected randomly from $[0.2, 1]$. ``Gaussian'' refers to considering a Gaussian distribution for each mode and calculating its weight by estimating its covariance matrix, while ``R\'enyi'' refers to our proposed weight-estimation method based on R\'enyi entropy with $\alpha=0.99$. Results for MH base samplers are shown on the left, while for MALA samplers on the right.}
		\label{fig:weight_estimation_2d}
	\end{center}
\end{figure}

\begin{figure}[th]
	\begin{center}
		\centerline{
		\includegraphics[width=0.5\columnwidth]{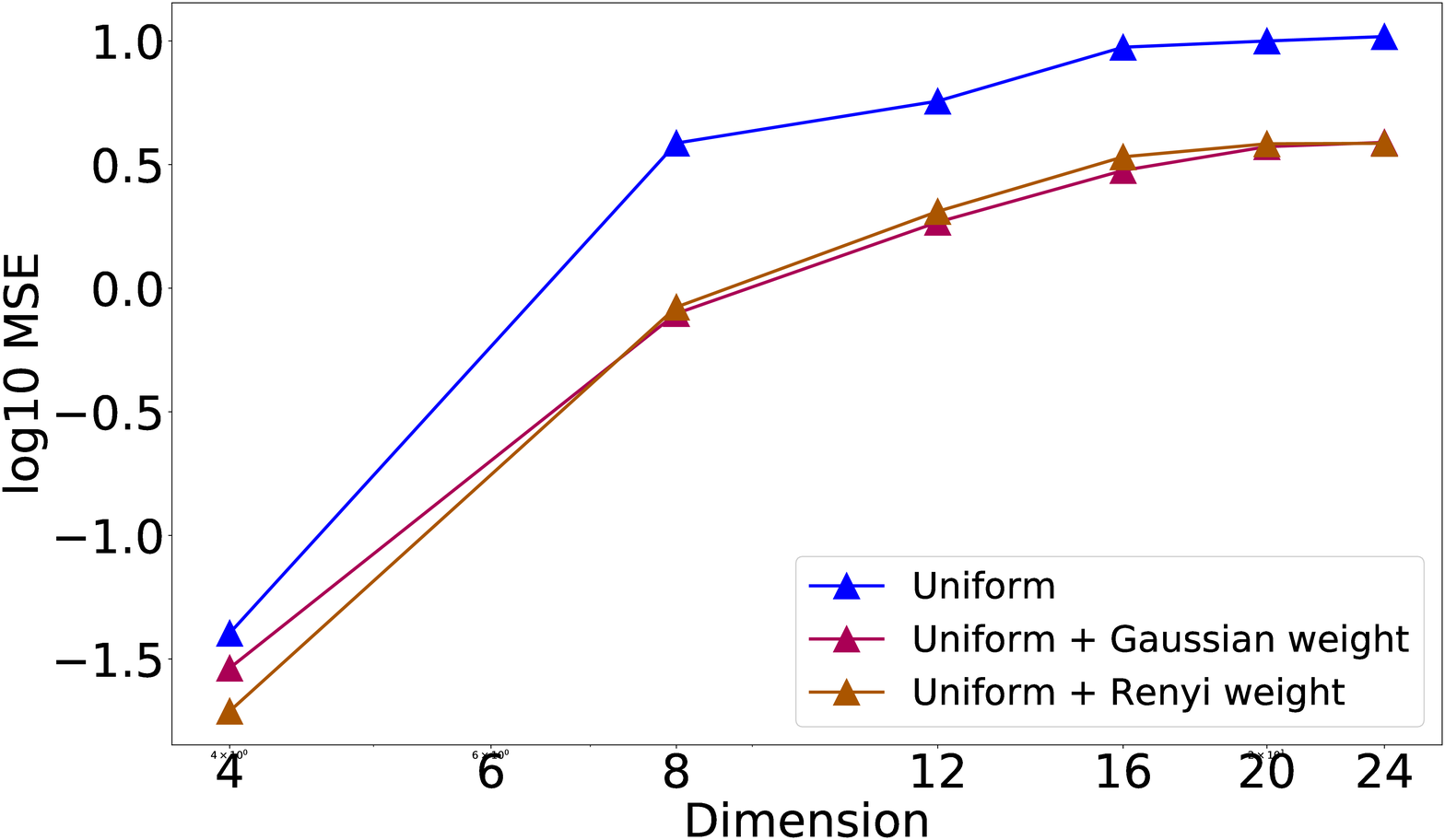}}
		\caption{Unknown number of modes:  Mean squared error of different weight estimation methods for a multimodal distribution in $d$-dimensions  when combining $10$ randomly initialized NUTS samplers. The samples are taken from a $d$-dimensional Gaussian mixture model with 5 isotropic modes. The mean of each mode is selected uniformly at random from $[-5, 5]^d$ and the variance of each component is selected randomly from $[0.5, 1]$. ``Gaussian'' refers to considering a Gaussian distribution for each mode and calculating its weight by estimating its covariance matrix, while ``R\'enyi'' refers to our proposed weight-estimation method based on R\'enyi entropy with $\alpha=0.99$. 		\label{fig:weight_dimension}}
	\end{center}
\end{figure}

A crucial step in the aggregation of the samples coming from different samplers is to estimate the weight of each mode. To this end, we repeated the same experiment with a uniformly random choice of $I_t$, but considering different methods for the final weight estimation. In particular, we compare the R\'enyi-entropy-based estimates with different $\alpha$ parameters and the natural weight estimates for Gaussian mixtures based on estimating the covariance matrix for each mode  \citep{scott_consensus_2016}. 
The results presented in \Cref{fig:std_vs_renyi_multimode} show that the weight estimation is not really sensitive to the choice of $\alpha$ (the recommendation of \citealp{szabo14information} is to use values close to $1$). Interestingly, in case of MALA base samplers, the R\'enyi-entropy-based methods outperform the ones designed specifically for the Gaussian distribution, which is indeed the underlying distribution in our case.
In a similar setup, \Cref{fig:weight_estimation_2d} compares several aggregation schemes, including the above tested R\'enyi- and, respectively, Gaussian-mixture-based combination, as well as uniform averaging and the computationally very expensive KSD-based reweighting scheme of \citet{liu_black-box_2016}. The results demonstrate that our method is able to outperform the others. On the other hand, extending the experiment to higher dimensions (and using NUTS as the base sampler, see \Cref{fig:weight_dimension}
), the Gaussian and the R\'enyi-entropy-based estimators perform very similarly (recall that the Gaussian estimator is specifically fitted to the Gauss mixtures considered in the experiments).

\subsection{The general case: unknown number of modes with multiple chains}
\label{sec:exp_general}

In this section, we consider a realistic  situation where we have access to an unnormalized density but the number of its separated regions/modes is unknown.
Again, the goal is to estimate the mean of the distribution.  Here we cannot guarantee that any MCMC chain used will only sample one mode; on the contrary, chains will usually move among different regions. We consider a Gaussian mixture model with $5$ modes of random parameters, and run 
$10$ base samplers whose parameters are also chosen randomly. We compare several combination methods discussed before: (i) each sampler is used to generate the same number of samples (Uniform); (ii) the same as the previous, but the final clustering and reweighting step of KSD-MCMC-WR (\Cref{algorithm:mcmc_bandit_global}) is used in the end (Uniform+clustering); finally, four versions of KSD-MCMC-WR are considered where the bandit method is $\eps$-greedy or UCB1, and the selection of $I_t$ is uniform at random (Equal probability) or random with probabilities proportional to $w_i \tilde{S}_{i,n_i}$. For clustering, we used the "Kmean" method of the scikit-learn package \citep{scikit-learn}. 

\begin{figure}[t]
\vspace{-0.5cm}
	\begin{center}
		\centerline{
			\includegraphics[width=0.5\columnwidth]{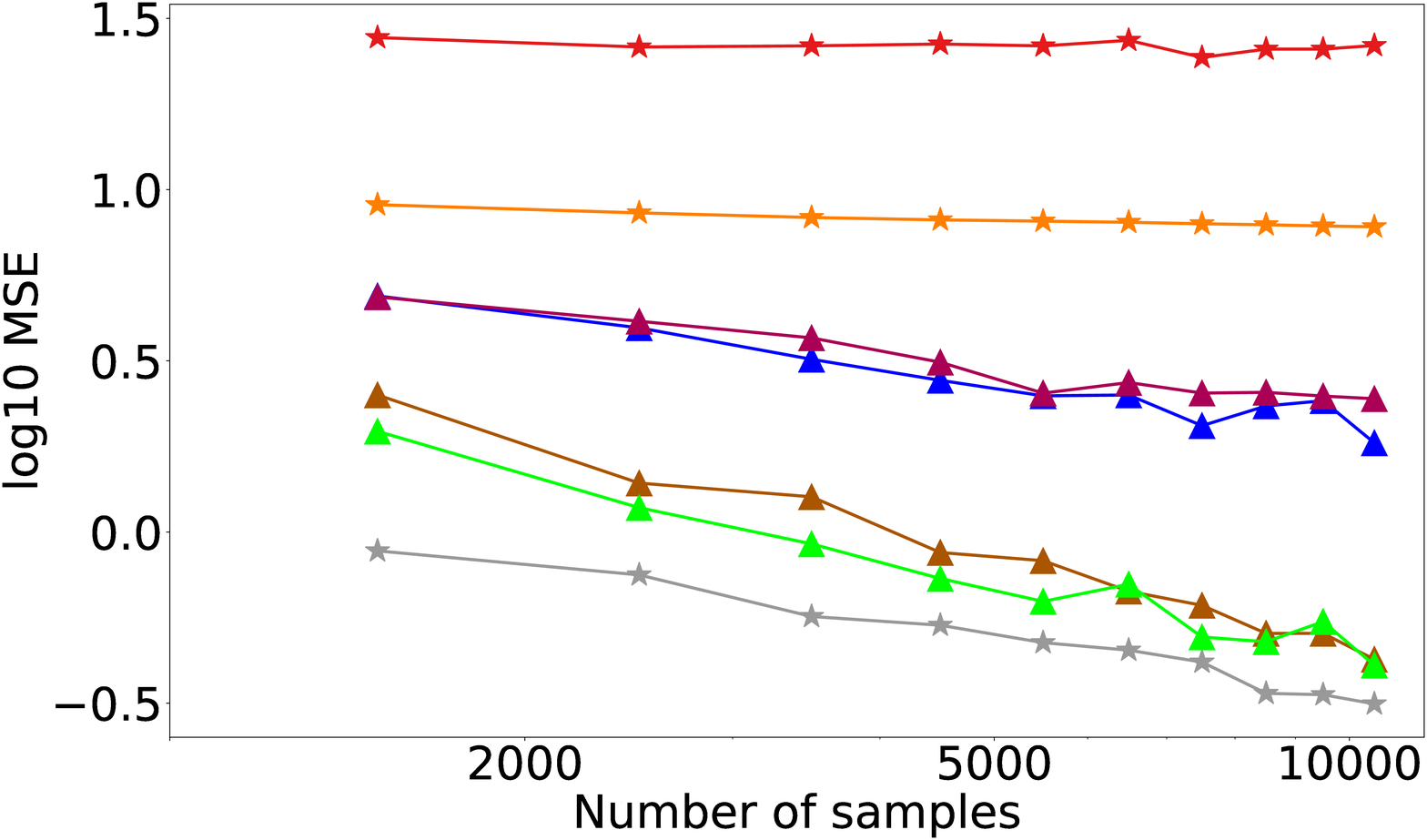} 
			\includegraphics[width=0.5\columnwidth]{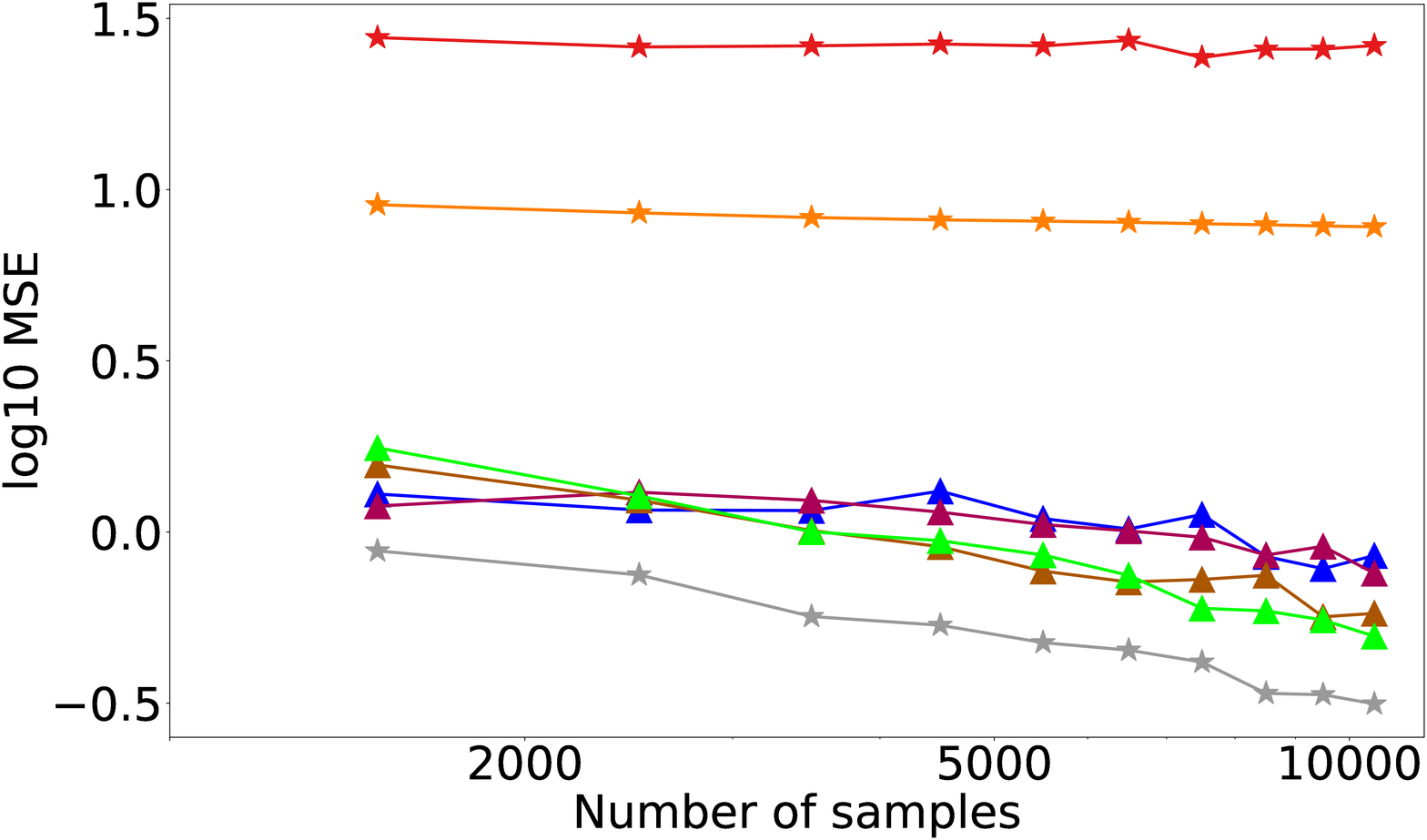}}	\vspace{-8pt}
		\centerline{
		\includegraphics[width=0.5\columnwidth]{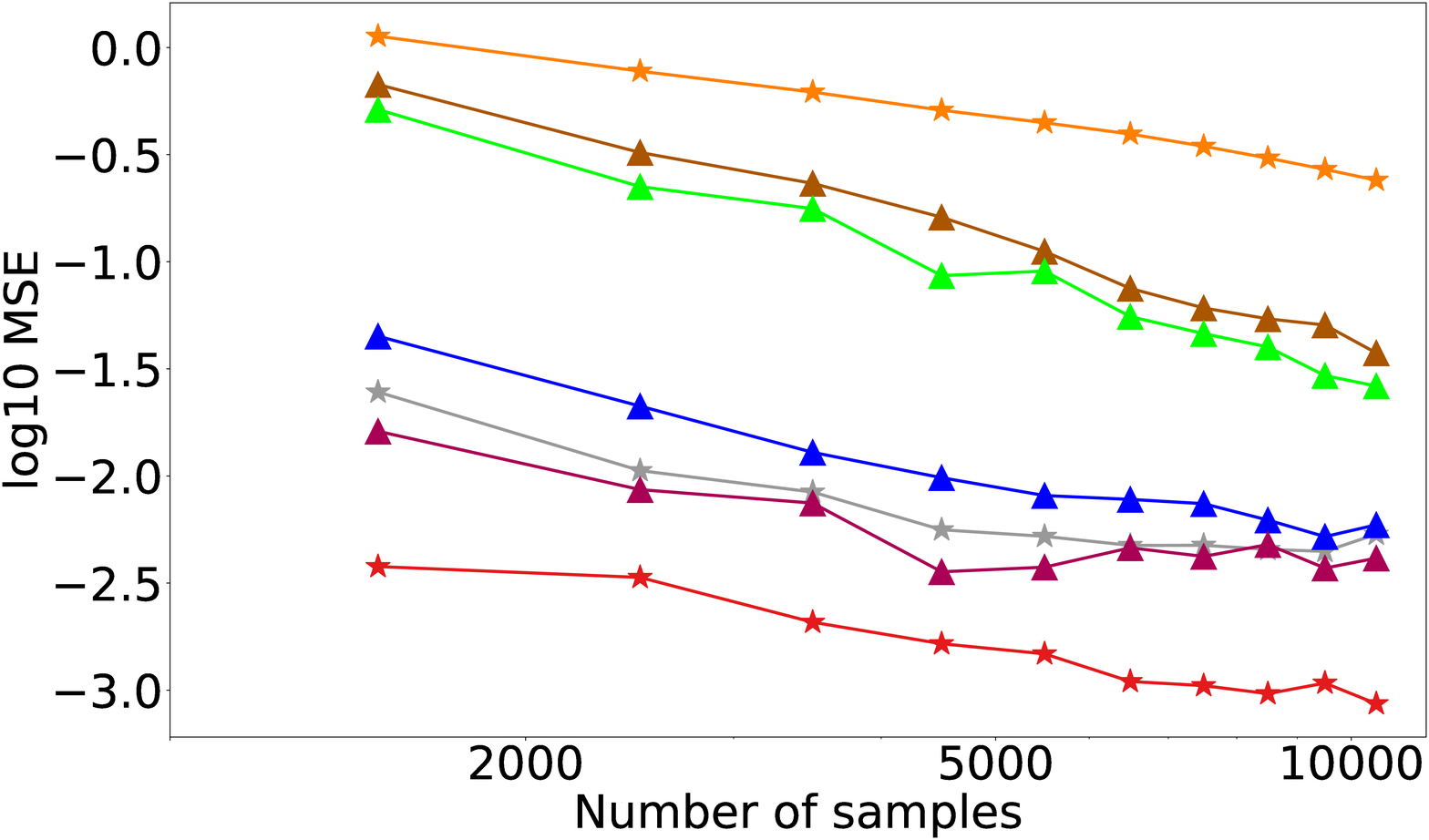} 
		\includegraphics[width=0.5\columnwidth]{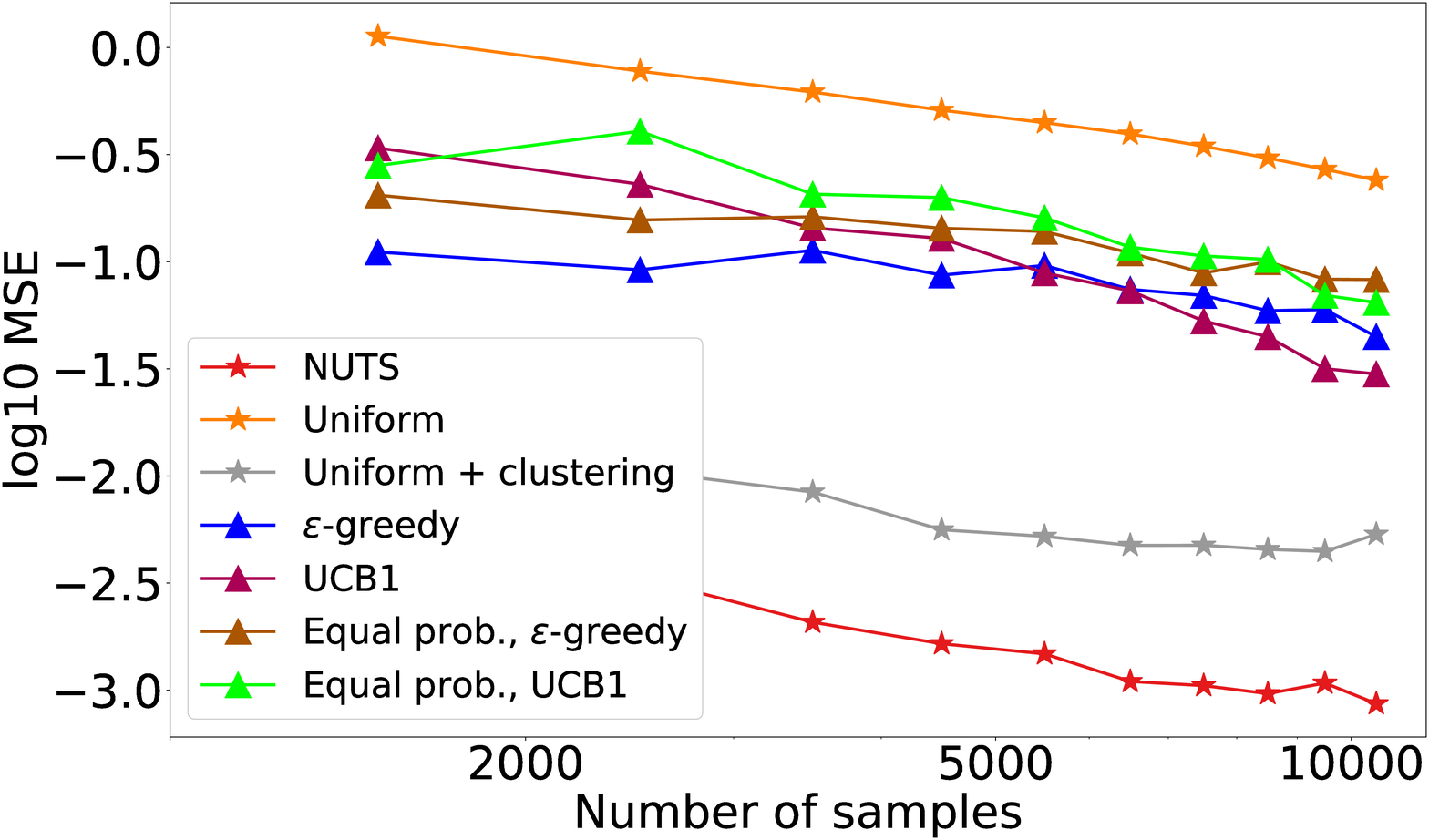}}
\tiny	\caption{Unknown number of modes: MSE for different sample sizes when combining $10$ randomly initialized MH samplers.  The samples are taken from a 2-dimensional Gaussian mixture model with 5 isotropic modes. The mean of each mode is selected uniformly at random from $[-5, 5]^2$ and the variance of each component is selected randomly from $[0.2, 1]$. Two representative cases are shown: in the top row the modes are far from each other, in the bottom row the modes are close. The left column corresponds to batch size $10$, while it is $100$ for the right column. R\'enyi entropy with $\alpha=0.99$ is used for weight estimation. }
	\label{fig:mse_vs_sample_multimode_bandit}
\end{center}
\end{figure}

After randomly choosing the distribution and the samplers, we considered them fixed and ran $100$ experiments; in each of them the sampling methods are  started from random initial points. The whole process was then repeated for $20$ different sets of distributions and samplers. \Cref{fig:mse_vs_sample_multimode_bandit} shows the results for two representative cases (out of $20$) when MH is used as the base sampler. The top row corresponds to a situation when the modes are far from each other; in this case our proposed  method significantly outperforms NUTS and also unweighted uniform usage of the samplers. The bottom row depicts a case when  the modes are really close together, in which case NUTS performs the best. Interestingly, in these experiments the application of any bandit algorithm is not really beneficial as the best combination algorithm is essentially using the samplers uniformly and then reweighting the samples via our clustering method. 

\begin{figure}[t]
	\begin{center}
		\centerline{
			\includegraphics[width=0.5\columnwidth]{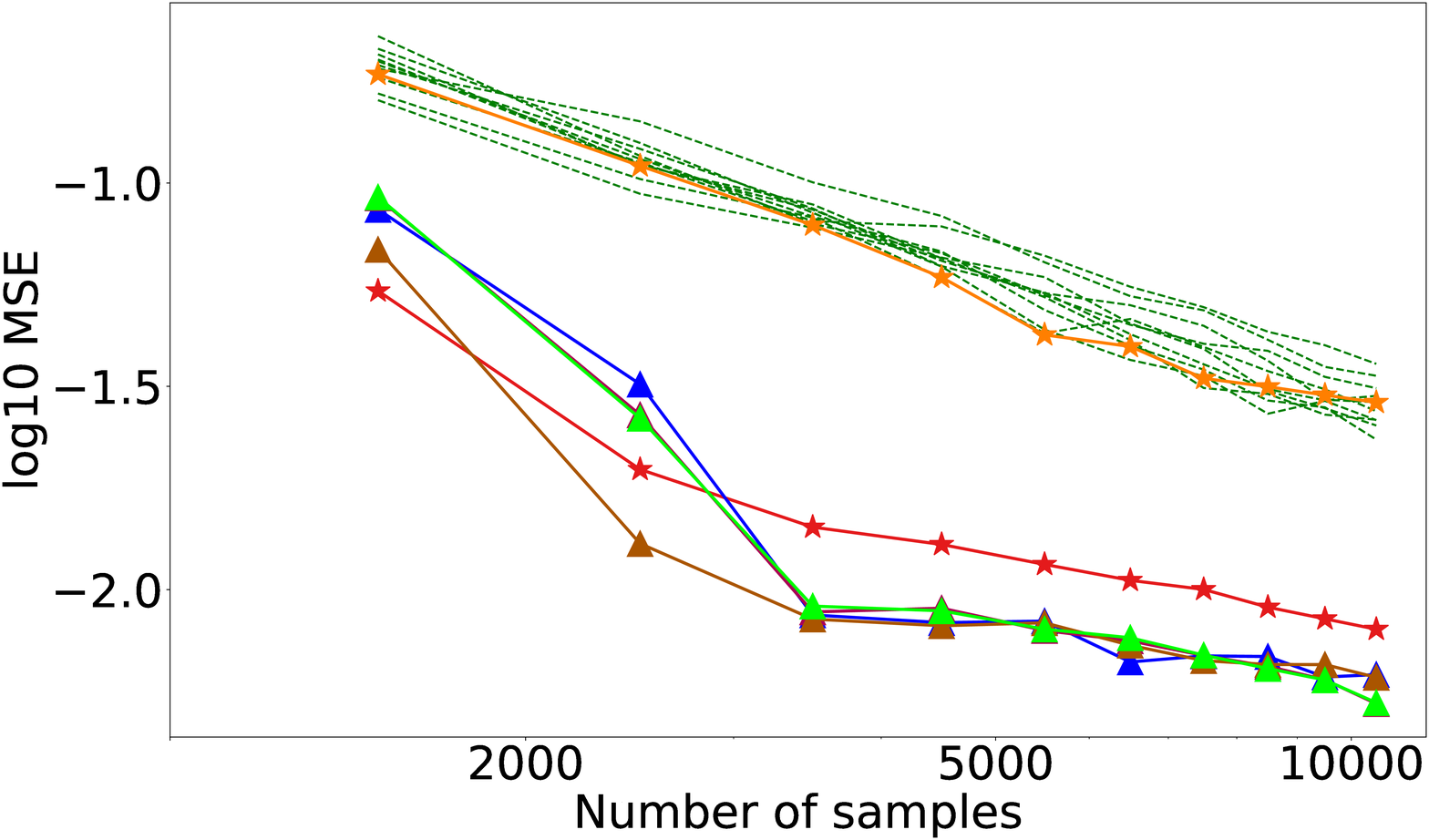} 
			\includegraphics[width=0.5\columnwidth]{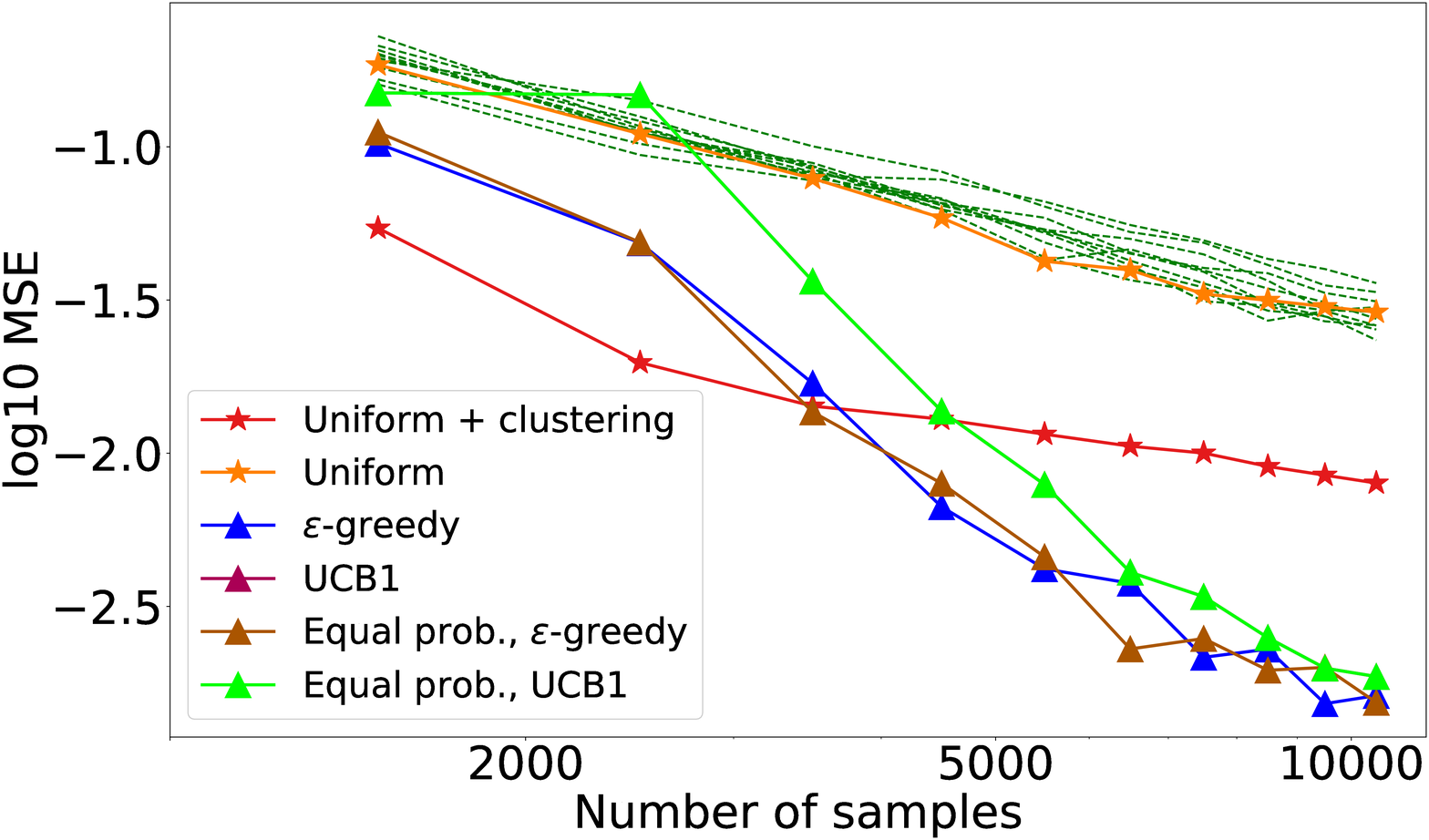}}	
\tiny		\caption{MSE for different sample sizes with separated modes: KSD-MCMC-WR with $10$ randomly initialized NUTS samplers for an unknown number of modes. The batch size on the left figure is $10$ and on the right figure is $100$. The dashed green lines show the results for the individual NUTS samplers. R\'enyi entropy  with $\alpha=0.99$ is used for weight estimation. The problem setup is the same as the one in \Cref{fig:mse_vs_sample_multimode_bandit}. }
		\label{fig:mse_vs_sample_multimode_bandit_nuts}
	\end{center}
	\vspace{-0.3cm}
\end{figure}

On the other hand, if  the MH samplers are replaced with NUTS samplers, a more careful selection of which sampler to use becomes important. This is because these samplers explore individual modes efficiently, thus using multiple samplers for the same mode is really wasteful.
\Cref{fig:mse_vs_sample_multimode_bandit_nuts} shows the results for another representative example out of the $20$ cases that we used in the previous experiment.
The results indicate that our combination method (with various choices for the details)  can significantly outperforms NUTS. Also, as mentioned before, using bandit algorithms to select from the base samplers is much better than just choosing them uniformly. 

\begin{figure}[t]
	\begin{center}
		\centerline{
			\includegraphics[width=0.5\columnwidth]{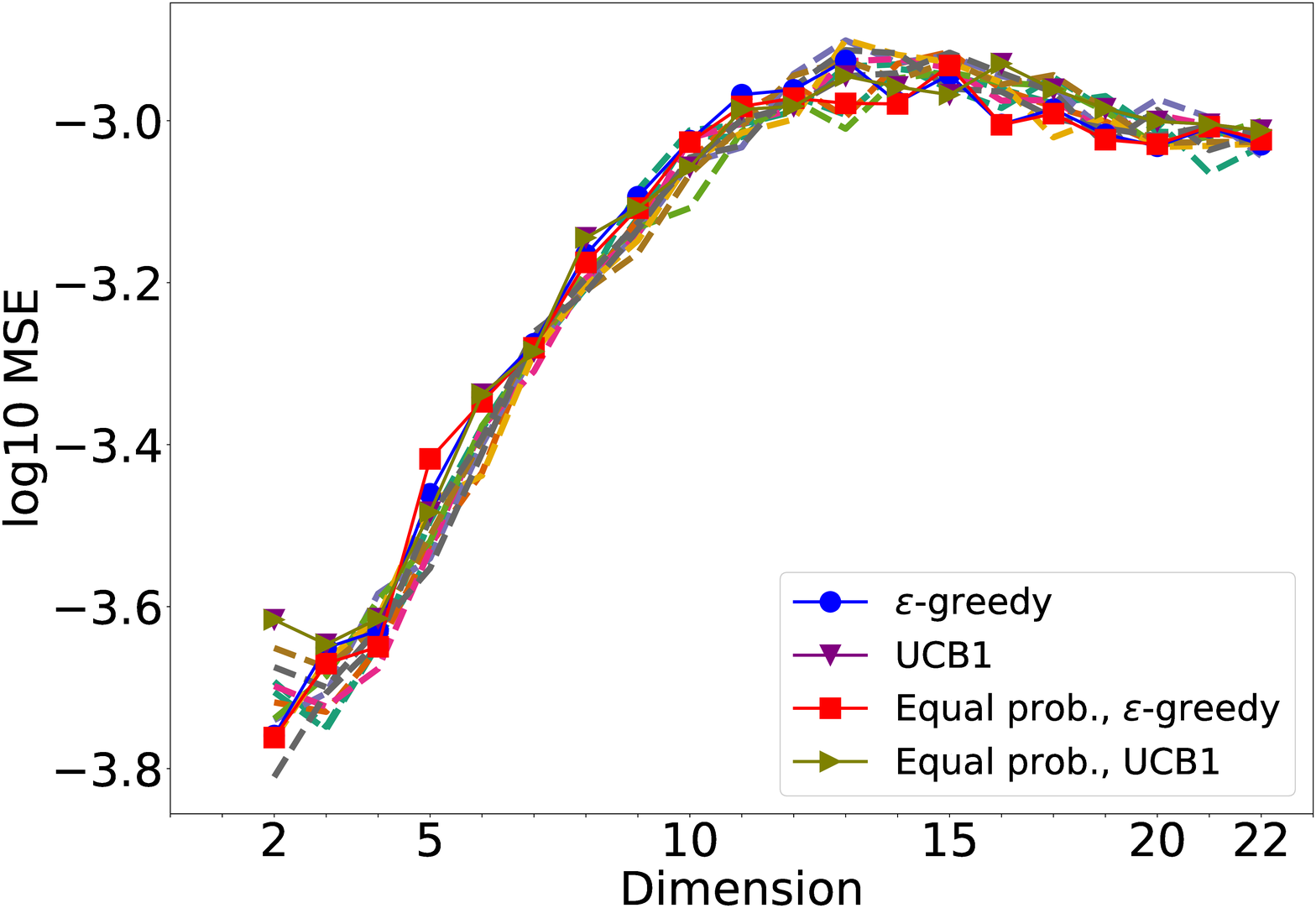} 
			\includegraphics[width=0.5\columnwidth]{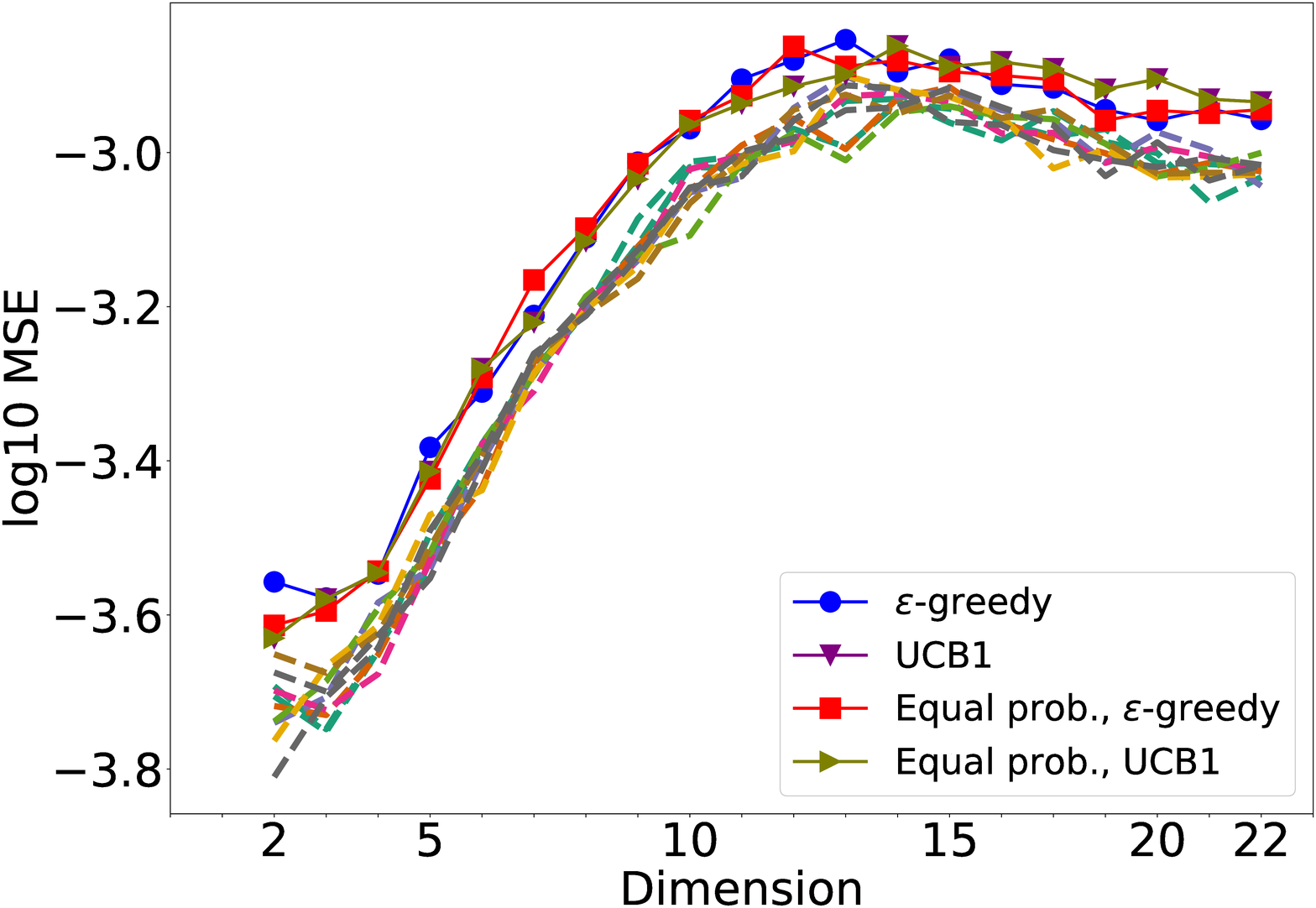} 
		}
		\tiny
		\caption{Running KSD-MCMC-WR for a multivariate standard normal distribution %
		with $10$ NUTS instances started with different (random) initializations. 
		The batch size is $10$ on the left and $100$ on the right.}
		\label{fig:bandit_singlemode_multinuts}
	\end{center}
	\vspace{-0.2cm}
\end{figure}

Although the previous experiment demonstrated that the KSD-MCMC-WR is preferable to NUTS when the modes of the distribution are separated, it is not sufficient to argue that KSD-MCMC-WR is preferable to NUTS. In order to be able to make such a claim, we should also look at cases in favor of NUTS; the most favorable being the case of a unimodal distribution. To this end, we compared the performance of NUTS and KSD-MCMC-WR when sampling from a multivariate standard normal distribution. 
\Cref{fig:bandit_singlemode_multinuts} shows that running KSD-MCMC-WR is competitive with the individual NUTS instances in this case.

These results suggest that \emph{one should always run KSD-MCMC-WR with NUTS base samplers}, as it provides significant improvements for multimodal distributions while remains a safe choice for the easier, unimodal cases.

\subsection{Comparison with parallel MCMC methods}
\label{sec:parallel}

In the previous sections, we mostly compared our algorithms with MCMC methods (MH, MALA, and NUTS) that are only expected to work well for essentially unimodal densities (i.e., densities with a single ``connected'' region), and the vanilla parallelization (i.e., uniformly mixing multiple chains) is not sufficient to achieve good performance on multimodal problems. 
Therefore, we also compare our algorithms  with two popular parallel MCMC methods, annealed sequential Monte Carlo with resampling (SMC) \citep{demoral_sequential_2006} and parallel tempering (PT) \citep{earl_parallelTT_2005}, which are proposed to deal with multimodal distributions. Following general practice in the literature (similarly, e.g., to \citealp{liu_black-box_2016}) and in line with our previous experiments, we use the number of times the samplers calculate $p(x)$ for generating their final samples as a proxy for the runtime (a comparison based on actual runtimes is impossible since the algorithms are implemented using different packages). Throughout the experiments, for SMC and PT, we use $10$ different temperature values for annealing with an exponential ladder of base $\sqrt{2}$ (with inverse temperature values $0, 2^{-4}, 2^{-3.5},\ldots,1$). For SMC, we take one MCMC step between two resampling steps, and for PT $25$ steps between two swapping attempts.

\begin{figure}[t]
	\begin{center}
\centerline{
	\includegraphics[width=0.49\columnwidth]{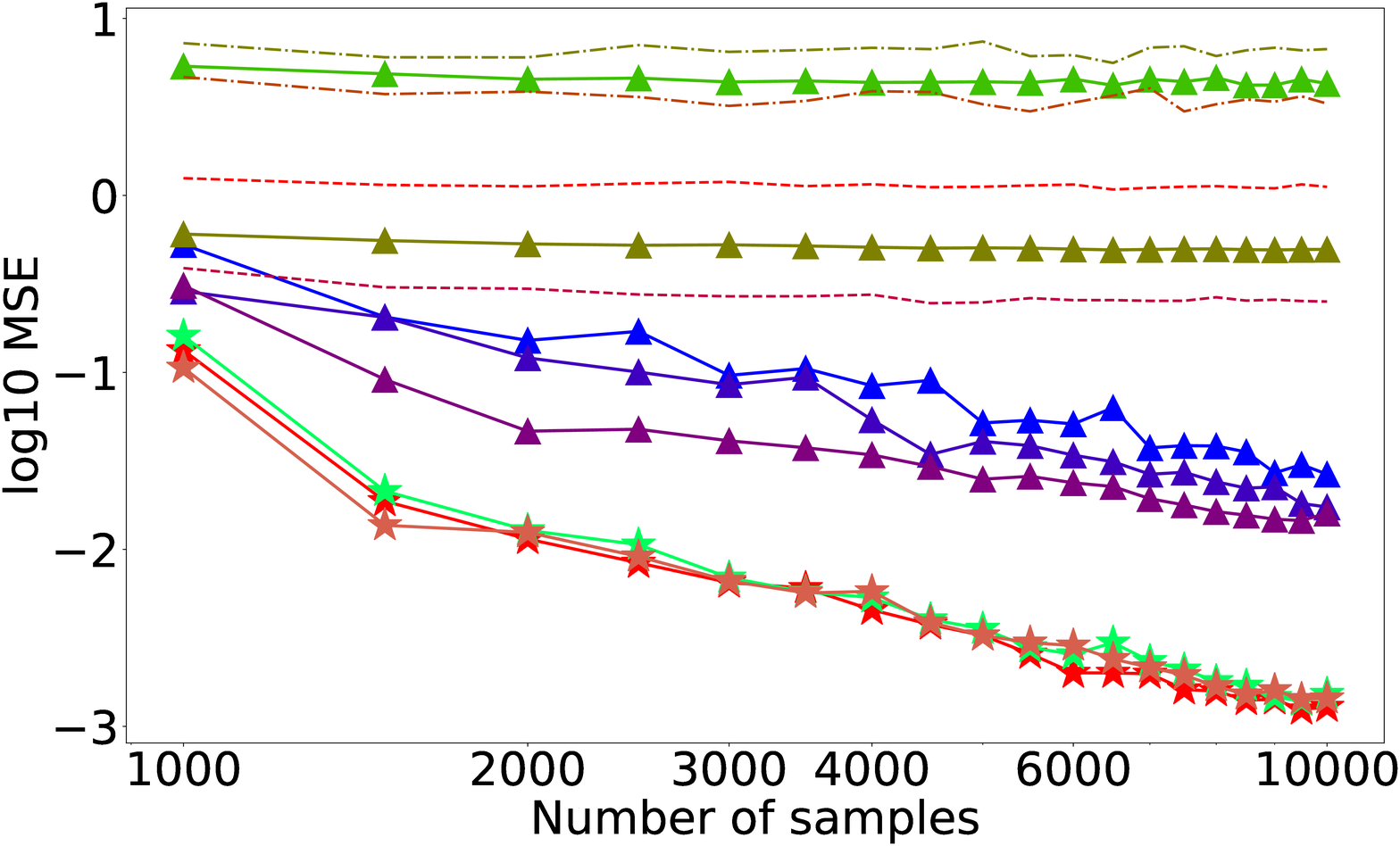}
	\hskip -22pt	\includegraphics[width=0.54\columnwidth]{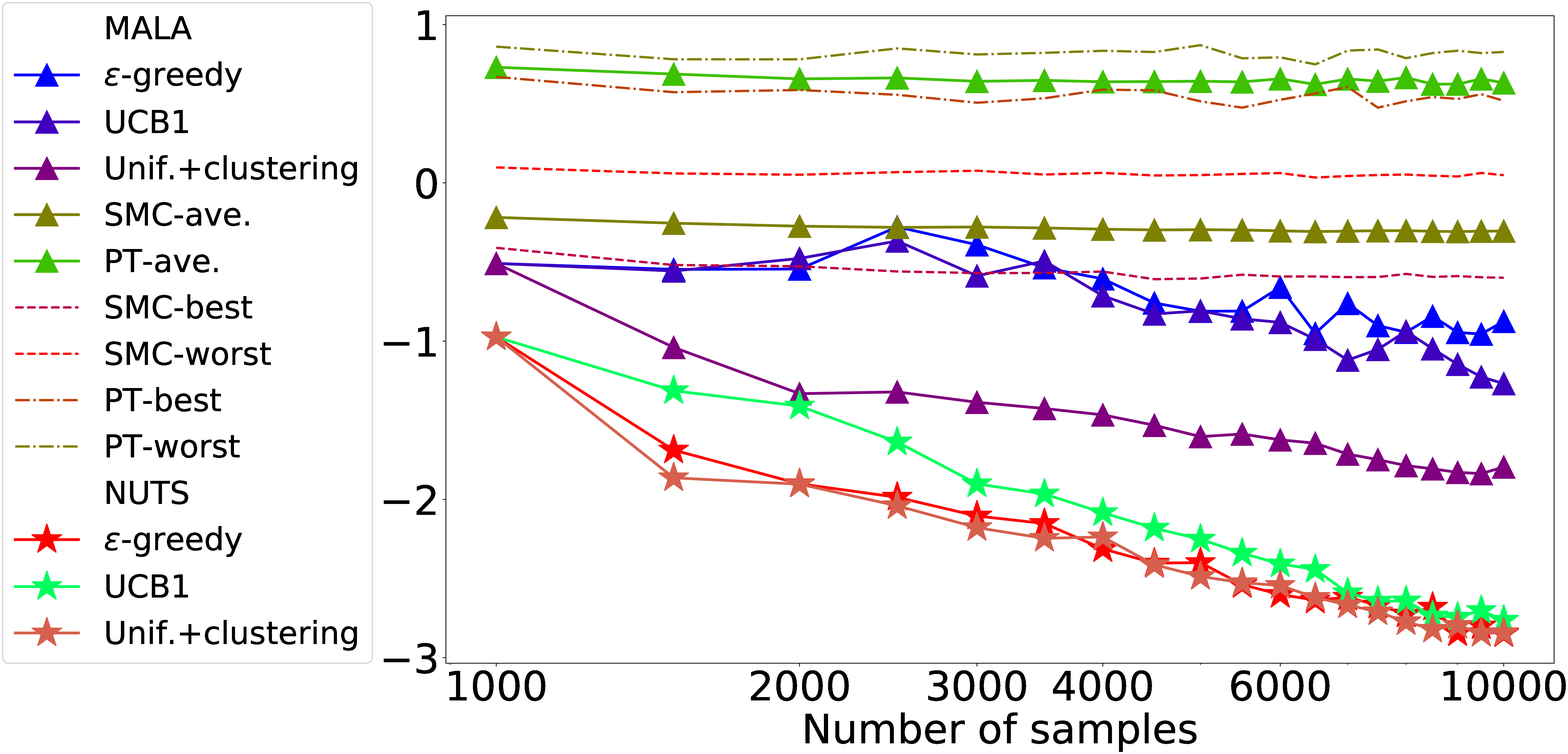} }
		\caption{Unknown number of modes: MSE for different sample sizes when combining $10$ randomly initialized  MALA samplers for a  2-dimensional Gaussian mixture model with 5 isotropic modes. The mean of each mode was selected uniformly at random from $[-5, 5]^2$ and the variance of each component was selected uniformly from $[0.2, 1]$. SMC and PT were run separately for every parameter setting of MALA ($10$ different parameters) with the best, worst and average values reported, while KSD-MCMC-WR combined the same samplers. In addition, the results for KSD-MCMC-WR with $10$ randomly initialized NUTS base samplers were included for completeness. Batch size for KSD-MCMC-WR is $10$ on the left figure, $100$ on the right. Each curve is obtained by averaging over $100$ runs with different random initializations of the MALA/NUTS samplers (with the same parameters).}
		\label{fig:ksd_parallel}
	\end{center}
\end{figure}

While our algorithm works best with NUTS as its base sampler, combining SMC and PT with NUTS is problematic: SMC uses a large initial population of samples at the beginning and so the corresponding chains started from these points are too short (at least in the sample sizes we consider) for NUTS to be able to adapt. Similar problems occur with PT: since PT runs parallel chains whose states are swapped frequently, after each swap NUTS should be run for sufficiently long time to adapt to the new location, which slows down the process.\footnote{We have run PT with the pym3 implementation of  NUTS, but the results were not good enough to be included.}  Of course, one could try to avoid these problems by coming up with some clever method for integrating NUTS, but this is out of the scope of this paper.
Therefore, as a fair comparison, first we tested KSD-MCMC-WR, SMC and PT all with MALA as their base sampler in a $2$-dimensional multimodal  problem.  
The parameters of the target distribution were selected randomly (as described in \Cref{fig:ksd_parallel}) and kept fixed during the experiment. Other target distributions yielded similar results.
We ran SMC and PT with $10$ different parameters of MALA, which were also used in KSD-MCMC-WR. The results, averaged over $100$ runs with different initializations of the base samplers, are shown in \Cref{fig:ksd_parallel}: it can be seen that our algorithms significantly outperformed PT and provided better results than SMC on average. The significance of our approach becomes clear by noticing that it is not even clear how one should set the MALA parameters for SMC in advance. It is interesting to note that in this experiment, replacing the bandit algorithms with a uniformly random selection of the chains followed by the R\'enyi entropy-based reweighting (``uniform+clustering'' in the figure) has the best performance among all the algorithms with MALA base samplers, and in this case KSD-MCMC-WR works better with the smaller batch size $10$ than with $100$. This phenomenon is related to how easily the MALA samplers move from one mode to another, how it affects clustering, and how suitable the KSD measure becomes in this case to compare the sample quality of different chains clustered together, and is discussed in detail in the next section.
For completeness, the results for KSD-MCMC-WR with NUTS as the base sampler are also shown in \Cref{fig:ksd_parallel}, which confirms again that our methods work much better with NUTS, and our best method significantly outperformed all the other competitor algorithms.

\begin{figure}[t]
	\begin{center}
		\centerline{
			\includegraphics[width=0.5\columnwidth]{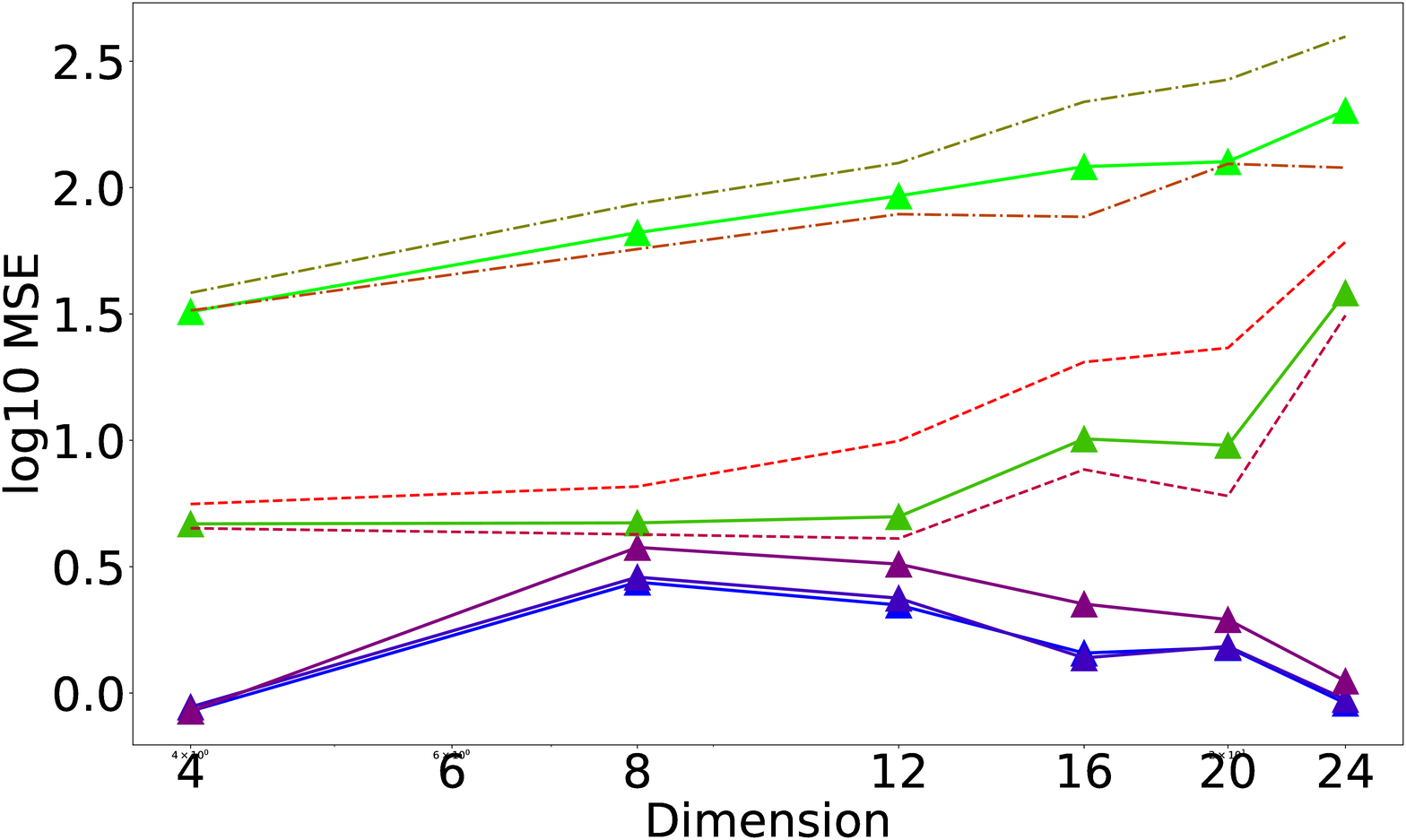}
			\hskip -22pt
			\includegraphics[width=0.56\columnwidth]{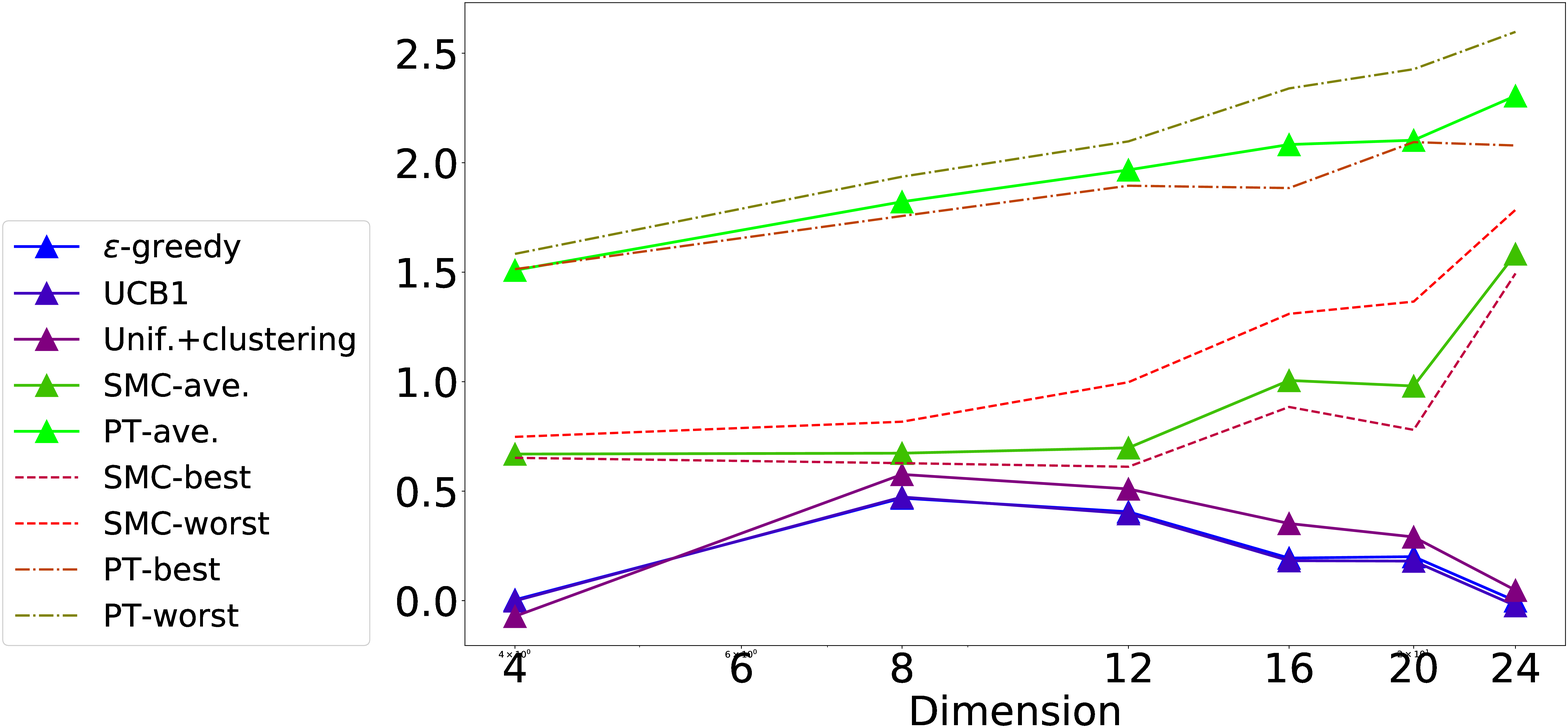} }
		\caption{Unknown number of modes: MSE for different dimensions when using KSD-MCMC-WR with $10$ randomly initialized  NUTS  samplers for a $d$-dimensional Gaussian mixture model with 5 isotropic modes. The mean of each mode was selected uniformly at random from $[-5, 5]^d$ and the variance of each component was selected randomly from $[0.5, 1]$. SMC and PT were run separately for every parameter setting of MALA ($10$ different parameters) with the best, worst and average values reported.  Batch size for KSD-MCMC-WR is $10$ on the left figure, $100$ on the right. Each curve is obtained by averaging over $100$ runs with different random initializations of the MALA/NUTS samplers (for the same parameters)}
		\label{fig:ksd_parallel_dimension_1}
	\end{center}
\vspace{-0.9cm}
\end{figure}

\begin{figure}[]
	\begin{center}
			\centerline{
			\includegraphics[width=0.5\columnwidth]{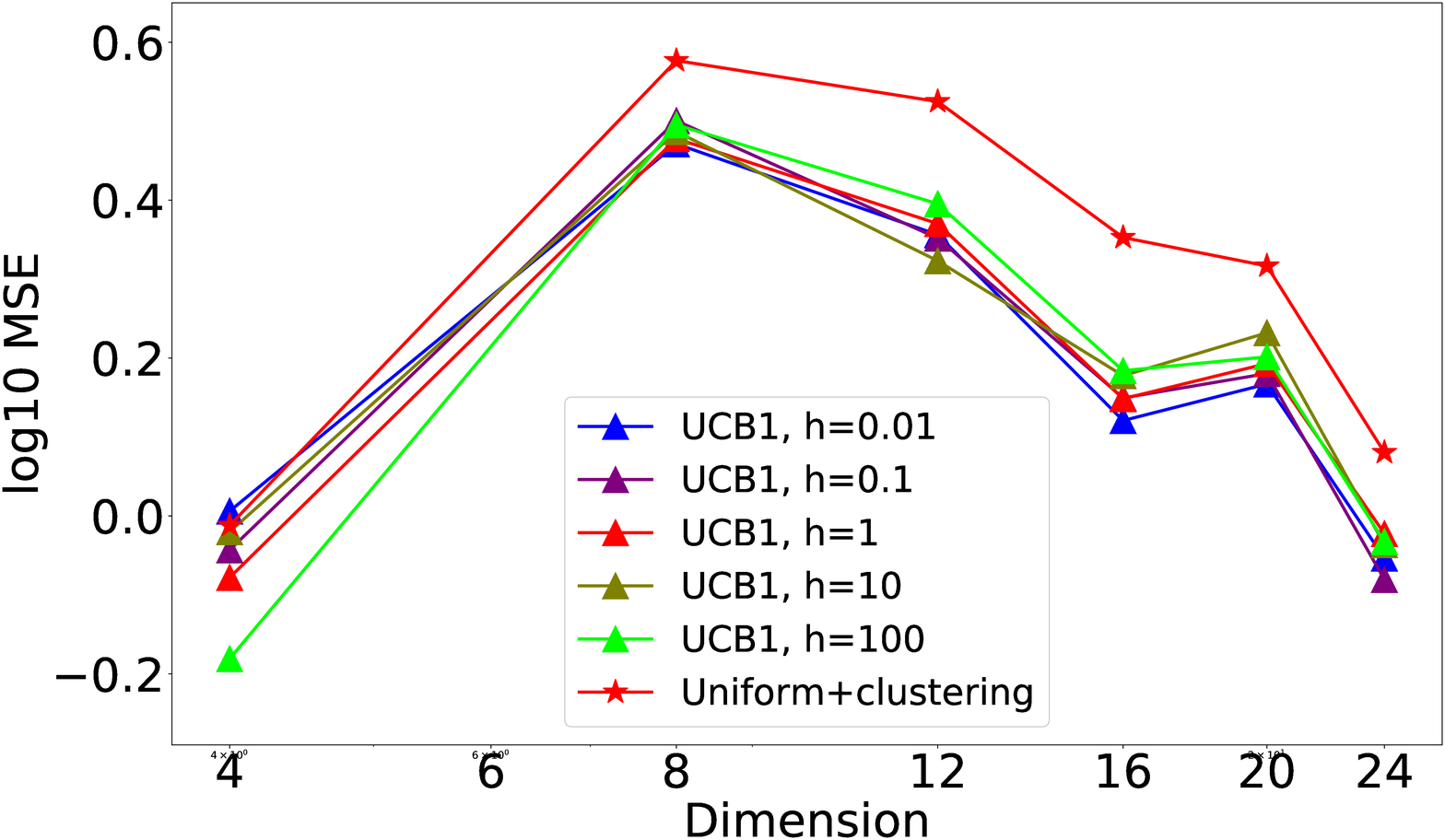}
			\includegraphics[width=0.5\columnwidth]{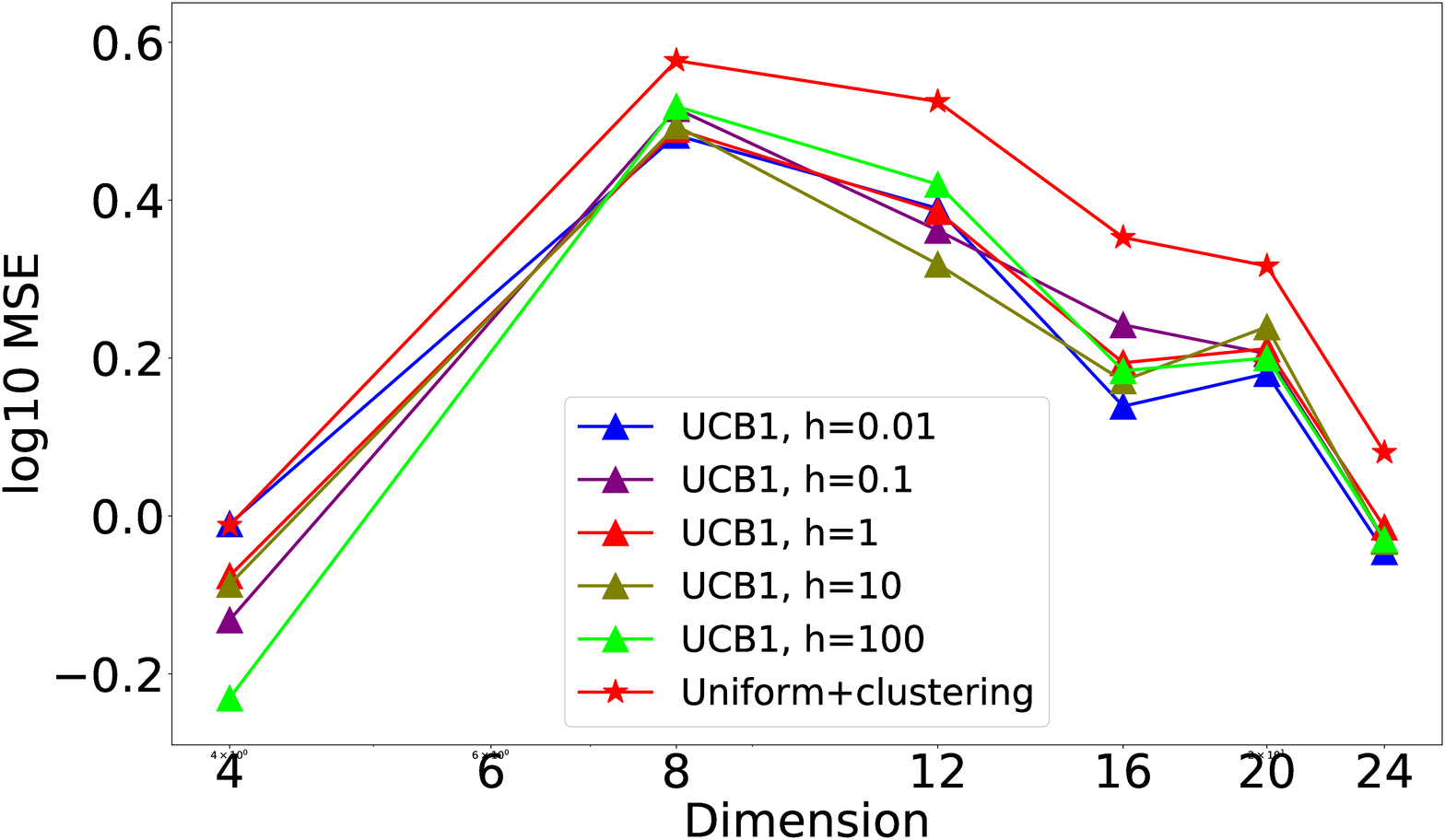}}\vspace{-8pt}
		\centerline{
			\includegraphics[width=0.5\columnwidth]{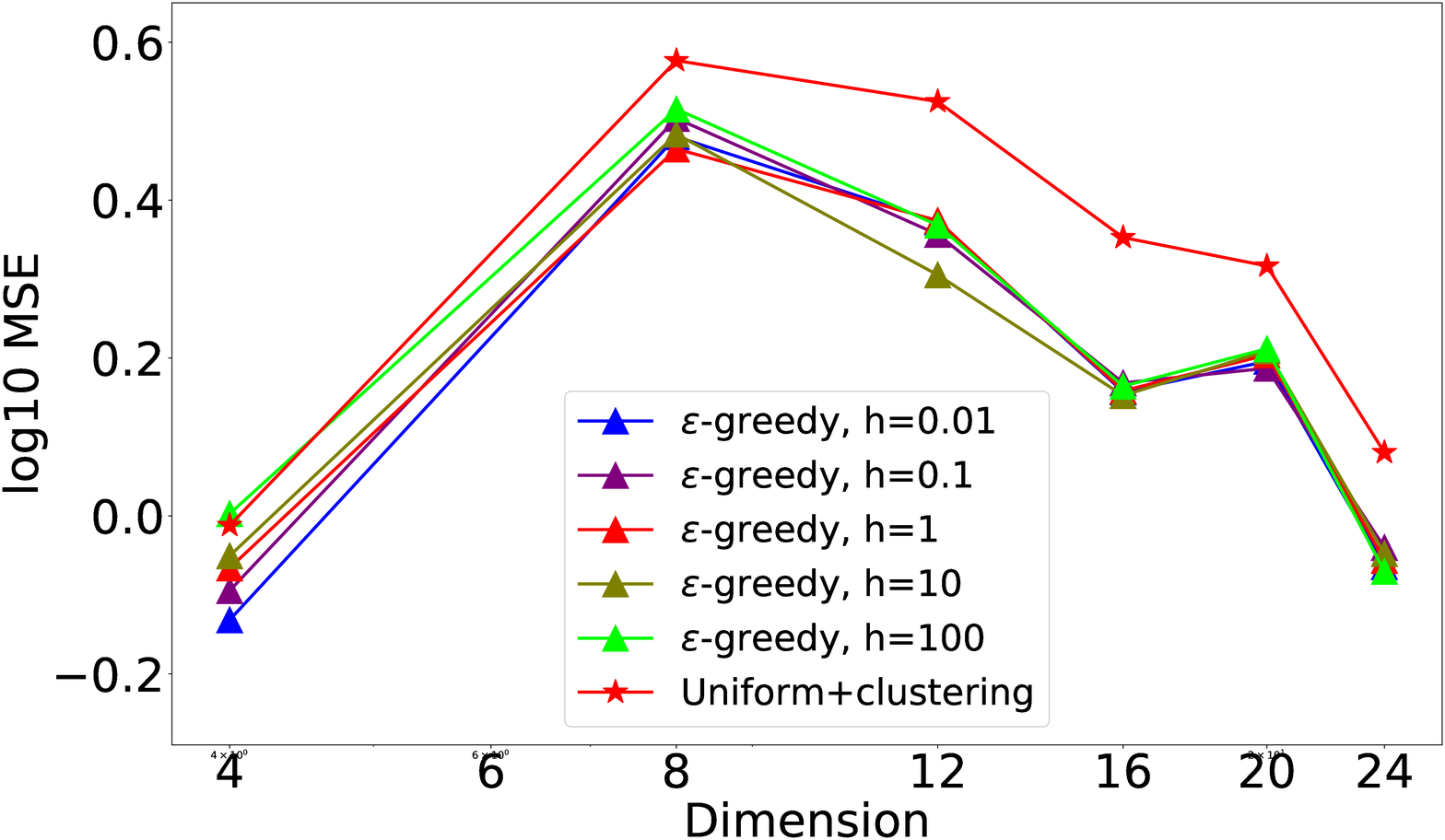} 
			\includegraphics[width=0.5\columnwidth]{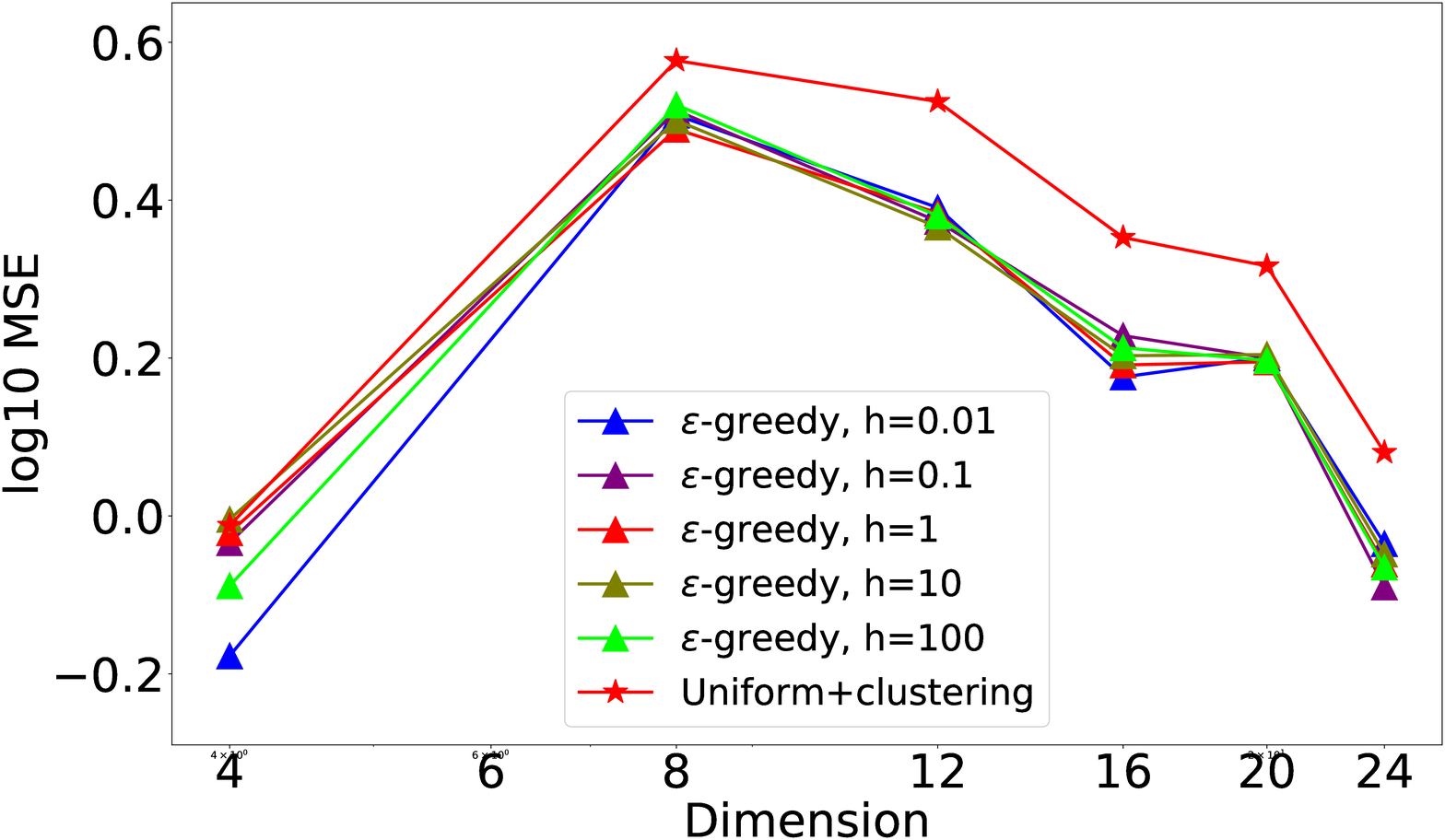}}
		\caption{Changing the KSD kernel width $h$: MSE for different dimensions when using KSD-MCMC-WR with $10$ randomly initialized  NUTS  samplers (the same setting as in \Cref{fig:ksd_parallel_dimension_1}). Batch size for KSD-MCMC-WR is $10$ on the left figures, $100$ on the right; the performance of the UCB1 variant in shown in the top row, while that of the $\eps$-greedy version in the bottom. Each curve is obtained by averaging over $100$ runs with different initializations.}
		\label{fig:ksd_free}
	\end{center}
\end{figure}

In the second experiment,  we compared the performance of our best algorithm, KSD-MCMC-WR  with a NUTS base sampler, and that of SMC and PT with MALA.
\Cref{fig:ksd_parallel_dimension_1} shows how the relative performance of the algorithms change with the dimension for randomly selected Gaussian mixture distributions (a single $24$-dimensional distribution with 5 modes was selected randomly and its first $d$ coordinates were used in the experiments for $d$-dimensions\footnote{Other target distributions gave similar results.}):
It can be seen that KSD-MCMC-WR with NUTS consistently outperformed SMC and PT, and the bandit selection methods (UCB1 and $\eps$-greedy) in KSD-MCMC-WR performed better than the baseline uniform selection.
Finally, the last experiment was repeated with different kernel widths, $h \in \{0.01, 0.1, 1, 10, 100\}$. The results in \Cref{fig:ksd_free} show that our proposed algorithm is insensitive to this parameter in a wide range, and also that the performance of the UCB1 and $\eps$-greedy variants are very close to each other.\footnote{To reduce clutter, the results for SMC and PT are not shown in this figure.}


\subsection{Comparing the components of KSD-MCMC-WR: bandit algorithms and clustering-reweighting}
\label{sec:ksd_vs_uniform}
The two novel components of our final algorithm, KSD-MCMC-WR, are (i) using bandit algorithms with clustering and approximations to KSD and (ii) clustering and reweigthing in the end via estimating R\'enyi-entropy. While we have already argued that (i) does not work without reweighting, to which our solution has been (ii), some of the experiments presented so far indicate that obtaining the same number of samples from each sampler followed by the reweighting process (ii) (the ``uniform+clustering'' method in our previous experiments) might lead to almost the same or even superior performance than that of KSD-MCMC-WR in certain cases (see \Cref{fig:mse_vs_sample_multimode_bandit}, \Cref{fig:mse_vs_sample_multimode_bandit_nuts}, the MALA-results in \Cref{fig:ksd_parallel}, and \Cref{fig:ksd_parallel_dimension_1}). 

The reason behind this behavior seems to be a combination of how easily the samplers move from one mode to another, how this affects the clustering of the samplers, how meaningful it is to use the KSD to compare the sample quality of different samplers clustered together, and how many decisions the bandit algorithm can make.
In essence, when some of the modes (``connected'' regions) are close to each other so that samplers (particularly NUTS samplers) can easily go from one mode to another, samplers that explore these modes are often clustered together. At the same time, if the modes are far enough so that the KSD can signal good fit even if samples do not explore all these modes, the bandit algorithms can easily end up choosing samplers that only explore a small number of modes. Furthermore, even if the clusters are properly selected, when a sampler switches clusters, it may take a large number of steps until its KSD is comparable with that of other chains in the same group (since its KSD is dominated by the sample quality in other modes). Using a smaller batch size in KSD-MCMC-WR can help create smaller clusters (and hence avoid the aforementioned problems). In addition, it also increases the number of decisions to be made by the bandit algorithm, which allows the bandit algorithm to adapt better (in cases when the KSD is a reasonable measure).

\begin{figure}[t]
	\begin{center}
		\centerline{
			\includegraphics[width=0.50\columnwidth]{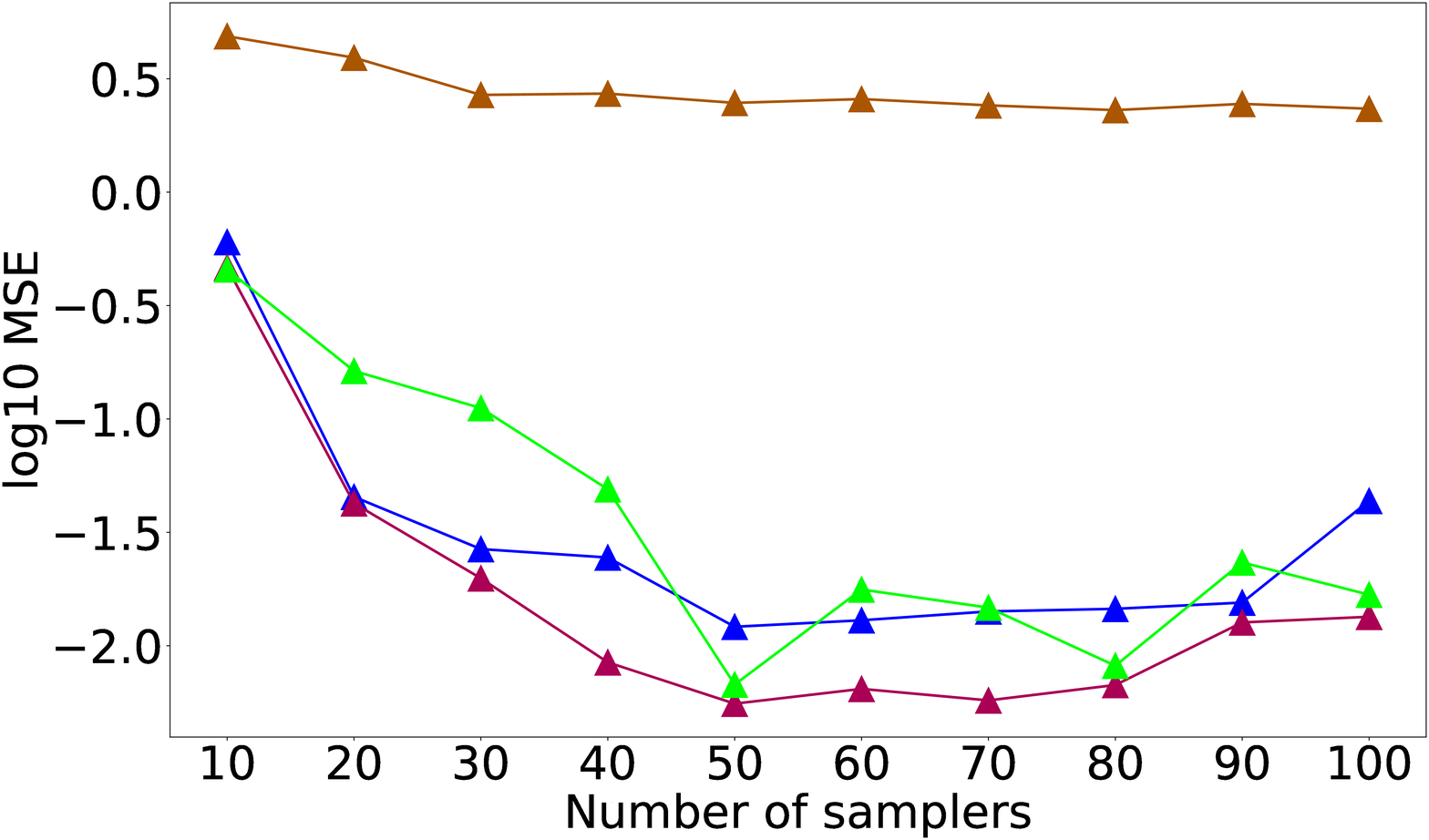}
			\hskip -22pt	\includegraphics[width=0.50\columnwidth]{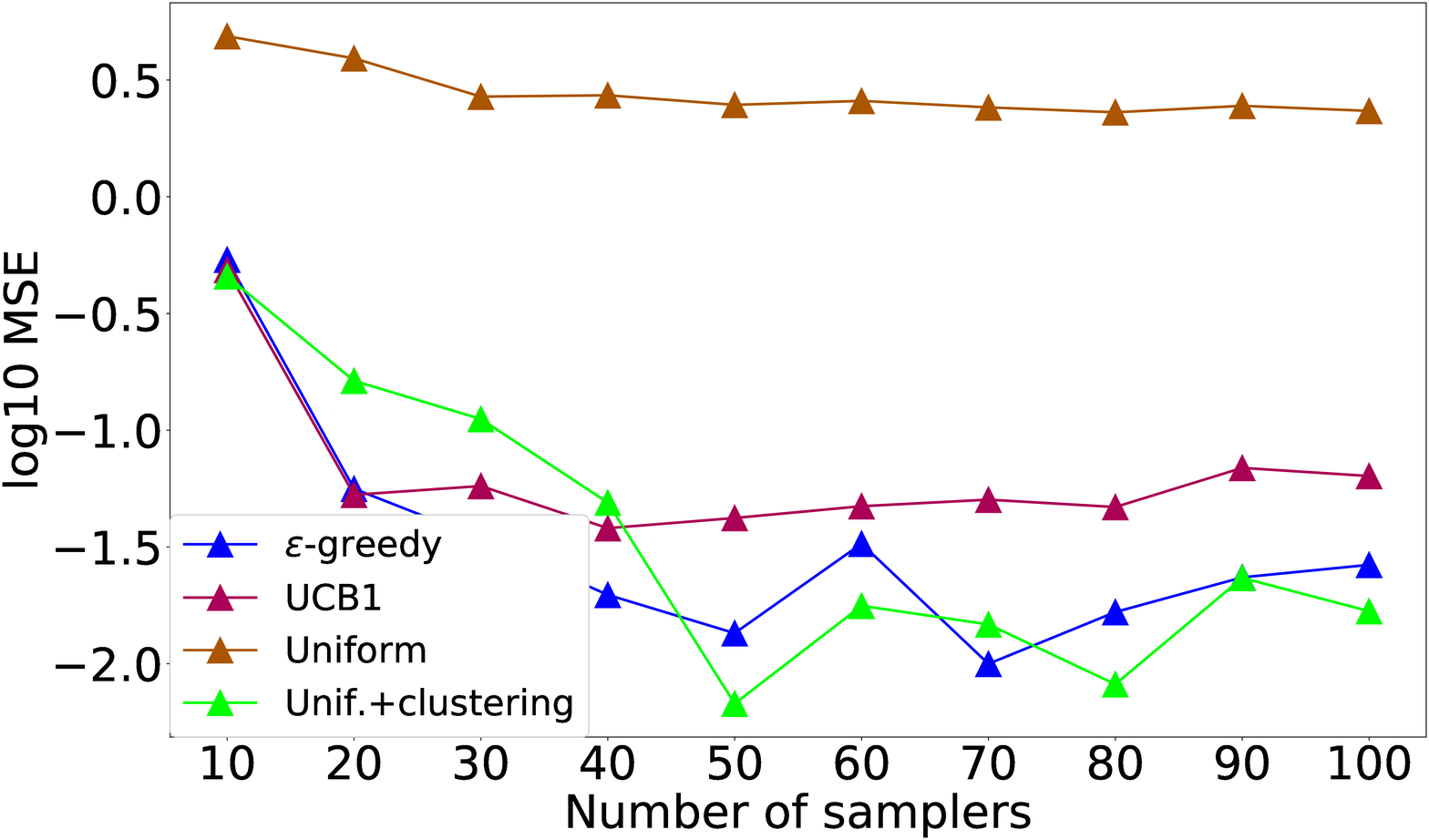} } 
		\centerline{
			\includegraphics[width=0.50\columnwidth]{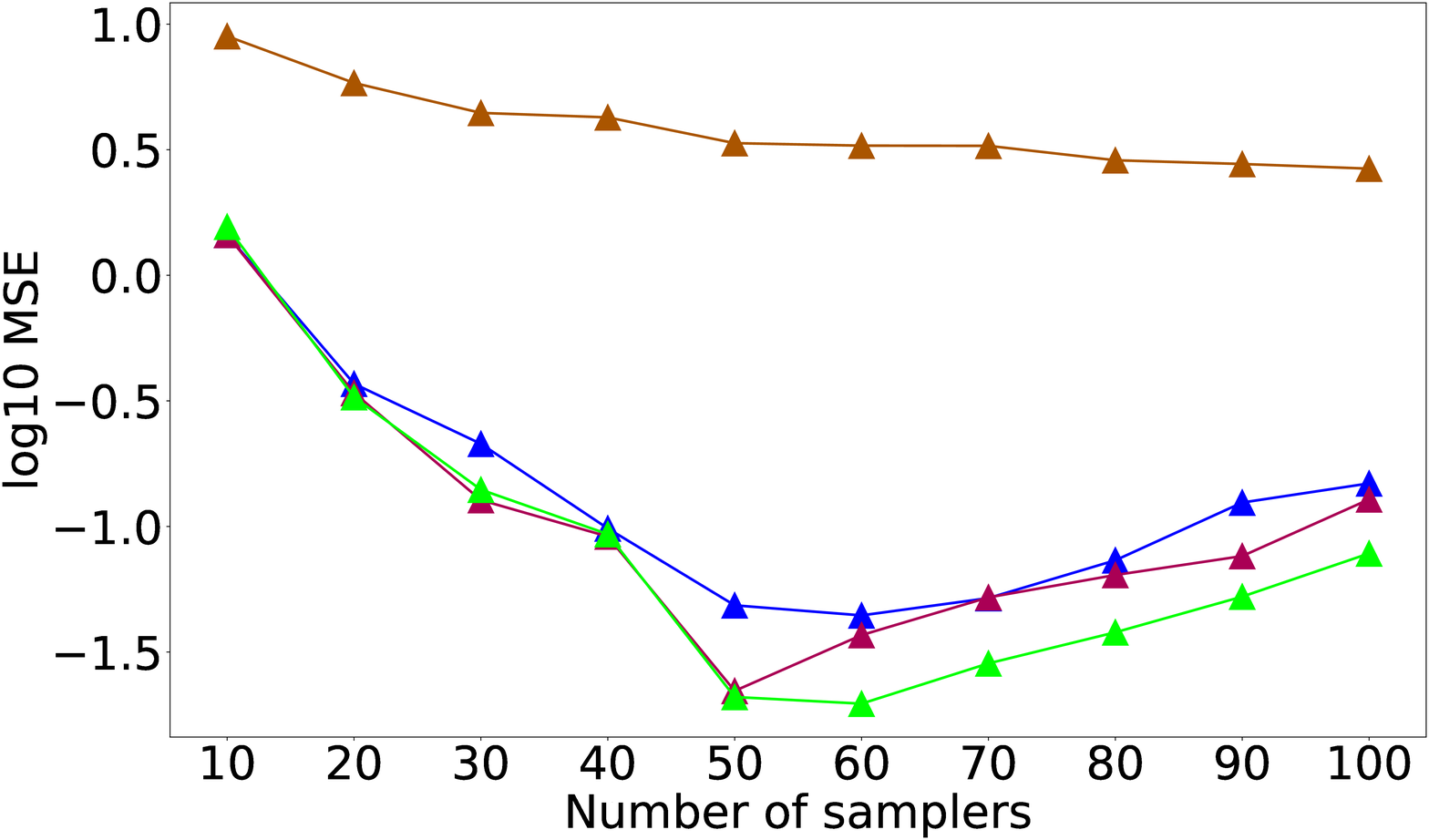}
			\hskip -22pt	\includegraphics[width=0.50\columnwidth]{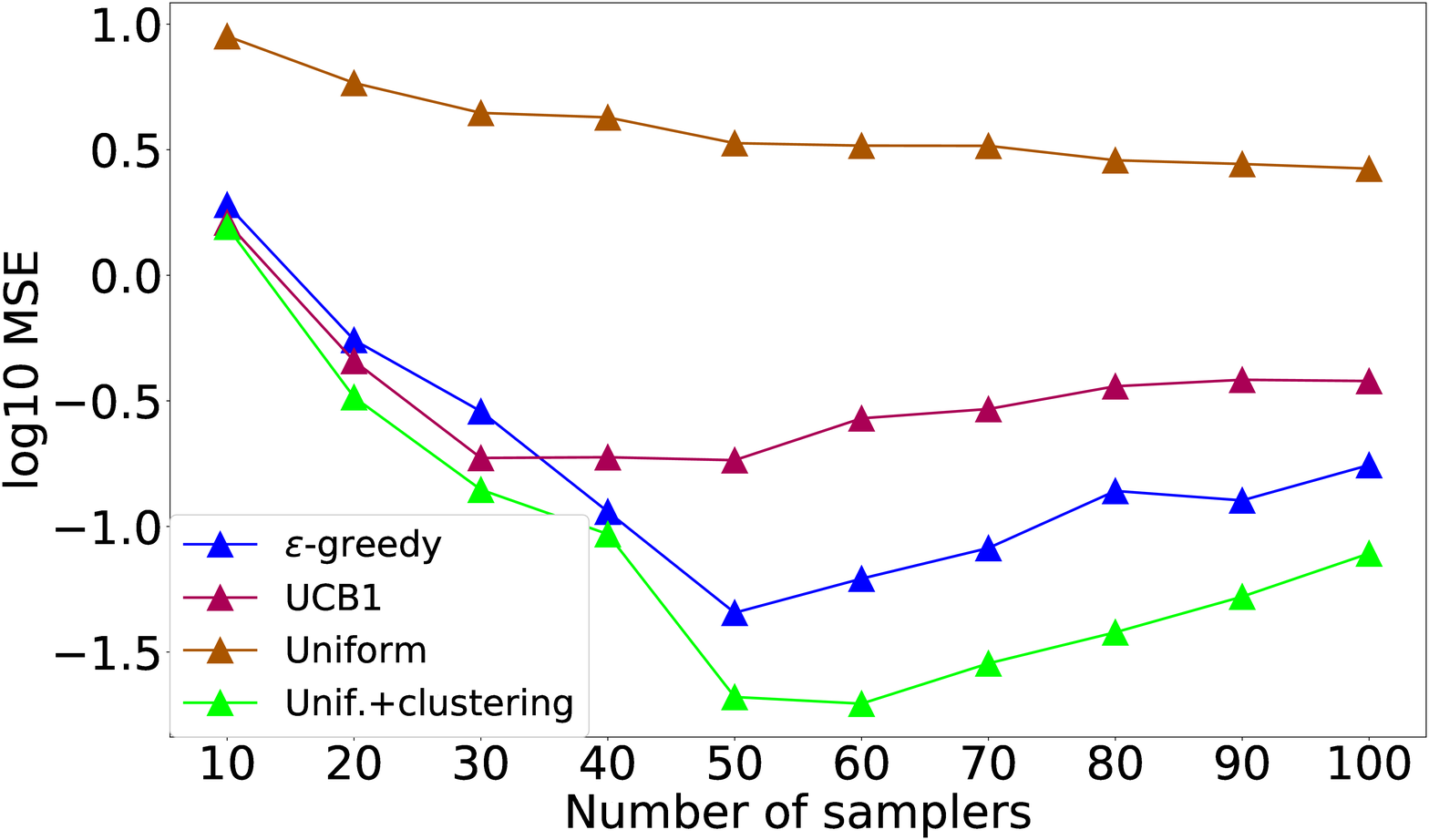} }
		\caption{MSE for different number of NUTS samplers as a function  of the number of samplers  for a  2-dimensional Gaussian mixture model with $20$ isotropic modes. The mean of each mode was selected uniformly at random from $[-10, 10]^2$ and the variance of each component was selected uniformly from $[0.2, 1]$. Two representative cases are shown: in the top row the modes are far from each other, in the bottom row some modes are close to each other. Batch size for KSD-MCMC-WR is $10$ on the left figure and $100$ on the right. Each curve is obtained by averaging over $100$ runs with different random initializations of the NUTS samplers; the total number of samples is $20000$ in each case.}
		\label{fig:chain_to_mode}
	\end{center}
\end{figure}

In the rest of this section we present some experiments verifying the above argument. Since we could only observe the superior behavior of the ``uniform+clustering'' algorithm for low-dimensional problems (this also holds for the relatively high, 18-dimensional experiments presented in the next section, where KSD-MCMC-WR consistently performs better), we consider  a Gaussian mixture model with $20$ modes of random parameters in $2$D and run KSD-MCMC-WR with $M$  NUTS samplers, $M \in \{10, 20, \ldots, 100\}$.%
\footnote{Note that here the number of samplers is much bigger (up to 100) than in any of our experiments.}
 As before, after randomly choosing the distribution and the samplers, we considered them fixed and ran $100$ experiments; in each of them the sampling methods are  started from random initial points. The whole process was then repeated for $10$ different sets of distributions and samplers with the total number of samples being $20000$ in each case. \Cref{fig:chain_to_mode} shows the results for two representative cases (out of $10$). The top row corresponds to a case that the modes are far from each other and UCB1 significantly outperforms the uniform+clustering algorithm, except for few points. The bottom row depicts a case when the modes are close to each other, in which case UCB1 performs slightly worse than uniform+clustering. This is more pronounced for when the number of samplers is large, as in this case more modes are clustered together, while the samplers belonging to each cluster do not explore all the modes belonging to the cluster. Also, a large number of samplers often results in a larger number of clusters, increasing the decision space of the bandit algorithm, which deteriorates its performance in general.

\subsection{Sensor network localization}
\label{sec:sensor}

In our last experiment we consider a simulated problem of  \emph{sensor network localization}. Following the setup of \citet{ihler_snl_2005} (see also the work of \citealp{ahan_regeneration_2013}), we consider $N$ sensors which are independently placed in a planar region $[-L/2, L/2]^2$ for some $L>0$ according to a density $p_x$. The distance between two sensors, $t$ and $u$, are observed with probability  $P_o(x_t, x_u) = \exp\left( -0.5 \| x_t - x_u\|_2^2/ R^2 \right)$ (for some $R>0$), and we define $o_{tu}=1$ if the distance is observed and $0$ otherwise. The observed distances are corrupted by independent zero-mean Gaussian noise $n_{tu}$ with variance $
\sigma^2$:  $d_{tu} = \|x_t-x_u\|_2 + n_{tu}$.

\begin{figure}[]
	\begin{center}
		\centerline{
			\includegraphics[width=0.5\columnwidth]{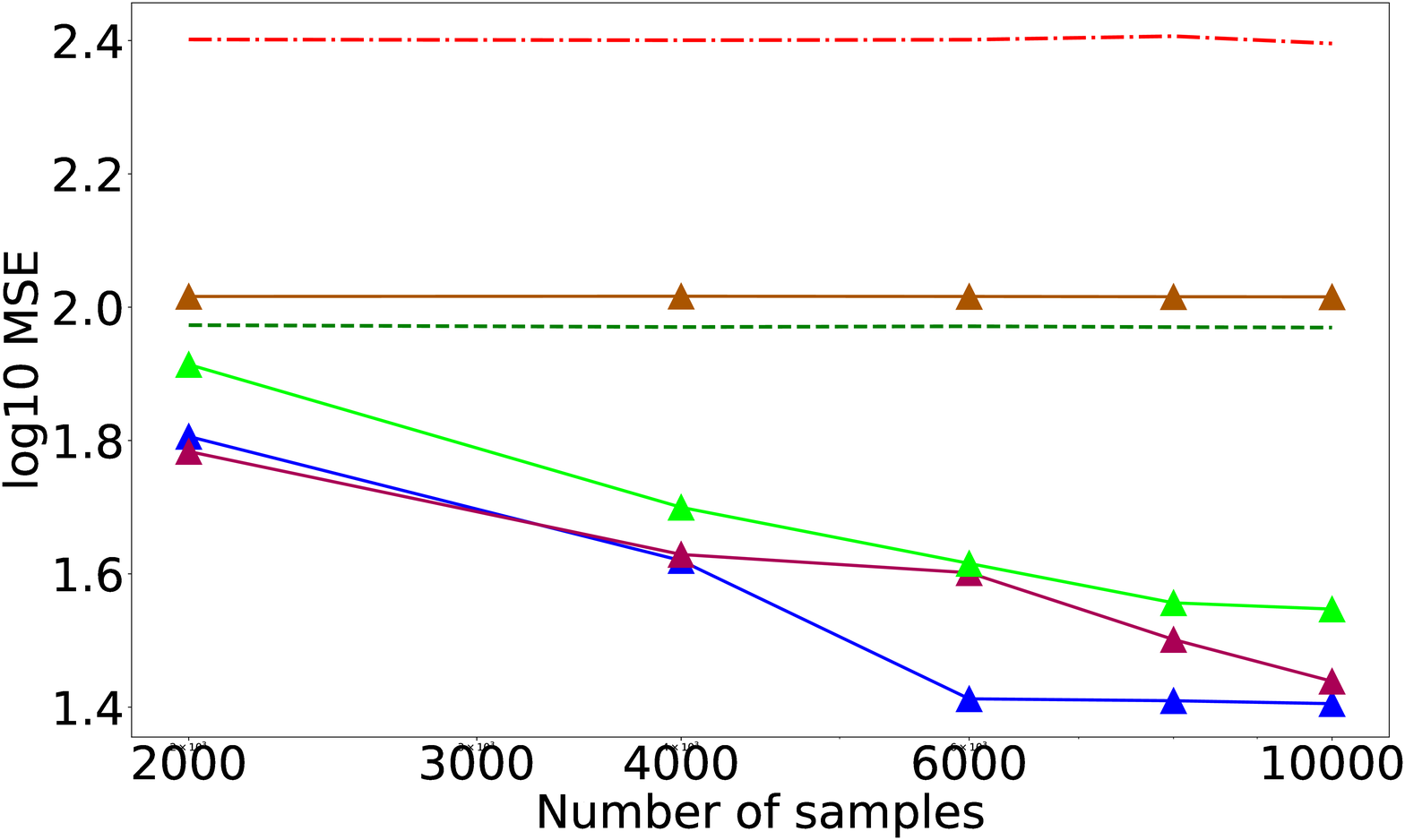} 
			\hskip -22pt
			\includegraphics[width=0.55\columnwidth]{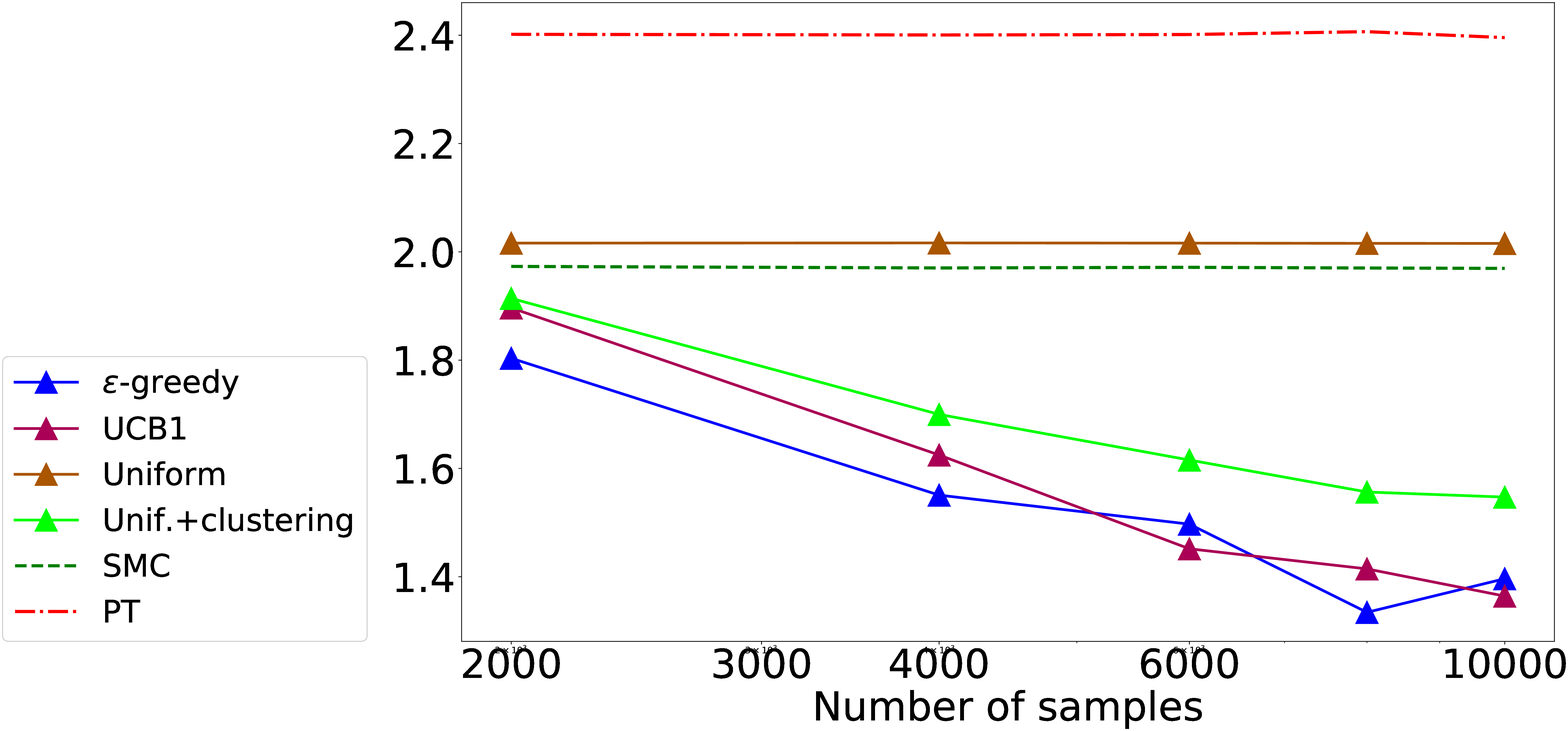} 
		}
		\tiny
		\caption{Running KSD-MCMC-WR for the sensor network localization problem with $10$ NUTS instances started with different (random) initialization. The dashed green lines show the results for the individual NUTS samplers. R\'enyi entropy  with $\alpha=0.99$ is used for weight estimation.The batch size is $10$ on the left and $100$ on the right. }
		\label{fig:bandit_snl_nuts}
	\end{center}
	\vspace{-0.2cm}
\end{figure}

Given a set of observations $\{d_{tu}\}$ (which only includes $t,u$ pairs with $o_{tu}=1$), the joint distribution of the observations and locations is given by  
\[
p(\{x_t\},  \{o_{tu}\}, \{d_{tu}\}) = \prod_{t, u}^{N}\psi(x_t,x_u,o_{tu})\prod_{t}^{N}p_x(x_t)
\]
where, denoting the density of $n_{tu}$ by $p_{\sigma^2}$,
\[
\psi(x_t,x_u,o_{tu}) = \begin{cases} P_o(x_t, x_u)p_{\sigma^2}(d_{tu}-\|x_t-x_u\|_2) & \quad \text{if $o_{tu}=1$;} \\
1-P_o(x_t,x_u) & \quad \text{if $o_{tu}=0$.}
\end{cases}
\]
The task is to infer the actual sensor locations from the observations.  In the experiment we set $N=8$, $R/L = 0.3$, $\sigma_v/L = 0.02$, $L=10$, $p_x$ is uniform, and we introduce three additional sensors with known locations, a.k.a. the anchor nodes, to avoid ambiguities due to translations, negations and rotations, as described by \citet{ahan_regeneration_2013}.  Since several pairwise distances are not available, this setup results in a multimodal posterior distribution given the observations. The performance is measured by the Euclidean distance between the actual  and the estimated locations from the sampling methods and we report the average results over $100$ runs.  We use KSD-MCMC-WR with NUTS and choosing each region uniformly at random.  \Cref{fig:bandit_snl_nuts} shows that our proposed algorithms significantly outperform all competitors, including single NUTS, parallel NUTS (labeled as uniform in the figures), SMC and PT for finding the sensor locations.%
\footnote{Unfortunately, we could not reproduce the results of \citet{ahan_regeneration_2013} because the paper does not provide enough details about how their algorithm is tuned.}

\section{Conclusion}
Selecting the best MCMC method and its best parameter setting for a given problem is a hard task. Even if the parameters are selected correctly, in case of multimodal distribution, MCMC chains can easily get trapped in different modes for a long time, which may lead to biased results. To solve these issues, in this paper we proposed an adaptive MCMC method, KSD-MCMC-WR,  that runs several chains in parallel, measures the quality of samples from each mode locally via kernel  Stein discrepancy, and decides sequentially how to allocate the sampling budget among the different samplers using multi-armed bandit algorithms. The final step of our method is to combine the samples obtained by the different samplers: for this we developed a novel weighting scheme based on R\'enyi-entropy estimation, which might be of independent interest. Extensive experiments on several  setups demonstrated that our proposed algorithm works well in both unimodal and multimodal problems, and it can perform particularly well when used in combination with NUTS as its base sampler.

\section*{Acknowledgement}

We would like to thank Firas Hamze for insightful discussions about different sampling methods and Tor Lattimore for his comments on earlier versions of the manuscript.

\bibliography{My_Library}

\end{document}